\documentclass{article}

\usepackage[margin=1in]{geometry}
\usepackage{amsfonts}       
\usepackage{bbm}
\usepackage{booktabs}       
\usepackage{clipboard}
\usepackage{comment}
\usepackage{esvect}
\usepackage{hyperref}       
\usepackage{mathtools}
\usepackage{microtype}      
\usepackage{nicefrac}       
\usepackage{tikz}
\usepackage{times}
\usepackage{url}            
\usepackage{xcolor}         
\usepackage[utf8]{inputenc} 
\usepackage[T1]{fontenc}    
\usepackage{amsthm}
\usepackage{natbib}

\usetikzlibrary{arrows, shapes}
\tikzstyle{bubble}= [rectangle, rounded corners, fill=gray!30, draw=black, minimum width=1.5cm]
\usepackage{graphicx}

\newenvironment{packed_itemize}{
\begin{itemize}
  \setlength{\itemsep}{1pt}
  \setlength{\parskip}{0pt}
  \setlength{\parsep}{0pt}
}{\end{itemize}}

\newcommand{\norm}[1]{\left\lVert#1\right\rVert}
\newcommand{\func}[2] {{#1}\left(#2\right)}
\newcommand{\tuple}[2] {\left(#1 , #2 \right)}
\newcommand{\prob}[2] {\underset{#1}{\text{Pr}}\left[#2\right]}
\newcommand{\expectation}[2] {\underset{#1}{\mathbb{E}}\left[#2\right]}

\newcommand{\cov}[3]{\underset{#1}{\text{Cov}} \left(#2, #3\right)}
\newcommand{\abs}[1]{\left| #1 \right|}
\newcommand{\naturals}{\mathbb{N}}
\newcommand{\reals}{\mathbb{R}}
\newcommand{\pos}{\reals^{+}}
\newcommand{\indicator}[2]{\func{\mathbbm{1}_{#1}}{#2}}
\newcommand{\lnFunc}[1]{\func{\ln}{#1}}
\newcommand{\expFunc}[1]{\func{\exp}{#1}}
\newcommand{\supFunc}[2]{\func{\underset{#1}{\sup}}{#2}}

\newcommand{\eqRes}[2]{\overset{\left( #1 \right)}{#2}}
\newcommand{\chiSqr}[2]{\func{\mathbf{D}_{\chi^{2}}}{#1 \Vert #2}}

\newcommand{\cpLem}[3]{
\begin{lemma}[#1] \label{lem:#2}
\Copy{cp:#2}{#3}
\end{lemma}
}
\newcommand{\pstLem}[1]{
\begin{lemma}[Lemma~\ref{lem:#1} restated]
\Paste{cp:#1}
\end{lemma}
}

\newcommand{\cpThm}[3]{
\begin{theorem}[#1] \label{thm:#2}
\Copy{cp:#2}{#3}
\end{theorem}
}
\newcommand{\pstThm}[1]{
\begin{theorem}[Theorem~\ref{thm:#1} restated]
\Paste{cp:#1}
\end{theorem}
}

\newcommand{\sampleSize}{n}
\newcommand{\numOfIterations}{k}
\newcommand{\rangeOfIterations}{\left[ \numOfIterations \right]}

\newcommand{\BayesF}[2]{\func{K}{#1, #2}}

\newcommand{\element}{x}
\newcommand{\elementRV}{X}
\newcommand{\elementsDomain}{\mathcal{\elementRV}}
\newcommand{\elementY}{y}
\newcommand{\elementRVY}{Y}

\newcommand{\domainOfSets}{\elementsDomain^{\sampleSize}}
\newcommand{\sampleSet}{s}
\newcommand{\sampleSetRV}{S}

\newcommand{\response}{r}
\newcommand{\responseRV}{R}
\newcommand{\responseFamily}{\mathcal{\responseRV}}

\newcommand{\view}{v}
\newcommand{\viewRV}{V}
\newcommand{\viewFamily}{\mathcal{\viewRV}}

\newcommand{\query}{q}
\newcommand{\queryRV}{Q}
\newcommand{\queryFamily}{\mathcal{\queryRV}}
\newcommand{\viewQuery}{\bar{\query}_\view}
\newcommand{\queryFunc}[1] {\func{\query}{#1}}
\newcommand{\queryFuncInd}[2] {\func{\query_{#1}}{#2}}

\newcommand{\mechanism}{M}
\newcommand{\mechanismFunc}[1] {\func{\mechanism}{#1}}

\newcommand{\analyst}{A}
\newcommand{\analystsFamily} {\mathcal{\analyst}}

\newcommand{\dist}{D}
\newcommand{\iiDist}{\dist^{\left( \sampleSize \right)}}
\newcommand{\distInd}[1] {\dist_{#1}}
\newcommand{\distP}{P}
\newcommand{\distQ}{Q}
\newcommand{\distUpInd}[2] {\distInd{#1}^{#2}}
\newcommand{\distDomain} {\distInd{\elementsDomain}}
\newcommand{\distDomainOfSets} {\distInd{\domainOfSets}}
\newcommand{\distDep}[3] {\distUpInd{#1|#2}{#3}}
\newcommand{\distJoint}[3] {\distUpInd{\tuple{#1}{#2}}{#3}}

\newcommand{\distFunc}[1] {\func{\dist}{#1}}
\newcommand{\distPfunc}[1]{\func{\distP}{#1}}
\newcommand{\distQfunc}[1]{\func{\distQ}{#1}}
\newcommand{\distFuncDep}[2] {\func{\dist}{#1 \,|\, #2}}
\newcommand{\distPost}{\distUpInd{}{\view}}

\newcommand{\DPMD}[2]{\le_{\tuple{\varepsilon_{#1}}{\delta_{#2}}}^{+}}
\newcommand{\DNMD}[2]{\le_{\tuple{\varepsilon_{#1}}{\delta_{#2}}}^{-}}

\newcommand{\PC}{\varphi}
    \newcommand{\PCfuncS}[2]{\func{\PC}{#1; #2}}
\newcommand{\PCfuncL}[3]{\func{\PC}{#1, #2; #3}}
\newcommand{\PCqFunc}[2]{\PCfuncL{#1}{#2}{\query}}
\newcommand{\PCvFunc}[2]{\PCfuncL{#1}{#2}{\view}}

\newcommand{\PI}{\varepsilon}
\newcommand{\PIfuncS}[2]{\func{\PI}{#1; #2}}
\newcommand{\PIfuncL}[3]{\func{\PI}{#1, #2; #3}}
\newcommand{\PIqFunc}[2]{\PIfuncL{#1}{#2}{\query}}
\newcommand{\PIvFunc}[2]{\PIfuncL{#1}{#2}{\view}}

\newcommand{\loss}[3]{\func{\ell}{#1, #2 ; #3}}
\newcommand{\lossS}[2]{\func{\ell}{#1; #2}}

\newcommand{\moshe}[1]{\textcolor{orange}{[M: #1]}}

\title{Generalization in the Face of Adaptivity: \\
A Bayesian Perspective}

\author{
    Moshe Shenfeld and  Katrina Ligett \\
    The Hebrew University of Jerusalem 
}

\begin{document}

\theoremstyle{plain}
\newtheorem{theorem}{Theorem}[section]
\newtheorem{lemma}[theorem]{Lemma}
\newtheorem{claim}[theorem]{Claim}
\newtheorem{corollary}[theorem]{Corollary}
\newtheorem{assumption}[theorem]{Assumption}
\newtheorem{proposition}[theorem]{Proposition}
\newtheorem{fact}[theorem]{Fact}

\theoremstyle{definition}
\newtheorem{definition}[theorem]{Definition}
\newtheorem{example}[theorem]{Example}
\theoremstyle{remark}
\newtheorem{remark}[theorem]{Remark}

\maketitle

\begin{abstract}
Repeated use of a data sample via adaptively chosen queries can rapidly lead to overfitting, wherein the empirical evaluation of queries on the sample significantly deviates from their mean with respect to the underlying data distribution. It turns out that simple noise addition algorithms suffice to prevent this issue, and differential privacy-based analysis of these algorithms shows that they can handle an asymptotically optimal number of queries.  However, differential privacy's worst-case nature entails scaling such noise to the range of the queries even for highly-concentrated queries, or introducing more complex algorithms.

In this paper, we prove that straightforward noise-addition algorithms already provide variance-dependent guarantees that also extend to unbounded queries. This improvement stems from a novel characterization that illuminates the core problem of adaptive data analysis. We show that the harm of adaptivity results from the covariance between the new query and a Bayes factor-based measure of how much information about the data sample was encoded in the responses given to past queries. We then leverage this characterization to introduce a new data-dependent stability notion that can bound this covariance.
\end{abstract}

\section{Introduction}
Recent years have seen growing recognition of the role of \emph{adaptivity} in causing overfitting and thereby reducing the accuracy of the conclusions drawn from data. Intuitively, allowing a data analyst to adaptively choose the queries that she issues potentially leads to misleading conclusions, because the results of prior queries might encode information that is specific to the available data sample and not relevant to the underlying distribution from which it was drawn. As a result, future queries might then be chosen to leverage these sample-specific properties, giving answers on the sample that differ wildly from the values of those queries on the data distribution. Such adaptivity has been blamed, in part, for the current reproducibility crisis in the data-driven sciences~\citep{Ioannidis05,GL14}.

A series of works catalyzed by~\citet{DFHPR15}
recently established a formal framework for understanding and analyzing adaptivity in data analysis, and introduced a general toolkit for provably preventing the harms of choosing queries adaptively---that is, as a function of the results of previous queries. This line of work has established that enforcing that computations obey a constraint of \emph{differential privacy}~\citep{DMNS06}---which one can interpret as a robustness or stability requirement on computations---provably limits the extent to which adaptivity can cause overfitting. Practically, for simple numerical computations, these results translate into adding a level of noise to the query results that would be sufficient to ensure \emph{worst-case} stability of the \emph{worst-case} query on the \emph{worst-case} dataset. In particular, when analyzed using differential privacy, simple noise-addition mechanisms must add noise that scales with the \emph{worst-case} change in query value that could be induced by changing a single input data element, in order to protect against adaptivity.

However, this can be overkill: for the purposes of statistical validity, we do not care about the worst case but the typical case.
In the present work, we prove the following new and better generalization guarantees for simple Gaussian noise-addition algorithms:
\begin{theorem} [Informal versions of main theorems]
    With probability $> 1 - \delta$, the error of the responses produced by a mechanism which only adds Gaussian noise to the empirical values of the queries it receives is bounded by $\epsilon$, even after responding to $\numOfIterations$ adaptively chosen queries, if
    \begin{packed_itemize}
        \item the range of the queries is bounded by $\Delta$, their variance is bounded by $\sigma^{2}$, and the size of the dataset $\sampleSize = \func{\Omega}{\max \left\{\frac{\Delta}{\epsilon}, \frac{\sigma^{2}}{\epsilon^{2}} \right\} \sqrt{\numOfIterations} \cdot \lnFunc{\frac{\numOfIterations \sigma}{\delta \epsilon}}}$ (Theorem \ref{thm:GaussMechAccBnd}), or
        \item the queries are $\sigma^{2}$-sub-Gaussian and the size of the dataset $\sampleSize = \func{\Omega}{\frac{\sigma^{2} \sqrt{\numOfIterations}}{\epsilon^{2}} \lnFunc{\frac{\numOfIterations \sigma}{\delta \epsilon}}}$ (Theorem \ref{thm:GaussMechAccSubGauss}).
    \end{packed_itemize}
    \end{theorem}

In contrast, analyzing the same mechanism using differential privacy yields a sample size requirement that scales with $\sampleSize = \func{\Omega}{\frac{\Delta^{2} \sqrt{\numOfIterations}}{\epsilon^{2}} \lnFunc{\frac{\numOfIterations \sigma}{\delta \epsilon}}}$ in the first setting, and provides no guarantee in the latter.

We prove these theorems via a new notion, \emph{pairwise concentration (PC)} (Definition \ref{def:PC}), which captures the extent to which replacing one dataset by another would be ``noticeable,'' given a particular query-response sequence. This is thus a function of particular differing datasets (instead of worst-case over elements), and it also depends on the actual issued queries. We then build a composition toolkit (Theorem \ref{thm:PCcomp}) that allows us to track PC losses over multiple computations. 
The PC notion allows for more careful analysis of the information encoded by the query-response sequence than differential privacy does.

In order to leverage this more careful analysis of the information encoded in query-response sequences, we rely on a simple new characterization (Lemma \ref{lem:covStab}) that 
shows explicitly that the harms of adaptivity come from the covariance between the behavior of the future queries and a Bayes factor-based measure of how much information about the data sample was encoded in the responses to the issued queries. Our characterization
gives insight into how differential privacy protects against adaptivity, and then allows us to step away from this worst-case approach. What differential privacy accomplishes is that it bounds a worst-case version of the Bayes factor; however, as this new characterization makes clear, it is sufficient to bound the ``typical'' Bayes factor in order to avoid overfitting.

We measure the harm that past adaptivity causes to a future query by considering the query as evaluated on a posterior data distribution and comparing this with its value on a prior. The prior is the true data distribution, and the posterior is induced by observing the responses to past queries and updating the prior. If the new query behaves similarly on the prior distribution as it does on this posterior (a guarantee we call \emph{Bayes stability}; Definition \ref{def:bysStab}), adaptivity has not led us too far astray.\footnote{This can be viewed as a generalization of the \emph{Hypothesis Stability} notion of~\citet{BE02}---which was proven to guarantee on-average generalization~\citep{SSSSS10}---where the hypothesis is a post-processing of the responses to past queries, and the future query is the loss function estimation.} If furthermore, the the response given by the mechanism is close to the query result on the posterior, then by a triangle inequality argument, that mechanism is distribution accurate. This type of triangle inequality first appeared as an analysis technique in \citet{JLNRSS20}.

The dependence of our PC notion on the actual adaptively chosen queries places it in the so-called \emph{fully-adaptive} setting~\citep{RRUV16, WRRW23}, which requires a fairly subtle analysis involving a set of tools and concepts that may be of independent interest. In particular, we establish a series of ``dissimilarity'' notions in Appendix \ref{apd:divAndDis}, which generalize the notion of divergence, replacing the scalar bound with a function. Our main stability notion (Definition \ref{def:PC}) can be viewed as an instance-tailored variant of \emph{zero-concentrated differential privacy~\citep{BS16}}, and we also make use of a similar extension of the classical max-divergence-based differential privacy definition (\ref{def:dynMaxDis}).

\paragraph{Related work}

\emph{Differential privacy} \citep{DMNS06} is a privacy notion based on a bound on the max divergence between the output distributions induced by any two neighboring input datasets (datasets which differ in one element). One natural way to enforce differential privacy is by directly adding noise to the results of a numeric-valued query, where the noise is calibrated to the \emph{global sensitivity} of the function to be computed---the maximal change in its value between any two neighboring datasets.~\citet{DFHPR15} and~\citet{BNSSSU21} showed that differential privacy is also useful as tool for ensuring generalization in settings where the queries are chosen adaptively. 

Differential privacy essentially provides the optimal asymptotic generalization guarantees given adaptive queries~\citep{HU14, SU15}. However, its optimality is for worst-case adaptive queries, and the guarantees that it offers only beat the naive intervention---of splitting a dataset so that each query gets fresh data---when the input dataset is quite huge~\citep{JLNRSS20}. A worst-case approach makes sense for privacy, but for statistical guarantees like generalization, we only need statements that hold with high probability with respect to the sampled dataset, and only on the actual queries issued.

One cluster of works that steps away from this worst-case perspective focuses on giving privacy guarantees that are tailored to the dataset at hand~\citep{NRS07,GR11, ESS15,Wang19}. In ~\citet{FZ20} in particular, the authors elegantly manage to track the individual privacy loss of the elements in the dataset. However, their results do not enjoy a dependence on the standard deviation in place of the range of the queries.
Several truncation-based specialized mechanisms have been proposed, both to provide differential privacy guarantees for Gaussian and sub-Gaussian queries even in the case of multivariate distribution with unknown covariance~\citep{KV17, AL21, DHK23} and, remarkably, design specialized algorithms that achieve adaptive data analysis guarantees that scale like the standard deviation of the queries~\citep{FS17}. Recently,~\citet{Blanc23} proved that randomized rounding followed by sub-sampling provides accuracy guarantees that scale with the queries' variance. But none of these results apply to simple noise addition mechanisms.

Another line of work (e.g.,~\citet{GHLP12,BGKS13,BBGLT11}) proposes relaxed privacy definitions that leverage the natural noise introduced by dataset sampling to achieve more average-case notions of privacy. This builds on intuition that average-case privacy can be viewed from a Bayesian perspective, by restricting some distance measure between some prior distribution and some posterior distribution induced by the mechanism's behavior~\citep{DMNS06, KS14}. This perspective was used \citet{LS19} to propose a stability notion which is both necessary and sufficient for adaptive generalization under several assumptions. Unfortunately, these definitions have at best extremely limited adaptive composition guarantees. ~\citet{BF16} connect this Bayesian intuition to statistical validity via \emph{typical stability}, an approach that discards ``unlikely'' databases that do not obey a differential privacy guarantee, but their results require a sample size that grows linearly with the number of queries even for iid distributions. \citet{TF20} propose the notion of \emph{Bayesian differential privacy} which leverages the underlying distribution to improve generalization guarantees, but their results still scale with the range in the general case.  

An alternative route for avoiding the dependence on worst case queries and datasets was achieved using expectation based stability notions such as \emph{mutual information} and \emph{KL stability}~\citet{RZ16, BNSSSU21, SZ20}. Using these methods \citet{FS18} presented a natural noise addition mechanism, which adds noise that scales with the empirical variance when responding to queries with known range and unknown variance. Unfortunately, in the general case, the accuracy guarantees provided by these methods hold only for the expected error rather than with high probability.

A detailed comparison to other lines of work can be found in Appendix \ref{apd:relNot}.

\section{Preliminaries}

\subsection{Setting} \label{subsec:setting}
We study datasets, each of fixed size $\sampleSize \in \naturals$, whose elements are drawn from some domain $\elementsDomain$.
We assume there exists some distribution $\distDomainOfSets$ defined over the datasets $\sampleSet \in \domainOfSets$.\footnote{Throughout the paper, the notation $\distFunc{\cdot}$ describing a distribution over a domain, will either represent the probability mass function defined by a discrete probability over a countable sample space or the probability density function (the Radon–Nikodym derivative) in case of a measure over measurable spaces.} We consider a family of functions (\emph{queries}) $\queryFamily$ of the form $\query: \domainOfSets \rightarrow \responseFamily \subseteq \reals$. We refer to the functions' outputs as \emph{responses}.\footnote{Most of the definitions and claims in this paper can be extended beyond $\reals$ and the Euclidean norm. To do so, the divergence in Lemma \ref{lem:covStab} and the $\PC$ function (Definition \ref{def:PC}) must be chosen accordingly.}

We denote by $\queryFunc{\distDomainOfSets} \coloneqq \expectation{\sampleSetRV \sim \distDomainOfSets}{\queryFunc{\sampleSetRV}}$ the mean of $\query$ with respect to the distribution $\distDomainOfSets$, and think of it as the true value of the query $\query$, which we wish to estimate. As is the case in many machine learning or data-driven science settings, we do not have direct access to the distribution over datasets, but instead receive a dataset $\sampleSet$ sampled from $\distDomainOfSets$, which can be used to compute $\queryFunc{\sampleSet}$---the empirical value of the query.

A \emph{mechanism} $\mechanism$ is a (possibly non-deterministic) function $\mechanism: \domainOfSets \times \queryFamily \times \Theta \rightarrow \responseFamily$ which, given a dataset $\sampleSet \in \domainOfSets$, a query $\query \in \queryFamily$, and some auxiliary parameters $\theta \in \Theta$, provides some response $\response \in \responseFamily$ (we omit $\theta$ when not in use). The most trivial mechanism would simply return $\response = \queryFunc{\sampleSet}$, but we will consider mechanisms that return a noisy version of $\queryFunc{\sampleSet}$.
A mechanism induces a distribution over the responses $\func{\distUpInd{\responseFamily | \domainOfSets}{\query, \theta}}{\cdot \,|\, \sampleSet}$---indicating the probability that each response $\response$ will be the output of the mechanism, given $\sampleSet$, $\query$, and $\theta$ as input. Combining this with $\distDomainOfSets$ induces the marginal distribution $\distUpInd{\responseFamily}{\query, \theta}$ over the responses, and the conditional distribution $\func{\distUpInd{\domainOfSets | \responseFamily}{\query, \theta}}{\cdot \,|\, \response}$ over sample sets. Considering the uniform distribution over elements in the sample set induces similar distributions over the elements $\distInd{\elementsDomain}$ and $\distUpInd{}{\response} \coloneqq  \func{\distUpInd{\elementsDomain | \responseFamily}{\query, \theta}}{\cdot \,|\, \response}$.
This last distribution, $\distUpInd{}{\response}$, is central to our analysis, and it represents the \emph{posterior distribution} over the sample elements given an observed response---a Bayesian update with respect to a prior of $\distDomain$. All of the distributions discussed in this subsection are more formally defined in Appendix \ref{apd:formDefs}.

An \emph{analyst} $\analyst$ is a (possibly non-deterministic) function $\analyst: \responseFamily^{*} \rightarrow \tuple{\queryFamily}{\Theta}$ which, given a sequence of responses, provides a query and parameters to be asked next. An interaction between a mechanism and an analyst--- wherein repeatedly (and potentially adaptively) the analyst generates a query, then the mechanism generates a response for the analyst to observe---generates a sequence of $\numOfIterations$ queries and responses for some $\numOfIterations \in \naturals$, which together we refer to as a \emph{view} $\view_{\numOfIterations} = \left(\response_{1}, \response_2, \ldots, \response_{\numOfIterations} \right) \in \viewFamily_{\numOfIterations}$ (we omit $\numOfIterations$ when it is clear from context). In the case of a non-deterministic analyst, we add its coin tosses as the first entry in the view, in order to ensure each $\tuple{\query_{i}}{\theta_{i}}$ is a deterministic function of $\view_{i-1}$, as detailed in Definition \ref{def:adap}; given a view $\view \in \viewFamily$, we denote by $\viewQuery$ the sequence of queries that it induces. Slightly abusing notation, we denote $\view = \mechanismFunc{\sampleSet, \analyst}$. The distributions $\distUpInd{\responseFamily | \domainOfSets}{\query, \theta}$, $\distUpInd{\responseFamily}{\query, \theta}$, $\distUpInd{\domainOfSets | \responseFamily}{\query, \theta}$, and $\distUpInd{}{\response}$ naturally extend to versions where a sequence of queries is generated by an analyst: $\distUpInd{\viewFamily | \domainOfSets}{\analyst}$, $\distUpInd{\viewFamily}{\analyst}$, $\distUpInd{\domainOfSets | \viewFamily}{\analyst}$, and $\distPost \coloneqq  \func{\distUpInd{\elementsDomain | \viewFamily}{\analyst}}{\cdot \,|\, \view}$.

Notice that $\mechanism$ holds $\sampleSet$ but has no access to $\dist$. On the other hand, $\analyst$ might have access to $\dist$, but her only information regarding $\sampleSet$ comes from $\mechanism$'s responses as represented by $\distPost$. This intuitively turns some metric of distance between $\dist$ and $\distPost$ into a measure of potential overfitting, an intuition that we formalize in Definition \ref{def:bysStab} and Lemma \ref{lem:covStab}.

\subsection{Notation}
Throughout the paper, calligraphic letters denote domains (e.g., $\elementsDomain$), lower case letters denote elements of domains (e.g., $\element \in \elementsDomain$), and capital letters denote random variables (e.g., $\elementRV \sim \distDomain$).

We omit most superscripts and subscripts when clear from context (e.g., $\distFuncDep{\response}{\sampleSet} = \func{\distUpInd{\responseFamily | \domainOfSets}{\query, \theta}}{\response \,|\, \sampleSet}$ is the probability to receive $\response \in \responseFamily$ as a response, conditioned on a particular input dataset $\sampleSet$, given a query $\query$ and parameters $\theta$). Unless specified otherwise, we assume $\distDomainOfSets$, $\sampleSize$, $\numOfIterations$, and $\mechanism$ are fixed, and omit them from notations and definitions.

We use  $\norm{\cdot}$  to denote the Euclidean norm, so $\norm{\func{\viewQuery}{\element}} = \sqrt{\sum_{i=1}^{\numOfIterations} \left(\queryFuncInd{i}{\element} \right)^{2}}$ will denote the norm of a concatenated sequence of queries.

\subsection{Definitions}
We introduce terminology to describe the accuracy of responses that a mechanism produces in response to queries.

\begin{definition} [Accuracy of a mechanism] \label{def:acc}
    Given a dataset $\sampleSet$, an analyst $\analyst$, and a view $\view = \mechanismFunc{\sampleSet, \analyst}$, we define three types of output error for the mechanism: \emph{sample error} $\func{\text{err}_{S}}{\sampleSet, \view} \coloneqq \underset{i \in \rangeOfIterations}{\max} \abs{\response_{i} - \queryFuncInd{i}{\sampleSet}}$, \emph{distribution error} $\func{\text{err}_{D}}{\sampleSet, \view} \coloneqq \underset{i \in \rangeOfIterations}{\max} \abs{\response_{i} - \queryFuncInd{i}{\dist}}$, and \emph{posterior error} $\func{\text{err}_{P}}{\sampleSet, \view} \coloneqq \underset{i \in \rangeOfIterations}{\max} \abs{\response_{i} - \queryFuncInd{i}{\distPost}}$, where $\query_{i}$ is the $i$th query and $\response_{i}$ is the $i$th response in $\view$.
    
    These errors can be viewed as random variables with a distribution that is induced by the underlying distribution $\dist$ and the internal randomness of $\mechanism$ and $\analyst$.
    Given $\epsilon, \delta \ge 0$ we call a mechanism $\mechanism$ \emph{$\tuple{\epsilon}{\delta}$-sample/distribution/posterior accurate} with respect to $\analyst$ if
    \[
        \prob{\sampleSetRV \sim \iiDist, \viewRV \sim \mechanismFunc{\sampleSetRV, \analyst}}{ \left( \func{\text{err}}{\sampleSetRV, \viewRV} \right) > \epsilon} \le \delta,
    \]
    for $\text{err} = \text{err}_{S}$/$\text{err}_{D}$/$\text{err}_{P}$, respectively.
    
    Notice that if each possible value of $\epsilon$ has a corresponding 
    $\delta$, then $\delta$ is essentially a function of $\epsilon$. Given such a function $\delta : \pos \rightarrow \left[0, 1 \right]$, we will say the mechanism is $\tuple{\epsilon}{\func{\delta}{\epsilon}}$ accurate, a perspective which will be used in Lemma \ref{lem:samAccImpPostAcc}.
\end{definition}

We start by introducing a particular family of queries known as \emph{linear queries}, which will be used to state the main results in this paper, but it should be noted that many of the claims extend to arbitrary queries as discussed in Section \ref{sec:arbQuer}

\begin{definition}[Linear queries]\label{def:linearqueries}
    A function $\query : \domainOfSets \rightarrow \reals$ is a \emph{linear query} if it is defined by a function $\query_{1} : \elementsDomain \rightarrow \reals$ such that $\queryFunc{\sampleSet} \coloneqq \frac{1}{\sampleSize} \overset{\sampleSize}{\underset{i = 1}{\sum}} \func{\query_{1}}{\sampleSet_{i}}$. For simplicity, we denote $\query_{1}$ as $\query$ throughout.
\end{definition}

The aforementioned error bounds implicitly assume some known scale of the queries, which is usually chosen to be their range, denoted by $\Delta_{\query} \coloneqq \underset{\element, \elementY \in \elementsDomain}{\sup}\abs{\queryFunc{\element} - \queryFunc{\elementY}}$ (we refer to such a query as $\Delta$-bounded). This poses an issue for concentrated random variables, where the ``typical'' range can be arbitrarily smaller than the range, which might even be infinite.
A natural alternative source of scale we consider in this paper is the query's variance $\sigma_{\query}^{2} \coloneqq \expectation{\elementRV \sim \distDomain}{\left(\queryFunc{\elementRV} - \queryFunc{\distDomain} \right)^{2}}$, or its variance proxy in the case of a sub-Gaussian query (Definition \ref{def:subGauss}). The corresponding definitions for arbitrary queries are presented in Section \ref{sec:arbQuer}.

Our main tool for ensuring generalization is the Gaussian mechanism, which simply adds Gaussian noise to the empirical value of a query.

\begin{definition} [Gaussian mechanism] \label{def:GausMech}
    Given $\eta > 0$ and a query $\query$, the \emph{Gaussian mechanism} with noise parameter $\eta$ returns its empirical mean $\queryFunc{\sampleSet}$ after adding a random value, sampled from an unbiased Gaussian distribution with variance $\eta^{2}$. Formally, $\mechanismFunc{\sampleSet, \query} \sim \func{\mathcal{N}}{\queryFunc{\sampleSet}, \eta^{2}}$.\footnote{In the case of an adaptive process, one can also consider the case where $\eta_{i}$ are adaptively chosen by the analyst and provided to the mechanism as the auxiliary parameter $\theta_{i}$.}
\end{definition}

\section{Analyzing adaptivity-driven overfitting} \label{sec:gen}
In this section, we give a clean, new characterization of the harms of adaptivity. Our goal is to bound the distribution error of a mechanism that responds to queries generated by an adaptive analyst.
This bound will be achieved via a triangle inequality, by bounding both the posterior accuracy and the Bayes stability (Definition \ref{def:bysStab}). Missing proofs from this section appear in Appendix \ref{apd:gen}.

The simpler part of the argument is posterior accuracy, which we prove can be inherited directly from the sample accuracy of a mechanism. This lemma resembles Lemma 6 in~\citet{JLNRSS20}, but has the advantage of being independent of the range of the queries.

\cpLem{Sample accuracy implies posterior accuracy}{samAccImpPostAcc}{
    Given a function $\delta : \reals \rightarrow \left[ 0, 1 \right]$ and an analyst $\analyst$, if a mechanism $\mechanism$ is $\tuple{\epsilon}{\func{\delta}{\epsilon}}$-sample accurate for all $\epsilon > 0$, then $\mechanism$ is $\tuple{\epsilon}{\func{\delta'}{\epsilon}}$-posterior accurate for $\func{\delta'}{\epsilon} \coloneqq \underset{\xi \in \left(0, \epsilon \right)}{\inf} \left(\frac{1}{\xi} \int_{\epsilon - \xi}^{\infty} \func{\delta}{t} dt \right)$.
}

We use this lemma to provide accuracy guarantees for the Gaussian mechanism.

\cpLem{Accuracy of Gaussian mechanism}{GaussAcc}{
    Given $\eta > 0$, the Gaussian mechanism with noise parameter $\eta$ that receives $\numOfIterations$ queries is $\tuple{\epsilon}{\func{\delta}{\epsilon}}$-sample accurate for $\func{\delta}{\epsilon} \coloneqq \frac{2 \numOfIterations}{\sqrt{\pi}} e^{-\frac{\epsilon^{2}}{2 \eta^{2}}}$, and $\tuple{\epsilon}{\func{\delta}{\epsilon}}$-posterior accurate for $\func{\delta}{\epsilon} \coloneqq 4 \numOfIterations \cdot e^{-\frac{\epsilon^{2}}{4 \eta^{2}}}$.
}

In order to complete the triangle inequality, we have to define the stability of the mechanism. Bayes stability captures the concept that the results returned by a mechanism and the queries selected by the adaptive adversary are such that the queries behave similarly on the true data distribution and on the posterior distribution induced by those results. This notion first appeared in ~\citet{JLNRSS20}, under the name \emph{Posterior Sensitivity}, as did the following theorem.

\begin{definition} [Bayes stability] \label{def:bysStab}
    Given $\epsilon, \delta > 0$ and an analyst $\analyst$, we say $\mechanism$ is \emph{$\tuple{\epsilon}{\delta}$-Bayes stable} with respect to $\analyst$, if 
    \[
        \prob{\genfrac{}{}{0pt}{}{\sampleSetRV \sim \iiDist}{\viewRV \sim \mechanismFunc{\sampleSetRV, \analyst}, \queryRV \sim \func{\analyst}{\viewRV}}}{\abs{\func{\queryRV}{\dist^{\viewRV}} - \func{\queryRV}{\dist}} > \epsilon} \le \delta.
    \]
\end{definition}

Bayes stability and sample accuracy (implying posterior accuracy) combine via a triangle inequality to give distribution accuracy.

\cpThm{Generalization}{BayesStabImpdistPcc}{
    Given two functions $\delta_{1} : \reals \rightarrow \left[ 0, 1 \right]$, $\delta_{2} : \reals \rightarrow \left[ 0, 1 \right]$, and an analyst $\analyst$, if a mechanism $\mechanism$ is $\tuple{\epsilon}{\func{\delta_{1}}{\epsilon}}$-Bayes stable and $\tuple{\epsilon}{\func{\delta_{2}}{\epsilon}}$-sample accurate with respect to $\analyst$, then $\mechanism$ is $\tuple{\epsilon}{\func{\delta'}{\epsilon}}$-distribution accurate for $\func{\delta'}{\epsilon} \coloneqq \underset{\epsilon' \in \left(0, \epsilon \right), \xi \in \left(0, \epsilon - \epsilon' \right)}{\inf} \left(\func{\delta_{1}}{\epsilon'} + \frac{1}{\xi} \int_{\epsilon - \epsilon' - \xi}^{\infty} \func{\delta_{2}}{t} dt \right)$.
}

Since achieving posterior accuracy is relatively straightforward, guaranteeing Bayes stability is the main challenge in leveraging this theorem to achieve distribution accuracy with respect to adaptively chosen queries. The following lemma gives a useful and intuitive characterization of the quantity that the Bayes stability definition requires be bounded. Simply put, the Bayes factor $\BayesF{\cdot} {\cdot}$ (defined in the lemma below) represents the amount of information leaked about the dataset during the interaction with an analyst, by moving from the prior distribution over 
data elements to the posterior induced by some view $\view$. The degree to which a query $\query$ overfits to the dataset is expressed by the correlation between the query and that Bayes factor. This simple lemma is at the heart of the progress that we make in this paper, both in our intuitive understanding of adaptive data analysis, and in the concrete results we show in subsequent sections. Its corresponding version for arbitrary queries are presented in Section \ref{sec:arbQuer}.

\begin{lemma}[Covariance stability] \label{lem:covStab}
    Given a view $\view \in \viewFamily$ and a linear query $\query$,
    \[
        \queryFunc{\distPost} - \queryFunc{\distDomain} = \cov{\elementRV \sim \distDomain}{\queryFunc{\elementRV}}{\BayesF{\elementRV}{\view}}.
    \]
    
    Furthermore, given $\Delta, \sigma > 0$,
    \[
        \underset{q \in \queryFamily ~\text{s.t.}~ \Delta_{\query} \le \Delta}{\sup} \abs{\queryFunc{\distPost} - \queryFunc{\distDomain}} = \Delta \cdot \func{\mathbf{D}_{\text{TV}}}{\distPost \Vert \distDomain}
    \]
    and
    \[
        \underset{q \in \queryFamily ~\text{s.t.}~ \sigma_{\query}^{2} \le \sigma^{2}}{\sup} \abs{\queryFunc{\distPost} - \queryFunc{\distDomain}} = \sigma \sqrt{\chiSqr{\distPost}{\distDomain}},
    \]
    where $\BayesF{\element}{\view} \coloneqq \frac{\distFuncDep{\element}{\view}}{\distFunc{\element}} = \frac{\distFuncDep{\view}{\element}}{\distFunc{\view}}$ 
    is the Bayes factor of $\element$ given $\view$ (and vice-versa), $\mathbf{D}_{\text{TV}}$ is the total variation distance (Definition \ref{def:TVDist}), and $\mathbf{D}_{\chi^{2}}$ is the chi-square divergence (Definition \ref{def:chiSqrtDiv}).
\end{lemma}

\begin{proof}
    By definition,
    $\queryFunc{\distPost} = \expectation{\elementRV \sim \distPost}{\queryFunc{\elementRV}} = \expectation{\elementRV \sim \dist}{\BayesF{\elementRV}{\view} \queryFunc{\elementRV}}$, so
    \[
        \queryFunc{\distPost} - \queryFunc{\dist} = \expectation{\elementRV \sim \dist}{\BayesF{\elementRV}{\view} \queryFunc{\elementRV}} - \overset{= 1}{\overbrace{\expectation{\elementRV \sim \dist}{\BayesF{\elementRV}{\view}}}} \cdot \overset{= \queryFunc{\dist}}{\overbrace{\expectation{\elementRV \sim \dist}{\queryFunc{\elementRV}}}} = \cov{\elementRV \sim \dist}{\queryFunc{\elementRV}}{\BayesF{\elementRV}{\view}}.
    \]
    
    The second part is a direct result of the known variational representation of total variation distance and $\chi^{2}$ divergence, which are both $f$-divergences (see Equations 7.88 and 7.91 in \cite{PW22} for more details).
    \end{proof}

Using the first part of the lemma, we guarantee Bayes stability by bounding the correlation between specific $\query$ and $\BayesF{\cdot}{\view}$ as discussed in Section \ref{sec:disc}. The second part of this Lemma implies that bounding the appropriate divergence is necessary and sufficient for bounding the Bayes stability of the worst query in the corresponding family, which is how the main theorems of this paper are all achieved, using the next corollary.

\begin{corollary} \label{cor:BayesStabBnd}
    Given an analyst $\analyst$, for any $\epsilon > 0$ we have
    \[
        \prob{\genfrac{}{}{0pt}{}{\sampleSetRV \sim \iiDist}{\viewRV \sim \mechanismFunc{\sampleSetRV, \analyst}, \queryRV \sim \func{\analyst}{\viewRV}}}{\abs{\func{\queryRV}{\dist^{\viewRV}} - \func{\queryRV}{\dist}} > \sigma_{\queryRV} \cdot \epsilon} \le \prob{\genfrac{}{}{0pt}{}{\sampleSetRV \sim \iiDist}{\viewRV \sim \mechanismFunc{\sampleSetRV, \analyst}}}{\chiSqr{\dist_{\elementsDomain}^{\view}}{\dist_{\elementsDomain}} > \epsilon^{2}}.
    \]
\end{corollary}

The corresponding version of this corollary for bounded range queries provides an alternative proof to the generalization guarantees of the the LS stability notion \citep{LS19}.

\section{Pairwise concentration} \label{sec:PC}
In order to leverage Lemma \ref{lem:covStab}, we need a stability notion that implies Bayes stability of query responses in a manner that depends on the actual datasets and the actual queries (not just the worst case). In this section we propose such a notion and prove several key properties of it. Missing proofs from this section can be found in Appendix \ref{apd:PC}.

We start by introducing a measure of the stability loss of the mechanism.

\begin{definition} [Stability loss] \label{def:stabLoss}
    Given two sample sets $\sampleSet, \sampleSet' \in \domainOfSets$, a query $\query$, and a response $\response \in \responseFamily$, we denote by $\func{\ell_{\query}}{\sampleSet, \sampleSet'; \response} \coloneqq \lnFunc{\frac{\func{\distUpInd{}{\query}}{\response \,|\, \sampleSet}}{\func{\distUpInd{}{\query}}{\response \,|\, \sampleSet'}}}$ the \emph{stability loss} of $\response$ between $\sampleSet$ and $\sampleSet'$ for $\query$ (we omit $\query$ from notation for simplicity).\footnote{This quantity is sometimes referred to as \emph{privacy loss} in the context of differential privacy, or \emph{information density} in the context of information theory.} Notice that if $\responseRV \sim \mechanismFunc{\sampleSet, \query}$, the stability loss defines a random variable. 
    Similarly, given two elements $\element, \elementY \in \elementsDomain$, we denote $\loss{\element}{\elementY}{\response} \coloneqq \lnFunc{\frac{\distFuncDep{\response}{\element}}{\distFuncDep{\response}{\elementY}}}$ and $\lossS{\element}{\response} \coloneqq \lnFunc{\frac{\distFuncDep{\response}{\element}}{\distFunc{\response}}}$. This definition extends to views as well.
\end{definition}

Next we introduce a notion of a bound on the stability loss, where the bound is allowed to depend on the pair of swapped inputs and on the issued queries.

\begin{definition} [Pairwise concentration] \label{def:PC}
    Given a non-negative function $\PC: \domainOfSets \times \domainOfSets \times \queryFamily \rightarrow \pos$ which is symmetric in its first two arguments, a mechanism $\mechanism$ will be called \emph{$\PC$-Pairwise Concentrated} (or PC, for short), if for any two datasets $\sampleSet, \sampleSet' \in \domainOfSets$ and $\query \in \queryFamily$, for any $\alpha \ge 0$,\footnote{The requirement for the range $0 \le \alpha \le 1$ might appear somewhat counter-intuitive, and results from the fact that $\PC$ provides simultaneously a bound on the stability loss random variable's mean (the $-\left(\alpha-1\right)$ coefficient) and variance proxy (the $\left(\alpha-1\right)^{2}$ coefficient), as discusses in Definitions \ref{def:CDP}, \ref{def:zCDP}.}
    \[
        \expectation{\responseRV \sim \mechanismFunc{\sampleSet, \query}}{\expFunc{\left(\alpha - 1 \right) \loss{\sampleSet}{\sampleSet'}{\responseRV}}} \le \expFunc{\alpha \left(\alpha - 1 \right) \PCqFunc{\sampleSet}{\sampleSet'}}.
    \]
    
    Given a non-negative function $\PC: \domainOfSets \times \domainOfSets \times \viewFamily \rightarrow \pos$ which is symmetric in its first two arguments and an analyst $\analyst$, a mechanism $\mechanism$ will be called $\PC$-Pairwise Concentrated with respect to $\analyst$, if for any $\sampleSet, \sampleSet' \in \domainOfSets$ and $\alpha \ge 0$,
    \[
        \expectation{\viewRV \sim \mechanismFunc{\sampleSet, \analyst}}{\expFunc{\left(\alpha - 1 \right) \left(\loss{\sampleSet}{\sampleSet'}{\viewRV} - \alpha \PCfuncL{\sampleSet}{\sampleSet'}{\viewRV} \right)}} \le 1.
    \]
    
    We refer to such a function $\PC$ as a \emph{similarity function} over responses or views, respectively.
\end{definition}

The similarity function serves as a measure of the local sensitivity of the issued queries with respect to the replacement of the two datasets, by quantifying the extent to which they differ from each other with respect to the query $\query$. The case of noise addition mechanisms provides a natural intuitive interpretation, where the $\PC$ scales with the difference between $\queryFunc{s}$ and $\queryFunc{s'}$, which governs how much observing the response $r$ distinguishes between the two datasets, as stated in the next lemma.

We note that the first part of this definition can be viewed as a refined version of zCDP (Definition \ref{def:zCDP}), where the bound on the R\'{e}nyi divergence (Definition \ref{def:RenDiv}) is a function of the sample sets and the query. As for the second part, since the bound depends on the queries, which themselves are random variables, it should be viewed as a bound on the R\'{e}nyi dissimilarity notion that we introduce in the appendix (Definition \ref{def:dynRenyiDis}). This kind of extension is not limited to R\'{e}nyi divergence, as discussed in Appendix \ref{apd:divAndDis}.

\cpLem{Gaussian mechanism is PC}{GaussPC}{
    Given $\eta > 0$ and an analyst $\analyst$, if $\mechanism$ is a Gaussian mechanism with noise parameter $\eta$, then it is $\PC$-PC for $\PCqFunc{\sampleSet}{\sampleSet'} \coloneqq \frac{\left(\queryFunc{\sampleSet} - \queryFunc{\sampleSet'} \right)^{2}}{2 \eta^{2}}$ and $\PC$-PC with respect to $\analyst$ for $\PCvFunc{\sampleSet}{\sampleSet'} \coloneqq \frac{\norm{\func{\viewQuery}{\sampleSet} - \func{\viewQuery}{\sampleSet'}}^{2}}{2 \eta^{2}}$.
}

To leverage this stability notion in an adaptive setting, it must hold under adaptive composition, which we prove in the next theorem.
\cpThm{PC composition}{PCcomp}{
    Given $\numOfIterations \in \naturals$, a similarity function over responses $\PC$, and an analyst $\analyst$ issuing $\numOfIterations$ queries, if a mechanism $\mechanism$ is $\PC$-PC, then it is $\tilde{\PC}$-PC with respect to $\analyst$, where
    $\func{\tilde{\PC}}{\sampleSet, \sampleSet'; \view} \coloneqq \sum_{i=1}^{\numOfIterations} \PCfuncL{\sampleSet}{\sampleSet'}{\query_{i}}$.
}

Before we state the stability properties of PC mechanisms, we first transition to its element-wise version, which while less general than the original definition, is more suited for our use.

\cpLem{Element-wise PC}{PCelem}{
Given a similarity function $\PC$ and an analyst $\analyst$, if a mechanism $\mechanism$ that is $\PC$-PC with respect to $\analyst$ receives an iid sample from $\domainOfSets$, then for any two elements $\element, \elementY \in \elementsDomain$ and $\alpha \ge 1$ we have
    \[
        \expectation{\viewRV \sim \distFuncDep{\cdot}{\element}}{\expFunc{\left(\alpha - 1 \right) \left(\loss{\element}{\elementY}{\viewRV} - \alpha \PCfuncL{\element}{\elementY}{\viewRV} \right)}} \le 1
    \]
    and
    \[
        \expectation{\viewRV \sim \distFuncDep{\cdot}{\element}}{\expFunc{\left(\alpha - 1 \right) \left(\lossS{\element}{\viewRV} - \alpha \PCfuncS{\element}{\viewRV} \right)}} \le 1,
    \]
    where $\PCvFunc{\element}{\elementY} \coloneqq \supFunc{\sampleSet \in \elementsDomain^{\sampleSize - 1}}{\PCvFunc{\tuple{\sampleSet}{\element}}{\tuple{\sampleSet}{\elementY}}}$ and  $\PCfuncS{\element}{\view} \coloneqq \lnFunc{\expectation{\elementRVY \sim \dist}{e^{\PCvFunc{\element}{\elementRVY}}}}$.
}

Finally, we bound the Bayes stability of PC mechanisms.

\begin{theorem} [PC stability] \label{thm:PCstab}
    Given a similarity function over views $\PC$ and an analyst $A$, if a mechanism $\mechanism$ that is $\PC$-PC with respect to $\analyst$ receives an iid sample from $\domainOfSets$, then
    for any $\epsilon, \delta > 0$, $\phi \ge \expectation{\genfrac{}{}{0pt}{}{\elementRV \sim \dist, \sampleSetRV \sim \iiDist}{\viewRV \sim \mechanismFunc{\sampleSetRV, \analyst}}}{\PCfuncS{\elementRV}{\viewRV}}$, we have
    \begin{align*}
        \underset{\genfrac{}{}{0pt}{}{\sampleSetRV \sim \iiDist}{\viewRV \sim \mechanismFunc{\sampleSetRV, \analyst}, \queryRV \sim \func{\analyst}{\viewRV}}}{{\text{Pr}}} & \left[\abs{\func{\queryRV}{\dist_{\elementsDomain}^{\view}} - \func{\queryRV}{\dist_{\elementsDomain}}} > \sigma_{\queryRV} \epsilon \right]
        \\ & \le \prob{\genfrac{}{}{0pt}{}{\sampleSetRV \sim \iiDist}{\viewRV \sim \mechanismFunc{\sampleSetRV, \analyst}}}{\expectation{\elementRV \sim \dist}{e^{27 \lnFunc{\frac{1}{\delta}} \left(\PCfuncS{\elementRV}{\viewRV} + \phi \right)}} > 1 + \frac{\epsilon^{2}}{6}} + \func{O}{\frac{e^{\phi} \delta}{\phi \epsilon^{2}}},
    \end{align*}
    where $\PCfuncS{\element}{\view}$ is as defined in Lemma \ref{lem:PCelem}.
\end{theorem}

An exact version of the bound can be found in Theorem \ref{thm:PCstabExt}.

\begin{proof}[Proof outline]
    From Corollary \ref{cor:BayesStabBnd},
    \[
        \prob{\genfrac{}{}{0pt}{}{\sampleSetRV \sim \iiDist}{\viewRV \sim \mechanismFunc{\sampleSetRV, \analyst}, \queryRV \sim \func{\analyst}{\viewRV}}}{\abs{\func{\queryRV}{\dist_{\elementsDomain}^{\view}} - \func{\queryRV}{\dist_{\elementsDomain}}} > \sigma_{\queryRV} \epsilon} \le \prob{\genfrac{}{}{0pt}{}{\sampleSetRV \sim \iiDist}{\viewRV \sim \mechanismFunc{\sampleSetRV, \analyst}}}{\chiSqr{\dist_{\elementsDomain}^{\view}}{\dist_{\elementsDomain}} > \epsilon^{2}}.
    \]
    
    Denoting $\PIfuncS{\element}{\view} \coloneqq \PCfuncS{\element}{\view} + \phi + 4 \sqrt{\lnFunc{\frac{1}{\delta}} \left(\PCfuncS{\element}{\view} + \phi \right)}$, this can be bounded by
    \[
        \prob{\genfrac{}{}{0pt}{}{\sampleSetRV \sim \iiDist}{\viewRV \sim \mechanismFunc{\sampleSetRV, \analyst}}}{\chiSqr{\dist_{\elementsDomain}^{\view}}{\dist_{\elementsDomain}} > \expectation{\elementRV \sim \dist}{ \left(e^{\PIfuncS{\elementRV}{\viewRV}} - 1 \right)^{2}} + \frac{\epsilon^{2}}{2}} + \prob{\genfrac{}{}{0pt}{}{\sampleSetRV \sim \iiDist}{\viewRV \sim \mechanismFunc{\sampleSetRV, \analyst}}}{\expectation{\elementRV \sim \dist}{\left(e^{\PIfuncS{\elementRV}{\viewRV}} - 1 \right)^{2}} > \frac{\epsilon^{2}}{2}}.
    \]
    
    We then show via algebraic manipulation that this second term can be bounded using the inequality
    \[
        \expectation{\elementRV \sim \dist}{ \left(e^{\PIfuncS{\elementRV}{\view}} - 1 \right)^{2}} \le 3 \left(\expectation{\elementRV \sim \dist}{e^{27 \lnFunc{\frac{1}{\delta}} \left(\PCfuncS{\elementRV}{\view} + \phi \right)}} - 1 \right),
    \]
    and, recalling the definition of the chi square divergence between $\distP$ and $\distQ$ (Definition \ref{def:chiSqrtDiv}) $\chiSqr{\distP}{\distQ} \coloneqq \expectation{\elementRV \sim \dist_{2}}{\left(\frac{\func{\distP}{\elementRV}}{\func{\distQ}{\elementRV}} - 1 \right)^{2}}$, the first term can be bounded using Markov's inequality followed by the Cauchy-Schwarz inequality by the term: 
    \begin{align*}
        \frac{2}{\epsilon^{2}} \underset{\genfrac{}{}{0pt}{}{\elementRV \sim \dist, \sampleSetRV \sim \iiDist}{\viewRV \sim \mechanismFunc{\sampleSetRV, \analyst}}}{\mathbb{E}} & \left[\left(\frac{\distFuncDep{\viewRV}{\elementRV}}{\distFunc{\viewRV}} - 1 \right)^{2} \cdot \indicator{\func{B}{\PI}}{\tuple{\elementRV}{\viewRV}} \right]
        \\ & \le \frac{2}{\epsilon^{2}} \sqrt{\overset{*}{\overbrace{\left(\expectation{\genfrac{}{}{0pt}{}{\elementRV \sim \dist, \sampleSetRV \sim \iiDist}{\viewRV \sim \mechanismFunc{\sampleSetRV, \analyst}}}{\left(\frac{\distFuncDep{\viewRV}{\elementRV}}{\distFunc{\viewRV}} \right)^{4}} + 1 \right)}} \cdot \overset{**}{\overbrace{\prob{\genfrac{}{}{0pt}{}{\elementRV \sim \dist, \sampleSetRV \sim \iiDist}{\viewRV \sim \mechanismFunc{\sampleSetRV, \analyst}}}{\tuple{\elementRV}{\viewRV} \in \func{B}{\PI}}}}}
    \end{align*}
    where $\func{B}{\PI} \coloneqq \left\{\tuple{\element}{\view} \in \elementsDomain \times \viewFamily \,|\, \abs{\lnFunc{\frac{\distFuncDep{\view}{\element}}{\distFunc{\view}}}} > \PIfuncS{\element}{\view} \right\}$.
    
    We then bound the quantity * by $\func{O}{e^{\phi}}$, and we bound **  by $\func{O}{\frac{\delta^{2}}{\phi}}$, which completes the proof.
\end{proof}

Combining this theorem with Theorem \ref{thm:BayesStabImpdistPcc} we get that PC and sample accuracy together imply distribution accuracy.

\section{Variance-based generalization guarantees for the Gaussian mechanism} \label{sec:varAcc}
In this section we leverage the theorems of the previous sections to prove variance-based generalization guarantees for the Gaussian mechanism under adaptive data analysis. All proofs from this section appear in Appendix \ref{apd:varAcc}.

We first provide generalization guarantees for bounded queries.

\begin{theorem} [Generalization guarantees for bounded queries] \label{thm:GaussMechAccBnd}
    Given $\numOfIterations \in \naturals$; $\Delta, \sigma, \epsilon \ge 0$; $0 < \delta \le \frac{1}{e}$; and an analyst $\analyst$ issuing $\numOfIterations$ $\Delta$-bounded linear queries with variance bounded by $\sigma^{2}$, if $\mechanism$ is a Gaussian mechanism with noise parameter $\eta = \func{\Theta}{\sigma \sqrt{\frac{\sqrt{\numOfIterations}}{\sampleSize}}}$ that receives an iid dataset of size $\sampleSize = \func{\Omega}{\max \left\{\frac{\Delta}{\epsilon}, \frac{\sigma^{2}}{\epsilon^{2}} \right\} \sqrt{\numOfIterations} \cdot \lnFunc{\frac{\numOfIterations \sigma}{\delta \epsilon}}}$, 
    then $\mechanism$ is $\tuple{\epsilon}{\delta}$-distribution accurate.
    
    An exact version of the bound can be found in Theorem \ref{thm:GaussMechAccBndExt}.    
\end{theorem}

This theorem provides significant improvement over similar results that were achieved using differential privacy (see, e.g., Theorem 13 in~\citet{JLNRSS20}, which is defined for $\Delta = 1$), by managing to replace the $\Delta^{2}$ term with $\sigma^{2}$. The remaining dependence on $\Delta$ is inevitable, in the sense that such a dependence is needed even for non-adaptive analysis. This improvement resembles the improvement provided by Bernstein's inequality over Hoeffding's inequality.

This generalization guarantee, which nearly avoids dependence on the range of the queries, begs the question of whether it is possible to extend these results to handle unbounded queries. Clearly such a result would not be true without some bound on the tail distribution for a single query, so we focus in the next theorem on the case of sub-Gaussian queries. Formally, we will consider the case where $\queryFunc{\elementRV} - \queryFunc{\dist}$ is a sub-Gaussian random variable, for all queries.

\begin{definition}[Sub-Gaussian random variable] \label{def:subGauss}
    Given $\sigma > 0$, a random variable $X \in \reals$ will be called \emph{$\sigma^{2}$-sub-Gaussian} if for any $\lambda \in \reals$ we have $\expectation{}{e^{\lambda X}} \le e^{\frac{\lambda^{2}}{2} \nu}$.
\end{definition}

We can now state our result for unbounded queries.

\begin{theorem} [Generalization guarantees for sub-Gaussian queries] \label{thm:GaussMechAccSubGauss}
    Given $\numOfIterations \in \naturals$; $\sigma, \epsilon \ge 0$; $0 < \delta \le \frac{1}{e}$; and an analyst $\analyst$ issuing $\numOfIterations$ $\sigma^{2}$-sub-Gaussian linear queries, if $\mechanism$ is a Gaussian mechanism with noise parameter $\eta = \func{\Theta}{\sigma \sqrt{\frac{\sqrt{\numOfIterations}}{\sampleSize}}}$ that receives an iid dataset of size $\sampleSize = \func{\Omega}{\frac{\sigma^{2}}{\epsilon^{2}} \sqrt{\numOfIterations} \lnFunc{\frac{\numOfIterations \sigma}{\delta \epsilon}}}$, 
    then $\mechanism$ is $\tuple{\epsilon}{\delta}$-distribution accurate.
    
    An exact version of the bound can be found in Theorem \ref{thm:GaussMechAccSubGaussExt}.
\end{theorem}

These results extend to the case where the variance (or variance proxy) of each query $\query_{i}$ is bounded by a unique value $\sigma_{i}^{2}$, by simply passing this value to the mechanism as auxiliary information and scaling the added noise $\eta_{i}$ accordingly. Furthermore, using this approach we can quantify the extent to which incorrect bounds affect the accuracy guarantee. Overestimating the bound on a query's variance would increase the error of the response to this query by a factor of square root of the ratio between the assumed and the correct variance, while the error of the other responses would only decrease. On the other hand, underestimating the bound on a query's variance would only decrease the error of the response to this query, while increasing the error of each subsequent query by a factor of the square root of the ratio between the assumed and the correct variance, divided by the number of subsequent queries. A formal version of this claim can be found in Section \ref{sec:difIncoVar}.

\section{Discussion} \label{sec:disc}
The contribution of this paper is two-fold. In Section \ref{sec:gen}, we provide a tight measure of the level of overfitting of some query with respect to previous responses. In Sections \ref{sec:PC} and \ref{sec:varAcc}, we demonstrate a toolkit to utilize this measure, and use it to prove new generalization properties of fundamental noise-addition mechanisms. The novelty of the PC definition stems replacing the fixed parameters that appear in the differential privacy definition with a function of the datasets and the query. The definition presented in this paper provides a generalization of zero-concentrated differential privacy, and future work could study similar generalizations of other privacy notions, as discussed in Section \ref{sec:stabMeas}.

One small extension of the present work would be to consider queries with range $\reals^{d}$. It would also be interesting to extend our results to handle arbitrary normed spaces, using appropriate noise such as perhaps the Laplace mechanism. It might also be possible to relax our assumption that data elements are drawn iid to a weaker independence requirement. Furthermore, it would be interesting to explore an extension from linear queries to general low-sensitivity queries.

We hope that the mathematical toolkit that we establish in Appendix~\ref{apd:divAndDis} to analyze our stability notion may find additional applications, perhaps also in context of privacy accounting. Furthermore, the max divergence can be generalized analogously to the ``dynamic'' generalization of R\'{e}nyi divergence proposed in this paper (\ref{def:dynRenyiDis}), perhaps suggesting that this approach may be useful in analyzing other mechanisms as well.

Our Covariance Lemma (\ref{lem:covStab}) shows that there are two possible ways to avoid adaptivity-driven overfitting---by bounding the Bayes factor term, which induces a bound on $\abs{\queryFunc{\distPost} - \queryFunc{\dist}}$, as we do in this work, or by bounding the correlation between $\query$ and $\BayesF{\cdot}{\view}$. This second option suggests interesting directions for future work. For example, to capture an analyst that is non-worst-case in the sense that she ``forgets'' some of the information that she has learned about the dataset, both the posterior accuracy and the Bayes stability could be redefined with respect to the internal state of the analyst instead of with respect to the full view. This could allow for improved bounds in the style of~\citet{HZ19}.

\paragraph{Acknowledgements}
We gratefully acknowledge productive discussions with Etam Benger, Vitaly Feldman, Yosef Rinott, Aaron Roth, and Tomer Shoham. 
This work was supported in part by a gift to the McCourt School of Public Policy and Georgetown University, Simons Foundation Collaboration 733792, Israel Science Foundation (ISF) grant 2861/20, and a grant from the Israeli Council for Higher Education. Shenfeld's work was also partly supported by the Apple Scholars in AI/ML PhD Fellowship. Part this work was completed while Ligett was visiting Princeton University's Center for Information Technology Policy.

\bibliographystyle{plainnat}
\bibliography{bibliography}

\appendix

\newpage
\section{Formal definitions} \label{apd:formDefs}

\subsection{Distributions of interest}

\begin{definition} [Distributions over $\domainOfSets$ and $\responseFamily$] \label{def:distsOfSets}
    A distribution $\distDomainOfSets$, a query $\query$, and a mechanism $\mechanism : \domainOfSets \times \queryFamily \rightarrow \responseFamily$, together induce a set of distributions over $\domainOfSets$, $\responseFamily$, and $\domainOfSets \times \responseFamily$.
    
    The \emph{conditional distribution} $\distDep{\responseFamily}{\domainOfSets}{\query}$ over $\responseFamily$ represents the probability to get $\response$ as the output of $\mechanismFunc{\sampleSet, \query}$. That is, $\forall \sampleSet \in \domainOfSets, \response\in\responseFamily$, if $\responseFamily$ is countable,
    \[
        \func{\distDep{\responseFamily}{\domainOfSets}{\query}}{\response\,|\,\sampleSet} \coloneqq \prob{\responseRV \sim \mechanismFunc{\sampleSet, \query}}{\responseRV = \response \,|\, \sampleSet},
    \]
    where the probability is taken over the internal randomness of $\mechanism$. In case $\responseFamily$ is a measurable spaces, it represents the probability density function (the Radon–Nikodym derivative).
    
    The \emph{joint distribution} $\distJoint{\domainOfSets}{\responseFamily}{\query}$ over $\domainOfSets \times \responseFamily$ represents the probability to sample a particular $\sampleSet$ and get $\response$ as the output of $\mechanismFunc{\sampleSet, \query}$. That is, $\forall \sampleSet \in \domainOfSets, \response \in \responseFamily$,
    \[
        \func{\distJoint{\domainOfSets}{\responseFamily}{\query}}{\sampleSet, \response} \coloneqq \func{\distDomainOfSets}{\sampleSet} \cdot \func{\distDep{\responseFamily}{\domainOfSets}{\query}}{\response \,|\, \sampleSet}.
    \]
    
    The \emph{marginal distribution} $\distUpInd{\responseFamily}{\query}$ over $\responseFamily$ represents the prior probability to get output $\response$ without any knowledge of $\sampleSet$. That is, $\forall \response \in \responseFamily,$
    \[
        \func{\distUpInd{\responseFamily}{\query}}{\response} \coloneqq
        \int_{\sampleSet \in \domainOfSets} \func{\distJoint{\domainOfSets}{\responseFamily}{\query}}{\sampleSet, \response} d \response.
    \]
    
    The \emph{conditional distribution} $\distDep{\domainOfSets}{\responseFamily}{\query}$ over $\domainOfSets$ represents the posterior probability that the input dataset to $\mechanism$ was $\sampleSet$, given that $\mechanismFunc{\cdot, \query}$ returns $\response$. That is, $\forall \sampleSet \in \domainOfSets, \response\in\responseFamily,$
    \[
        \func{\distDep{\domainOfSets}{\responseFamily}{\query}}{\sampleSet \,|\, \response} \coloneqq \frac{\func{\distJoint{\domainOfSets}{\responseFamily}{\query}}{\sampleSet, \response}}{\func{\distUpInd{\responseFamily}{\query}}{\response}}.
    \]
\end{definition}

    \begin{definition} [Distributions over $\elementsDomain$ and $\responseFamily$] \label{def:distsOfDomain}
    The \emph{marginal distribution} $\distDomain$ over $\elementsDomain$ represents the probability to get $\element$ by sampling a dataset and then sampling one element from that dataset uniformly at random. That is, $\forall \element \in \elementsDomain,$
    \[
        \func{\distDomain}{\element} \coloneqq \int_{\sampleSet \in \domainOfSets} \func{\distDomainOfSets}{\sampleSet} \cdot \func{\distDep{\elementsDomain}{\domainOfSets}{}}{\element \,|\, \sampleSet} d \sampleSet,
    \]
    where $\func{\distDep{\elementsDomain}{\domainOfSets}{}}{\element \,|\, \sampleSet}$ denotes the probability to get $\element$ by sampling $\sampleSet$ uniformly at random.
    
    The \emph{joint distribution} $\distJoint{\elementsDomain}{\responseFamily}{\query}$ over $\elementsDomain\times\responseFamily$ represents the probability to get $\element$ by sampling a dataset uniformly at random and also get $\response$ as the output of $\mechanismFunc{\cdot, \query}$ from the same dataset. That is, $\forall \element\in\elementsDomain,\response\in\responseFamily,$
    \[
        \func{\distJoint{\elementsDomain}{\responseFamily}{\query}}{\element,\response} \coloneqq \int_{\sampleSet \in \domainOfSets} \func{\distDomainOfSets}{\sampleSet} \cdot \func{\distDep{\elementsDomain}{\domainOfSets}{}}{\element \,|\, \sampleSet} \cdot \func{\distDep{\responseFamily}{\domainOfSets}{\query}}{\response \,|\, \sampleSet} d \sampleSet.
    \]
    where $\func{\distDep{\elementsDomain}{\domainOfSets}{}}{\element \,|\, \sampleSet}$ denotes the probability to get $\element$ by sampling $\sampleSet$ uniformly at random.
    
    The \emph{conditional distribution} $\distDep{\elementsDomain}{\responseFamily}{\query}$ over $\elementsDomain$ represents the probability to get $\element$ by sampling a dataset uniformly at random, given the fact that we got $\response$ as the output of $\mechanismFunc{\cdot, \query}$ from that dataset. That is, $\forall \element \in \elementsDomain, \response\in\responseFamily,$
    \[
        \func{\distUpInd{}{\response}}{\element} \coloneqq \func{\distDep{\elementsDomain}{\responseFamily}{\query}}{\element \,|\, \response} \coloneqq \int_{\sampleSet \in \domainOfSets} \func{\distDep{\domainOfSets}{\responseFamily}{\query}}{\sampleSet \,|\, \response} \cdot \func{\distDep{\elementsDomain}{\domainOfSets}{}}{\element\,|\,\sampleSet} d \sampleSet.
    \]
\end{definition}

Although all of these definitions depend on $\distDomainOfSets$ and $\mechanism$, we typically omit these from the notation for simplicity, and usually omit the superscripts and subscripts entirely. We include them only when necessary for clarity.

Though the conditional distributions $\distDep{\responseFamily}{\elementsDomain}{\query}$ and $\distDep{\elementsDomain}{\responseFamily}{\query}$ were not defined as the ratio between the joint and marginal distribution, the analogue of Bayes' rule still holds for these distributions. A formal proof of this claim can be found in Appendix A in~\citet{LS19}.

\subsection{A Formal treatment of adaptivity} \label{sec:formAdap}

\begin{definition}[Adaptive mechanism] \label{def:adap}
    Illustrated below, the \emph{adaptive mechanism} $\text{Adp}_{\mechanism}: \domainOfSets \times \analystsFamily \rightarrow \viewFamily_{\numOfIterations}$ is a particular type of mechanism, which inputs an analyst as its query and which returns a view as its response and is parametrized by a mechanism $\mechanism$ and number of iterations $\numOfIterations$.
    Given a dataset $\sampleSet$ and an analyst $\analyst$ as input, the adaptive mechanism iterates $\numOfIterations$ times through the process where $\analyst$ sends a query and auxiliary information to $\mechanism$ and receives its response to that query on the dataset. The adaptive mechanism returns the resulting sequence of $\numOfIterations$ responses $\view_{\numOfIterations}$. Naturally, this requires $\analyst$ to match $\mechanism$ such that $\mechanism$'s output can be $\analyst$'s input, and vice versa. 
    
    If $\analyst$ is randomized, we add one more step at the beginning where $\text{Adp}_{\mechanism}$ randomly generates some bits $c$---$\analyst$'s ``coin tosses.'' In this case, $v_{\numOfIterations} \coloneqq \left( c, \response_{1}, \ldots, \response_{i\numOfIterations} \right)$ and $\analyst$ receives the coin tosses as an input as well. This addition turns $\query_{\numOfIterations + 1}$ and $\theta_{\numOfIterations + 1}$ into a deterministic function of $\view_{i}$ for any $i \in \naturals$, a fact that we later rely on. In this situation, the randomness of $\text{Adp}_{\mechanism}$ results both from the randomness of the coin tosses and from that of $\mechanism$. We will denote $\mechanismFunc{\sampleSet, \analyst} \coloneqq \func{\text{Adp}_{\mechanism}}{\sampleSet, \analyst}$ for simplicity.
    
    \begin{figure}[h]
         \centering
        \fbox{
             \begin{minipage}{\linewidth - 15pt}
                Adaptive mechanism $\text{Adp}_{\mechanism}$
                \hrule
                
                \textbf{Input:} $\sampleSet \in \domainOfSets, \, \analyst \in \analystsFamily$
                
                \textbf{Output:} $\view_{\numOfIterations} \in \viewFamily_{\numOfIterations}$
                
                $\view_{0} \gets \emptyset$ or $c$
                
                \textbf{for} $i \in \left[ \numOfIterations \right]$ :
                
                ~~~$\tuple{\query_{i}}{\theta_{i}} \gets \func{\analyst}{\view_{i-1}}$
                
                ~~~$\response_{i} \gets \mechanismFunc{\sampleSet, \query_{i}, \theta_{i}}$
                
                ~~~$\view_{i} \gets \tuple{\view_{i-1}}{\response_{i}}$
                
                \textbf{return} $\view_{\numOfIterations}$
             \end{minipage}
        }
    \end{figure}
    
    Using the adaptive mechanism we can extend the set of distributions presented in Definitions \ref{def:distsOfSets} and \ref{def:distsOfDomain}, to distributions over $\viewFamily$ and $\elementsDomain$, with $\viewFamily$ taking the place of $\responseFamily$ and $\analyst$ replacing $\query$.
\end{definition}

\newpage
\section{Divergence, dissimilarity, and stability measures} \label{apd:divAndDis}
Given two probability distributions $\distP, \distQ$ over some domain $\Omega$, there are many useful measures of their dissimilarity. In this section we develop what is as far as we know a new type of dissimilarity measure between distributions, which generalizes the notion of divergences. We start by recalling several common divergence measures in Section \ref{sec:divMeas}. Next, Section \ref{sec:DynMeas}, we present in a generalization of the divergence notion, where a scalar is replaced by a function, yielding dynamic measures which we refer to as dissimilarities. We then prove (Section \ref{sec:conv}) several convexity properties for max and R\'{e}nyi dissimilarities. Finally (Section \ref{sec:stabMeas}) we propose several novel stability measures which build on the proposed dissimilarity measures and are used throughout the paper.

\subsection{Divergence measures} \label{sec:divMeas}
Dissimilarity between distributions is typically measured using a divergence, that is, a function receiving two distributions as input, and outputting a non-negative value which is equal to $0$ if and only if the two distributions are identical almost surely. One commonly used family of divergences is the $f$-divergences, defined as $\func{\mathbf{D}_{f}}{\distP \Vert \distQ} \coloneqq \expectation{\omega \sim \distQ}{\func{f}{\frac{\distPfunc{\omega}} {\distQfunc{\omega}}}}$. Here we recall several instances which are used in this paper.

\begin{definition}[Total variation distance] \label{def:TVDist}
The \emph{total variation distance} (also known as \emph{statistical distance}) between $\distP$ and $\distQ$ is defined as
\[
\func{\mathbf{D}_{\text{TV}}}{\distP \Vert \distQ} \coloneqq 
\frac{1}{2} \expectation{\omega \sim \distQ}{\abs{\frac{\distPfunc{\omega}}{\distQfunc{\omega}} - 1}}.
\]
\end{definition}

\begin{definition}[Chi-square divergence] \label{def:chiSqrtDiv}
The \emph{Chi-square divergence} (also known as Pearson's $\chi^{2}$ divergence) between $\distP$ and $\distQ$ is defined as
\[
\chiSqr{\distP}{\distQ} \coloneqq 
\expectation{\omega \sim \distQ}{\left(\frac{\distPfunc{\omega}}{\distQfunc{\omega}} - 1 \right)^{2}}.
\]
\end{definition}

\begin{definition}[KL divergence] \label{def:KLDiv}
The \emph{Kullback–Leibler divergence} (or \emph{KL divergence}) between $\distP$ and $\distQ$ is defined as
\[
\func{\mathbf{D}_{KL}}{\distP \Vert \distQ} \coloneqq \expectation{\omega \sim \distP}{\lnFunc{\frac{\distPfunc{\omega}} {\distQfunc{\omega}}}} = \expectation{\omega \sim \distQ}{\frac{\distPfunc{\omega}} {\distQfunc{\omega}} \cdot \lnFunc{\frac{\distPfunc{\omega}} {\distQfunc{\omega}}}}.
\]
\end{definition}

Not all divergences are $f$-divergences. For example:
\begin{definition}[($\delta$-Approximate) max divergence] \label{def:maxDiv}
Given $\delta \ge 0$, the \emph{$\delta$-approximate max divergence} between $\distP$ and $\distQ$ is defined as
\[
\func{\mathbf{D}_{\infty}^{\delta}}{\distP \Vert \distQ} \coloneqq \underset{\boldsymbol{\omega} \subseteq \Omega \,|\, \distPfunc{\boldsymbol{\omega}} \ge \delta}{\sup} \left(\lnFunc{\frac{\distPfunc{\omega} - \delta}{\distQfunc{\omega}}} \right).
\]
The case of $\delta = 0$ is simply called the max divergence and is denoted by $\mathbf{D}_{\infty}$.
\end{definition}

These two last definitions are generalized by the following one:
\begin{definition} [R\'{e}nyi divergence] \label{def:RenDiv}
Given $\alpha \in \left[0, \infty \right]$, the \emph{R\'{e}nyi divergence} between two probability distributions $\distP$ and $\distQ$ is defined as
\[
\func{\mathbf{D}_{\alpha}}{\distP \Vert \distQ} \coloneqq \frac{1}{\alpha - 1} \lnFunc{\expectation{\omega \sim \distP}{\left(\frac{\distPfunc{\omega}} {\distQfunc{\omega}} \right)^{\alpha-1}}}.
\]

The cases of $\alpha = 0, 1$, and $\infty$ are defined using the limit and converge to $-\lnFunc{\prob{\omega \sim \distQ}{\omega > 0}}$, the KL divergence, and the max divergence, respectively.\footnote{This definition generalizes the original one introduced by Alfr\'{e}d R\'{e}nyi to include the range $\left[0, 1 \right)$, which will become useful later in Theorem \ref{thm:dynRenImpDynMax}. For more details, see~\citet{VT14}}.
\end{definition}

\subsection{Dissimilarity measures} \label{sec:DynMeas}
In this section we present a generalization of the R\'{e}nyi divergence and the max divergence. We start by recalling two useful identities that are a direct result of the definitions.

\begin{fact}
Given $\epsilon, \delta \ge 0$ and two probability distributions $\distP, \distQ$, we have $\func{\mathbf{D}_{\infty}^{\delta}}{\distP \Vert \distQ} \le \epsilon$ if and only if for any $\boldsymbol{\omega} \subseteq \Omega$, $\prob{\omega \sim \distP}{\omega \in \boldsymbol{\omega}} \le e^{\epsilon} \prob{\omega \sim \distQ}{\omega \in \boldsymbol{\omega}} + \delta$.
\end{fact}

\begin{fact}
Given $\phi \ge 0$, $\alpha \ge 1$, and two probability distributions $\distP, \distQ$, we have $\func{\mathbf{D}_{\alpha}}{\distP \Vert \distQ} \le \phi$ if and only if \[
\expectation{\omega \sim \distP}{\expFunc{\left(\alpha - 1 \right) \lnFunc{\frac{\distPfunc{\omega}}{\distQfunc{\omega}}}}} \le e^{\left(\alpha - 1 \right) \phi}.
\]
\end{fact}

Motivated by these two identities, we generalize these divergences by replacing the scalar bounding the divergence by a function, leading to a more ``dynamic'' measure that can no longer be regarded a divergence, hence the names. 

We first generalize the notion of max divergence.
\begin{definition}[Max dissimilarity] \label{def:dynMaxDis}
Given $\delta \ge 0$, and a non-negative function $\varepsilon: \Omega \rightarrow \pos$, we say the \emph{($\delta$-approximate) max dissimilarity} between $\distP$ and $\distQ$ is bounded by $\varepsilon$ if for any $\boldsymbol{\omega} \subseteq \Omega$, 
\[
\int_{\omega \in \boldsymbol{\omega}} \distPfunc{\omega} d\omega \le \int_{\omega \in \boldsymbol{\omega}} e^{\func{\varepsilon}{\omega}} \distQfunc{\omega} d\omega + \delta.
\]

We denote this fact by $\distP \DPMD{}{} \distQ$.

Similarly, we denote $\distP \DNMD{}{} \distQ$ if for any $\boldsymbol{\omega} \subseteq \Omega$, 
\[
\int_{\omega \in \boldsymbol{\omega}} e^{-\func{\varepsilon}{\omega}} \distPfunc{\omega} d\omega \le \int_{\omega \in \boldsymbol{\omega}} \distQfunc{\omega} d\omega + \delta.
\]
\end{definition}

Next we generalize the notion of R\'{e}nyi divergence. This dissimilarity measure serves as the mathematical basis for the new stability measure that we introduce (Definition \ref{def:PC}).

\begin{definition}[R\'{e}nyi dissimilarity] \label{def:dynRenyiDis}
Given $\alpha \ge 0$ and a non-negative function $\varphi: \Omega \rightarrow \pos$, we say the \emph{$\alpha$-R\'{e}nyi dissimilarity} between $\distP$ and $\distQ$ is bounded by $\varphi$ if
\[
\expectation{\omega \sim \distP}{\expFunc{\left(\alpha - 1 \right) \left( \lnFunc{\frac{\distPfunc{\omega}}{\distQfunc{\omega}}} - \func{\varphi}{\omega} \right)}} \le 1.
\]
\end{definition}
The following theorem states several implications between the various max dissimilarity definitions.

\begin{theorem} \label{thm:maxDisImp}
Given $\delta \ge 0$, a non-negative function $\varepsilon : \Omega \rightarrow \pos$, and two distributions $\distP, \distQ$:

\begin{enumerate}
\item $\prob{\omega \sim \distP}{\lnFunc{\frac{\distPfunc{\omega}}{\distQfunc{\omega}}} > \func{\varepsilon}{\omega}} \le \delta$ implies $\distP \DPMD{}{} \distQ$,
\item $\distP \DPMD{}{} \distQ$ implies $\distP \DNMD{}{} \distQ$,
\item $\distP \DPMD{}{} \distQ$ implies $\prob{\omega \sim \distP}{\lnFunc{\frac{\distPfunc{\omega}}{\distQfunc{\omega}}} > \func{\varepsilon}{\omega} + a} < \frac{e^{a} \delta}{e^{a} - 1}$ 
for any $a>0$, and
\item $\distQ \DNMD{}{} \distP$ implies $\prob{\omega \sim \distP}{\lnFunc{\frac{\distQfunc{\omega}}{\distPfunc{\omega}}} > \func{\varepsilon}{\omega} + a} < \frac{\delta}{e^{a} - 1}$ 
for any $a>0$.
\end{enumerate}
\end{theorem}

\begin{proof}
\textbf{Part 1}: Denoting $\func{B_{\varepsilon}}{\distP, \distQ} \coloneqq \left\{ \omega \in \Omega \,|\, \lnFunc{\frac{\distPfunc{\omega}}{\distQfunc{\omega}}} > \func{\varepsilon}{\omega} \right\}$
we have

\begin{align*}
\int_{\omega \in \boldsymbol{\omega}} \distPfunc{\omega} d\omega & = \int_{\omega \in \boldsymbol{\omega} \backslash \func{B_{\varepsilon}}{\distP, \distQ}} \distPfunc{\omega} d\omega + \int_{\omega \in \boldsymbol{\omega} \cap \func{B_{\varepsilon}}{\distP, \distQ}} \distPfunc{\omega} d\omega \\
& \eqRes{a}{\le} \int_{\omega \in \boldsymbol{\omega} \backslash \func{B_{\varepsilon}}{\distP, \distQ}} e^{\func{\varepsilon}{\omega}} \distQfunc{\omega} d\omega + \int_{\omega \in \func{B_{\varepsilon}}{\distP, \distQ}} \distPfunc{\omega} d\omega \\
& = \int_{\omega \in \boldsymbol{\omega} \backslash \func{B_{\varepsilon}}{\distP, \distQ}} e^{\func{\varepsilon}{\omega}} \distQfunc{\omega} d\omega + \prob{\omega \sim \distP}{\lnFunc{\frac{\distPfunc{\omega}}{\distQfunc{\omega}}} > \func{\varepsilon}{\omega}} \\
& \eqRes{b}{\le} \int_{\omega \in \boldsymbol{\omega}} e^{\func{\varepsilon}{\omega}} \distQfunc{\omega} d\omega + \delta,
\end{align*}
where (a) results from the definition of $\func{B_{\varepsilon}}{\distP, \distQ}$ and (b) from the assumption that $\prob{\omega \sim \distP}{\lnFunc{\frac{\distPfunc{\omega}}{\distQfunc{\omega}}} > \func{\varepsilon}{\omega}} \le \delta$.

\textbf{Part 2}: Denoting $\func{\delta}{\omega} \coloneqq \max \left\{\distPfunc{\omega} -e^{\func{\varepsilon}{\omega}} \distQfunc{\omega}, 0 \right\}$,
which implies $\distPfunc{\omega} \le e^{\func{\varepsilon}{\omega}} \distQfunc{\omega} + \func{\delta}{\omega}$, we have

\begin{align*}
\int_{\omega \in \boldsymbol{\omega}} e^{-\func{\varepsilon}{\omega}} \distPfunc{\omega} d\omega & \eqRes{a}{\le} \int_{\omega \in \boldsymbol{\omega}} e^{-\func{\varepsilon}{\omega}} \left(e^{\func{\varepsilon}{\omega}} \distQfunc{\omega} + \func{\delta}{\omega} \right) d\omega \\
& = \int_{\omega \in \boldsymbol{\omega}} \left(\distQfunc{\omega} + e^{-\func{\varepsilon}{\omega}} \cdot \func{\delta}{\omega} \right) d\omega \\
& \eqRes{b}{\le} \int_{\omega \in \boldsymbol{\omega}} \distQfunc{\omega} d\omega + \int_{\omega \in \boldsymbol{\omega}} \func{\delta}{\omega} d\omega \\
& \eqRes{c}{\le} \int_{\omega \in \boldsymbol{\omega}} \distQfunc{\omega} d\omega + \delta
\end{align*}

where (a) results from the fact the inequality holds for any element
$\omega \in \Omega$, (b) from the non-negativity of $\varepsilon$, and (c) from the assumption that $\distP \DPMD{}{} \distQ$.

\textbf{Part 3}: Denoting $\epsilon_{\distQ}^{ + } \coloneqq \lnFunc{\expectation{\omega \sim \distQ}{e^{\varepsilon\left(X\right)}}}$
and $\func{\distQ_{\distP}^{+}}{\omega}
\coloneqq e^{\func{\varepsilon}{\omega} -\epsilon_{\distQ}^{ + }} \distQfunc{\omega}$,
we notice that from the definition $\distQ_{\distP}^{+}$ is a valid distribution,
since 
\[
\int_{\omega \in \Omega} \func{\distQ_{\distP}^{+}}{\omega} d\omega = e^{-\epsilon_{\distQ}^{ + }} \int_{\omega \in \Omega} e^{\func{\varepsilon}{\omega}} \distQfunc{\omega} d\omega = \frac{1}{\expectation{\omega \sim \distQ}{e^{\varepsilon\left(X\right)}}} \expectation{\omega \sim \distQ}{e^{\varepsilon\left(X\right)}} = 1.
\]

We first transform the $\distP \DPMD{}{} \distQ$
bound into a bound on the max divergence between $\distP$ and $\distQ_{\distP}^{+}$.
\[
\prob{\omega \sim \distP}{\omega \in \boldsymbol{\omega}}
\eqRes{a}{\le} \int_{\omega \in \boldsymbol{\omega}} e^{\func{\varepsilon}{\omega}} \distQfunc{\omega} d\omega + \delta
\eqRes{b}{=} e^{\epsilon_{\distQ}^{ + }} \int_{\omega \in \boldsymbol{\omega}} \func{\distQ_{\distP}^{+}}{\omega} d\omega + \delta
= e^{\epsilon_{\distQ}^{ + }} \prob{\omega \sim \distQ_{\distP}^{+}}{\omega \in \boldsymbol{\omega}} + \delta,
\]
where (a) results from the assumption that $\distP \DPMD{}{} \distQ$ and (b) from the definition of $\func{\distQ_{\distP}^{+}}{\omega}$.

Next we denote $\func{B_{\varepsilon}^{a}}{\distP, \distQ} \coloneqq \left\{ \omega \in \Omega \,|\, \lnFunc{\frac{\distPfunc{\omega}}{\distQfunc{\omega}}} > \func{\varepsilon}{\omega} + a \right\}$,
and notice that for any $\omega \in \func{B_{\varepsilon}^{a}}{\distP, \distQ}$
we have $\distPfunc{\omega}>e^{\func{\varepsilon}{\omega} + a} \distQfunc{\omega}=e^{\epsilon_{\distQ}^{ + } + a} \func{\distQ_{\distP}^{+}}{\omega}$,
so

\begin{align*}
\prob{\omega \sim \distP}{\omega \in \func{B_{\varepsilon}^{a}}{\distP, \distQ}} & \eqRes{a}{\le} e^{\epsilon_{\distQ}^{ + }} \prob{\omega \sim \distQ_{\distP}^{+}}{\omega \in \func{B_{\varepsilon}^{a}}{\distP, \distQ}} + \delta \\
& = e^{\epsilon_{\distQ}^{ + }} \int_{\omega \in \func{B_{\varepsilon}^{a}}{\distP, \distQ}} \func{\distQ_{\distP}^{+}}{\omega} d\omega + \delta \\
& \eqRes{b}{<} e^{\epsilon_{\distQ}^{ + }} \int_{\omega \in \func{B_{\varepsilon}^{a}}{\distP, \distQ}} e^{- \left(\epsilon_{\distQ}^{ + } + a \right)} \distPfunc{\omega} d\omega + \delta \\
& = e^{-a} \prob{\omega \sim \distP}{\omega \in \func{B_{\varepsilon}^{a}}{\distP, \distQ}} + \delta
\end{align*}
where (a) results from the previous bound and (b) from the fact the inequality holds for any element $\omega \in \Omega$.

Reordering the terms we get
\begin{align*}
\prob{\omega \sim \distQ}{\lnFunc{\frac{\distPfunc{\omega}}{\distQfunc{\omega}}} > \func{\varepsilon}{\omega} + a} = \prob{\omega \sim \distP}{\omega \in \func{B_{\varepsilon}^{a}}{\distP, \distQ}} < \frac{e^{a} \delta}{e^{a} - 1}.
\end{align*}

\textbf{Part 4}: Denoting $\epsilon_{\distQ}^{-} \coloneqq -\lnFunc{\expectation{\omega \sim \distQ}{e^{-\func{\varepsilon}{\omega}}}}$ and $\func{\distQ_{\distP}^{-}}{\omega} \coloneqq e^{-\func{\varepsilon}{\omega} + \epsilon_{\distQ}^{-}} \distQfunc{\omega}$, we notice that from the definition $\distQ_{\distP}^{-} $ is a valid distribution, since 
\[
\int_{\omega \in \Omega} \func{\distQ_{\distP}^{-}}{\omega} d\omega = e^{\epsilon_{\distQ}^{-}} \int_{\omega \in \Omega} e^{-\func{\varepsilon}{\omega}} \distQfunc{\omega} d\omega = \frac{1}{\expectation{\omega \sim \distQ}{e^{-\func{\varepsilon}{\omega}}}} \expectation{\omega \sim \distQ}{e^{-\func{\varepsilon}{\omega}}} = 1.
\]

We first transform the $\distQ \DNMD{}{} \distP$
bound into a bound on the max divergence between $\distQ_{\distP}^{-}$ and
$\distP$.
\[
\prob{\omega \sim \distQ_{\distP}^{-}}{\omega \in \boldsymbol{\omega}} \eqRes{a}{=} e^{\epsilon_{\distQ}^{-}} \int_{\omega \in \boldsymbol{\omega}} e^{-\func{\varepsilon}{\omega}} \distQfunc{\omega} d\omega \eqRes{b}{\le} e^{\epsilon_{\distQ}^{-}}\left(\int_{\omega \in \boldsymbol{\omega}} \distPfunc{\omega} + \delta\right) = e^{\epsilon_{\distQ}^{-}} \prob{\omega \sim \distP}{\omega \in \boldsymbol{\omega}} + e^{\epsilon_{\distQ}^{-}}\delta
\]

where (a) results from the definition of $\distQ_{\distP}^{-} $ and (b) from the assumption that $\distQ \DNMD{}{} \distP$.

Next we denote $\func{B_{\varepsilon}^{a}}{\distQ, \distP} \coloneqq \left\{ \omega \in \Omega \,|\, \lnFunc{\frac{\distQfunc{\omega}}{\distPfunc{\omega}}} > \func{\varepsilon}{\omega} + a \right\}$, and notice that for any $\omega \in \func{B_{\varepsilon}^{a}}{\distQ, \distP}$
we have $\distPfunc{\omega} < e^{-\left(\func{\varepsilon}{\omega} + a \right)} \distQfunc{\omega} = e^{-\left(\epsilon_{\distQ}^{-} + a \right)} \func{\distQ_{\distP}^{-}}{\omega}$,
so

\begin{align*}
\prob{\omega \sim \distP}{\omega \in \func{B_{\varepsilon}^{a}}{\distQ, \distP}} & = \int_{\omega\in \func{B_{\varepsilon}^{a}}{\distQ, \distP}} \distPfunc{\omega} d\omega \\
& \eqRes{a}{<} \int_{\omega\in \func{B_{\varepsilon}^{a}}{\distQ, \distP}} e^{-\left(\epsilon_{\distQ}^{-} + a\right)}\func{\distQ_{\distP}^{-}}{\omega} d\omega \\
& = e^{-\left(\epsilon_{\distQ}^{-} + a\right)} \prob{\omega \sim \distQ_{\distP}^{-}}{\omega \in \func{B_{\varepsilon}^{a}}{\distQ, \distP}} \\
& \eqRes{b}{\le} e^{-\left(\epsilon_{\distQ}^{-} + a\right)} \left(e^{\epsilon_{\distQ}^{-}} \prob{\omega \sim \distP}{\omega \in \func{B_{\varepsilon}^{a}}{\distQ, \distP}} + e^{\epsilon_{\distQ}^{-}} \delta \right) \\
& = e^{-a}\left(\prob{\omega \sim \distP}{\omega \in \func{B_{\varepsilon}^{a}}{\distQ, \distP}} + \delta \right)
\end{align*}
where (a) results from the fact the inequality holds for any element
$\omega \in \Omega$ and (b) from the previous bound.

Reordering the terms we get
\begin{align*}
\prob{\omega \sim \distP}{\lnFunc{\frac{\distQfunc{\omega}}{\distPfunc{\omega}}} > \func{\varepsilon}{\omega} + a} = \prob{\omega \sim \distP}{\omega \in \func{B_{\varepsilon}^{a}}{\distQ, \distP}} < & \frac{\delta}{e^{a}-1}.
\end{align*}
\end{proof}

Next we prove an important connection between the two dissimilarity measures, starting with a supporting lemma which is a slight variation of Theorem 2.7 in~\citet{DKL07}. This technique is sometimes referred to as the \emph{method of mixtures}.

\begin{lemma} \label{lem:mthdMix}
Given two jointly distributed random variables $\tuple{A}{B} \sim \dist$, if $\expectation{\tuple{A}{B} \sim D}{\expFunc{\lambda A - \frac{\lambda^{2} B^{2}}{2}}} \le 1$ for all $\lambda \in \reals$, then for any $\delta, b > 0$,
\[
\prob{\tuple{A}{B} \sim D}{\abs{A} > 2 \sqrt{\lnFunc{\frac{1}{\delta}} \left(B^{2} + b^{2} \right)}} \le \delta \sqrt{\expectation{A, B \sim D}{\sqrt{\frac{B^{2}}{b^{2}} + 1}}}.
\]
\end{lemma}
As~\citet{DKL07} note, at first glance this claim might seem trivial, and one can expect to prove it by simply optimizing over $\lambda$ to show $\expectation{A, B, \sim D}{\expFunc{\frac{A}{2 B^{2}}}} \le 1$ and combining with Markov's inequality. The problem is, the optimal $\lambda$ depends on the value of $B$ which is a random variables, while $\lambda$ must be set beforehand. To solve this, we define a distribution over $\lambda$ and use Fubini's theorem. 

\begin{proof}
Consider a Gaussian distribution over $\lambda$ with parameters $\mu = 0$, $\sigma = \frac{1}{b}$.
\begin{align*}
1 & \eqRes{a}{\ge} \expectation{\lambda \sim \func{\mathcal{N}}{0, \frac{1}{b^{2}}}}{\expectation{A, B, \sim D}{\expFunc{\lambda A - \frac{\lambda^{2} B^{2}}{2}}}} \\
& \eqRes{b}{=} \expectation{A, B, \sim D}{\expectation{\lambda \sim \func{\mathcal{N}}{0, \frac{1}{b^{2}}}}{\expFunc{\lambda A - \frac{\lambda^{2} B^{2}}{2}}}} \\
& = \expectation{A, B, \sim D}{\frac{b}{\sqrt{2 \pi}} \int_{-\infty}^{\infty} \expFunc{\lambda A - \frac{\lambda^{2} B^{2}}{2}} \expFunc{- \frac{\lambda^{2} b^{2}}{2}} d\lambda} \\
& = \expectation{A, B, \sim D}{\frac{b}{\sqrt{2 \pi}} \int_{-\infty}^{\infty} \expFunc{-\frac{B^{2} + b^{2}}{2} \left(\lambda^{2} - 2 \lambda \frac{A}{B^{2} + b^{2}} \right)} d\lambda} \\
& \eqRes{c}{=} \expectation{A, B, \sim D}{\frac{b}{\sqrt{B^{2} + b^{2}}} \expFunc{\frac{A^{2}}{2 \left(B^{2} + b^{2} \right)}} \frac{\sqrt{B^{2} + b^{2}}}{\sqrt{2 \pi}} \int_{-\infty}^{\infty} \expFunc{-\frac{B^{2} + b^{2}}{2} \left(\lambda - \frac{A}{B^{2} + b^{2}} \right)^{2}} d\lambda} \\
& \eqRes{d}{=} \expectation{A, B, \sim D}{\frac{b}{\sqrt{B^{2} + b^{2}}} \expFunc{\frac{A^{2}}{2 \left(B^{2} + b^{2} \right)}} \expectation{\lambda \sim \func{\mathcal{N}}{\frac{A}{B^{2} + b^{2}}, \frac{1}{B^{2} + b^{2}}}}{1}} \\
& = \expectation{A, B, \sim D}{\frac{b}{\sqrt{B^{2} + b^{2}}} \expFunc{\frac{A^{2}}{2 \left(B^{2} + b^{2} \right)}}}.
\end{align*}
where (a) results from the fact the inequality $\expectation{A, B, \sim D}{\expFunc{\lambda A - \frac{\lambda^{2} B^{2}}{2}}} \le 1$ holds for all $\lambda$, which implies it holds for the expectation taken with respect to any distribution over $\lambda$ as well, (b) results from Fubini's theorem, (c) is simply completing the square, and (d) from the definition of the Gaussian distribution.

Using this bound we get
\begin{align*}
\underset{A, B, \sim D}{\text{Pr}} & \left[\abs{A} > 2 \sqrt{\lnFunc{\frac{1}{\delta}} \left(B^{2} + b^{2} \right)} \right] \\
& = \prob{A, B, \sim D}{\expFunc{\frac{A^{2}}{4 \left(B^{2} + b^{2} \right)}} > \frac{1}{\delta}} \\
& \eqRes{a}{\le} \delta \expectation{A, B, \sim D}{\expFunc{\frac{A^{2}}{4 \left(B^{2} + b^{2} \right)}}} \\
& = \delta \expectation{A, B, \sim D}{\sqrt{\frac{\sqrt{B^{2} + b^{2}}}{b}} \sqrt{\frac{b}{\sqrt{B^{2} + b^{2}}}} \expFunc{\frac{A^{2}}{4 \left(B^{2} + b^{2} \right)}}} \\
& \eqRes{b}{\le} \delta \sqrt{\expectation{A, B, \sim D}{\frac{\sqrt{B^{2} + b^{2}}}{b}} \expectation{A, B, \sim D}{\frac{b}{\sqrt{B^{2} + b^{2}}} \expFunc{\frac{A^{2}}{2 \left(B^{2} + b^{2} \right)}}}} \\
& \eqRes{c}{\le} \delta \sqrt{\expectation{A, B, \sim D}{\sqrt{\frac{B^{2}}{b^{2}} + 1}}},
\end{align*}
where (a) results from Markov's inequality, (b) from the Cauchy-Schwarz inequality, and (c) from the previous inequality.
\end{proof}

We can now prove the implication Theorem.
\begin{theorem} \label{thm:dynRenImpDynMax}
Given a non-negative function $\varphi: \Omega \rightarrow \pos$ and two distributions $\distP, \distQ$, if the $\alpha$-R\'{e}nyi dissimilarity between $\distP$ and $\distQ$ is bounded by $\alpha \varphi$ for any $\alpha \ge 0$, then $\prob{\omega \sim \distP}{\abs{\lnFunc{\frac{\distPfunc{\omega}}{\distQfunc{\omega}}}} > \func{\varepsilon}{\omega}} \le \delta \sqrt{\expectation{\omega \sim \distP}{\sqrt{\frac{\func{\varphi}{\omega}}{\phi} + 1}}}$ for any $\phi, \delta > 0$, where $\func{\varepsilon}{\omega} \coloneqq \func{\varphi}{\omega} + 2 \sqrt{2 \lnFunc{\frac{1}{\delta}} \left(\func{\varphi}{\omega} + \phi \right)}$.
\end{theorem}

\begin{proof}
    Denoting $A \coloneqq \lnFunc{\frac{\distPfunc{\omega}}{\distQfunc{\omega}}} - \func{\varphi}{\omega}$ and $B \coloneqq \sqrt{2 \func{\varphi}{\omega}}$, for any $\lambda \ge 0$ we denote $\alpha = \lambda + 1$ and from the assumption we get that
    \[
        \expectation{A,B \sim \dist}{\expFunc{\lambda A - \frac{\lambda^{2} B^{2}}{2}}}
        \eqRes{a}{\le} \expectation{\omega \sim \distP}{\expFunc{\left(\alpha - 1 \right) \left(\lnFunc{\frac{\distPfunc{\omega}}{\distQfunc{\omega}}} - \alpha \func{\varphi}{\omega} \right)}} \eqRes{b}{\le} 1
    \]
    
    For any $\lambda < 0$ we denote $\alpha = -\lambda$ and from the assumption we get that
    \begin{align*}
        \expectation{A,B \sim \dist}{\expFunc{\lambda A - \frac{\lambda^{2} B^{2}}{2}}} & \eqRes{a}{\le} \expectation{\omega \sim \distP}{\expFunc{\lambda \left(\lnFunc{\frac{\distPfunc{\omega}}{\distQfunc{\omega}}} - \left(\lambda + 1 \right) \func{\varphi}{\omega} \right)}}
        \\ & \eqRes{c}{=} \expectation{\omega \sim \distP}{\expFunc{\alpha \left(\lnFunc{\frac{\distQfunc{\omega}}{\distPfunc{\omega}}} - \left(\alpha - 1 \right) \func{\varphi}{\omega} \right)}}
        \\ & = \expectation{\omega \sim \distP}{\left(\frac{\distQfunc{\omega}}{\distPfunc{\omega}} \right)^{\alpha} \expFunc{- \alpha \left(\alpha - 1 \right) \func{\varphi}{\omega}}}
        \\ & \eqRes{d}{=} \expectation{\omega \sim \distQ}{\left(\frac{\distQfunc{\omega}}{\distPfunc{\omega}} \right)^{\alpha - 1} \expFunc{- \alpha \left(\alpha - 1 \right) \func{\varphi}{\omega}}}
        \\ & = \expectation{\omega \sim \distQ}{\expFunc{\left(\alpha - 1 \right) \left(\lnFunc{\frac{\distQfunc{\omega}}{\distPfunc{\omega}}} - \alpha \func{\varphi}{\omega} \right)}}
        \\ & \eqRes{b}{\le} 1
    \end{align*}
    where (a) results from the definition of $\lambda$, (b) from the R\'{e}nyi dissimilarity assumption, (c) from the fact $\lnFunc{\frac{a}{b}} = -\lnFunc{\frac{b}{a}}$, and (d) is a renormalization step.
    
    Combining the two with Lemma \ref{lem:mthdMix} and setting $\phi \coloneqq 2 b^{2}$ we get
    \begin{align*}
        \prob{\omega \sim \distP}{\abs{\lnFunc{\frac{\distPfunc{\omega}}{\distQfunc{\omega}}}} > \func{\varepsilon}{\omega}} & = \prob{\omega \sim \distP}{\abs{\lnFunc{\frac{\distPfunc{\omega}}{\distQfunc{\omega}}}} > \func{\varphi}{\omega} + 2 \sqrt{2 \lnFunc{\frac{1}{\delta}} \left(\func{\varphi}{\omega} + \phi \right)}}
        \\ & \eqRes{a}{\le} \prob{\omega \sim \distP}{\abs{\lnFunc{\frac{\distPfunc{\omega}}{\distQfunc{\omega}}} - \func{\varphi}{\omega}} > + 2 \sqrt{2 \lnFunc{\frac{1}{\delta}} \left(\func{\varphi}{\omega} + \phi \right)}}
        \\ & = \prob{\tuple{A}{B} \sim D}{\abs{A} > 2 \sqrt{\lnFunc{\frac{1}{\delta}} \left(B^{2} + b^{2} \right)}}
        \\ & \eqRes{b}{\le} \delta \sqrt{\expectation{A, B \sim D}{\sqrt{\frac{B^{2}}{b^{2}} + 1}}}
        \\ & = \delta \sqrt{\expectation{\omega \sim \distP}{\sqrt{\frac{\func{\varphi}{\omega}}{\phi} + 1}}},
    \end{align*}
    where (a) results from the triangle inequality and (b) from Lemma \ref{lem:mthdMix}.
\end{proof}

\subsection{Convexity} \label{sec:conv}
Next we prove some convexity results for max and R\'{e}nyi dissimilarities.

\begin{theorem}[Max dissimilarity convexity] \label{thm:dynMaxDisConv}
Given $\lambda \in \left(0, 1\right)$, $\delta_{i} \ge 0$, distributions $\distP_{i},\distQ_{i}$, and non-negative functions $\varepsilon_{i} : \omega \rightarrow \pos$ for $i \in \left\{0, 1 \right\}$, denote $\distP_{\lambda} \coloneqq \lambda \distP_{0} + \left(1 - \lambda \right) \distP_{1}$,
$\distQ_{\lambda} \coloneqq \lambda \distQ_{0} + \left(1 - \lambda \right) \distQ_{1}$, $\func{\varepsilon_{\lambda}}{\omega} \coloneqq \lnFunc{\lambda e^{\func{\varepsilon_{0}}{\omega}} + \left(1-\lambda \right) e^{\func{\varepsilon_{1}}{\omega}}}$, and $\delta_{\lambda} \coloneqq \lambda \delta_{0} + \left(1 - \lambda \right) \delta_{1}$.

\begin{enumerate}
\item Joint quasi-convexity: If $\distP_{i} \DPMD{i}{i} \distQ_{i}$ for $i \in \left\{0, 1 \right\}$, then $\distP_{\lambda} \DPMD{\max}{\lambda} \distQ_{\lambda}$, where $\func{\varepsilon_{\max}}{\omega} \coloneqq \max \left\{\func{\varepsilon_{0}}{\omega}, \func{\varepsilon_{1}}{\omega} \right\}$
\item Left-hand log-convexity: If $\distP_{i} \DPMD{i}{i} \distQ$ for $i \in \left\{0, 1 \right\}$, then $\distP_{\lambda} \DPMD{\lambda}{\lambda} \distQ$
\item Right-hand log-convexity: If $\distP \DNMD{i}{i} \distQ_{i}$ for $i \in \left\{0, 1 \right\}$, then $\distP \DNMD{\lambda}{\lambda} \distQ_{\lambda}$
\end{enumerate}
\end{theorem}

\begin{proof}
\textbf{Part 1}: For any $\boldsymbol{\omega} \subseteq \Omega$,
\begin{align*}
\int_{\omega \in \boldsymbol{\omega}}\func{\distP_{\lambda}}{\omega} d\omega & \eqRes{a}{=} \int_{\omega \in \boldsymbol{\omega}} \left(\lambda \func{\distP_{0}}{\omega} + \left(1-\lambda\right) \func{\distP_{1}}{\omega} \right) d\omega \\
& = \lambda \int_{\omega \in \boldsymbol{\omega}} \func{\distP_{0}}{\omega} d\omega + \left(1 - \lambda \right) \int_{\omega \in \boldsymbol{\omega}} \func{\distP_{1}}{\omega} d\omega \\
& \eqRes{b}{\le} \lambda \left(\int_{\omega \in \boldsymbol{\omega}} e^{\func{\varepsilon_{0}}{\omega}} \func{\distQ_{0}}{\omega} d\omega + \delta_{0} \right) + \left(1 - \lambda \right) \left(\int_{\omega \in \boldsymbol{\omega}} e^{\func{\varepsilon_{1}}{\omega}} \func{\distQ_{1}}{\omega} d\omega + \delta_{1} \right) \\
& \eqRes{c}{\le} \lambda \left(\int_{\omega \in \boldsymbol{\omega}} e^{\func{\varepsilon_{\max}}{\omega}} \func{\distQ_{0}}{\omega} d\omega + \delta_{0} \right) + \left(1 - \lambda \right) \left(\int_{\omega \in \boldsymbol{\omega}} e^{\func{\varepsilon_{\max}}{\omega}} \func{\distQ_{1}}{\omega} d\omega + \delta_{1} \right) \\
& \eqRes{d}{=} \int_{\omega \in \boldsymbol{\omega}} e^{\func{\varepsilon_{\max}}{\omega}} \func{\distQ_{\lambda}}{\omega} d\omega + \delta_{\lambda}
\end{align*}

where (a) results from the definition of $\distP_{\lambda}$, (b) from the assumption $\distP_{i} \DPMD{i}{i} \distQ_{i}$, (c) from the definitions of $\varepsilon_{\max}$ and $\delta_{\lambda}$, and (d) from the definition of $\distQ_{\lambda}$.

\textbf{Part 2}: For any $\boldsymbol{\omega} \subseteq \Omega$,
\begin{align*}
\int_{\omega \in \boldsymbol{\omega}} \func{\distP_{\lambda}}{\omega} d\omega & \eqRes{a}{=} \int_{\omega \in \boldsymbol{\omega}} \left(\lambda \func{\distP_{0}}{\omega} + \left(1 - \lambda \right) \func{\distP_{1}}{\omega}\right) d\omega \\
& = \lambda \int_{\omega \in \boldsymbol{\omega}} \func{\distP_{0}}{\omega} d\omega + \left(1 - \lambda \right) \int_{\omega \in \boldsymbol{\omega}} \func{\distP_{1}}{\omega} d\omega \\
& \eqRes{b}{\le} \lambda \left(\int_{\omega \in \boldsymbol{\omega}} e^{\func{\varepsilon_{0}}{\omega}} \distQfunc{\omega} d\omega + \delta_{0} \right) + \left(1 - \lambda \right) \left(\int_{\omega \in \boldsymbol{\omega}} e^{\func{\varepsilon_{1}}{\omega}} \distQfunc{\omega} d\omega + \delta_{1} \right) \\
& \eqRes{c}{=} \int_{\omega \in \boldsymbol{\omega}} \left(\lambda e^{\func{\varepsilon_{0}}{\omega}} + \left(1 - \lambda \right) e^{\func{\varepsilon_{1}}{\omega}} \right) \distQfunc{\omega} d\omega + \delta_{\lambda} \\
& \eqRes{d}{=} \int_{\omega \in \boldsymbol{\omega}} e^{\func{\varepsilon_{\lambda}}{\omega}} \distQfunc{\omega} d\omega + \delta_{\lambda}
\end{align*}
where (a) results from the definition of $\distP_{\lambda}$, (b) from the assumption $\distP_{i}\DPMD{i}{i} \distQ$, (c) from the definition of $\delta_{\lambda}$, and (d) from the definition of $\varepsilon_{\lambda}$.

\textbf{Part 3}: For any $\boldsymbol{\omega} \subseteq \Omega$,
\begin{align*}
\int_{\omega \in \boldsymbol{\omega}} e^{-\func{\varepsilon_{\lambda}}{\omega}} \distPfunc{\omega} d\omega & \eqRes{a}{=} \int_{\omega \in \boldsymbol{\omega}} \frac{1}{\lambda e^{\func{\varepsilon_{0}}{\omega}} + \left(1 - \lambda \right) e^{\func{\varepsilon_{1}}{\omega}}} \distPfunc{\omega} d\omega \\
& \eqRes{b}{\le} \int_{\omega \in \boldsymbol{\omega}} \left(\lambda e^{-\func{\varepsilon_{0}}{\omega}} + \left(1 - \lambda \right) e^{-\func{\varepsilon_{1}}{\omega}} \right) \distPfunc{\omega} d\omega \\
& = \lambda \left(\int_{\omega \in \boldsymbol{\omega}} e^{-\func{\varepsilon_{0}}{\omega}} \distPfunc{\omega} d\omega \right) + \left(1 - \lambda \right) \left(\int_{\omega \in \boldsymbol{\omega}} e^{-\func{\varepsilon_{1}}{\omega}} \distPfunc{\omega} d\omega \right) \\
& \eqRes{c}{\le} \lambda \left(\int_{\omega \in \boldsymbol{\omega}} \func{\distQ_{0}}{\omega} d\omega + \delta_{0} \right) + \left(1 - \lambda \right) \left(\int_{\omega \in \boldsymbol{\omega}} \func{\distQ_{1}}{\omega} d\omega + \delta_{1} \right) \\
& \eqRes{d}{=} \int_{\omega \in \boldsymbol{\omega}} \func{\distQ_{\lambda}}{\omega} d\omega + \delta_{\lambda}
\end{align*}
where (a) results from the definition of $\varepsilon_{\lambda}$, (b) from Jensen's inequality for the convex function $\frac{1}{x}$, (c) from the assumption $\distP \DNMD{i}{i} \distQ_{i}$,
and (d) from the definitions of $\distQ_{\lambda}$ and $\delta_{\lambda}$.
\end{proof}

\begin{theorem} [R\'{e}nyi dissimilarity convexity] \label{thm:dynRenDisConv}
Given $\alpha \ge 1$, $\lambda \in \left(0, 1\right)$, and non-negative functions $\varphi_{i} : \omega \rightarrow \pos$ for $i \in \left\{0, 1 \right\}$, denote $\distP_{\lambda} \coloneqq \lambda \distP_{0} + \left(1 - \lambda \right) \distP_{1}$, $\distQ_{\lambda} \coloneqq \lambda \distQ_{0} + \left(1 - \lambda \right) \distQ_{1}$, and $\func{\varphi_{\lambda}}{\omega} \coloneqq \lambda \func{\varphi_{0}}{\omega} + \left(1-\lambda \right) \func{\varphi_{1}}{\omega}$.

\begin{enumerate}
\item Joint quasi-convexity: If the $\alpha$-R\'{e}nyi dissimilarity between $\distP_{i}$ and $\distQ_{i}$ is bounded by $\varphi_{i}$ for $i \in \left\{0, 1 \right\}$, then the $\alpha$-R\'{e}nyi dissimilarity between $\distP_{\lambda}$ and $\distQ_{\lambda}$ is bounded by $\varphi_{\max}$, where $\func{\varphi_{\max}}{\omega} \coloneqq \max \left\{\func{\varphi_{0}}{\omega}, \func{\varphi_{1}}{\omega} \right\}$
\item Right-hand convexity: If the $\alpha$-R\'{e}nyi dissimilarity between $\distP$ and $\distQ_{i}$ is bounded by $\varphi_{i}$ for $i \in \left\{0, 1 \right\}$, then the $\alpha$-R\'{e}nyi dissimilarity between $\distP$ and $\distQ_{\lambda}$ is bounded by $\varphi_{\lambda}$
\end{enumerate}
\end{theorem}

\begin{proof}
The case of $\alpha = 1$ is trivial for both parts, so we focus on the $\alpha > 1$ case.

\textbf{Part 1}:
We first notice that
\begin{align*}
\func{\distP_{\lambda}}{\omega} & = \lambda \func{\distP_{0}}{\omega} + \left(1 - \lambda \right) \func{\distP_{1}}{\omega} \\
& = \lambda \left(\func{\distQ_{0}}{\omega} \right)^{\frac{\alpha - 1}{\alpha}} \frac{\func{\distP_{0}}{\omega}}{\left(\func{\distQ_{0}}{\omega} \right)^{\frac{\alpha - 1}{\alpha}}} + \left(1 - \lambda \right) \left(\func{\distQ_{1}}{\omega} \right)^{\frac{\alpha - 1}{\alpha}} \frac{\func{\distP_{1}}{\omega}}{\left(\func{\distQ_{1}}{\omega} \right)^{\frac{\alpha - 1}{\alpha}}} \\
& \eqRes{a}{\le} \left(\lambda \left(\left(\func{\distQ_{0}}{\omega} \right)^{\frac{\alpha - 1}{\alpha}}\right)^{\frac{\alpha}{\alpha - 1}} + \left(1 - \lambda \right) \left(\left(\func{\distQ_{1}}{\omega} \right)^{\frac{\alpha - 1}{\alpha}}\right)^{\frac{\alpha}{\alpha - 1}}\right)^{\frac{\alpha - 1}{\alpha}} \\
& ~~~~ \cdot \left(\lambda \left(\frac{\func{\distP_{0}}{\omega}}{\left(\func{\distQ_{0}}{\omega} \right)^{\frac{\alpha - 1}{\alpha}}} \right)^{\alpha} + \left(1 - \lambda \right) \left(\frac{\func{\distP_{1}}{\omega}}{\left(\func{\distQ_{1}}{\omega} \right)^{\frac{\alpha - 1}{\alpha}}} \right)^{\alpha} \right)^{\frac{1}{\alpha}} \\
& = \left(\lambda \func{\distQ_{0}}{\omega} + \left(1 - \lambda \right) \func{\distQ_{1}}{\omega} \right)^{\frac{\alpha - 1}{\alpha}} \cdot \left(\lambda \frac{\left(\func{\distP_{0}}{\omega}\right)^{\alpha}}{\left(\func{\distQ_{0}}{\omega} \right)^{\alpha - 1}} + \left(1 - \lambda \right) \frac{\left(\func{\distP_{1}}{\omega}\right)^{\alpha}}{\left(\func{\distQ_{1}}{\omega} \right)^{\alpha - 1}} \right)^{\frac{1}{\alpha}} \\
& = \func{\distQ_{\lambda}}{\omega} \cdot \left(\lambda \frac{\func{\distP_{0}}{\omega}}{\func{\distQ_{\lambda}}{\omega}} \left(\frac{\func{\distP_{0}}{\omega}}{\func{\distQ_{0}}{\omega}}\right)^{\alpha - 1} + \left(1 - \lambda \right) \frac{\func{\distP_{1}}{\omega}}{\func{\distQ_{\lambda}}{\omega}} \left(\frac{\func{\distP_{1}}{\omega}}{\func{\distQ_{1}}{\omega}}\right)^{\alpha - 1} \right)^{\frac{1}{\alpha}} \\
\end{align*}
where (a) results from the generalized version of H\"{o}lder's inequality, $\expectation{}{X \cdot Y} \le \left(\expectation{}{X^p} \right)^{\frac{1}{p}} \cdot \left(\expectation{}{Y^q} \right)^{\frac{1}{q}}$ for any $p, q > 0$ such that $\frac{1}{p} + \frac{1}{q} = 1$.

Using this bound we get
\begin{align*}
\underset{\omega \sim \distP_{\lambda}}{\mathbb{E}} & \left[\expFunc{\left(\alpha - 1 \right) \left( \lnFunc{\frac{\func{\distP_{\lambda}}{\omega}}{\func{\distQ_{\lambda}}{\omega}}} - \func{\varphi_{\max}}{\omega} \right)} \right] \\
& = \expectation{\omega \sim \distP_{\lambda}}{\left(\frac{\func{\distP_{\lambda}}{\omega}}{\func{\distQ_{\lambda}}{\omega}} \right)^{\alpha - 1} \expFunc{-\left(\alpha - 1 \right) \func{\varphi_{\max}}{\omega}}} \\
& = \expectation{\omega \sim \distQ_{\lambda}}{\left(\frac{\func{\distP_{\lambda}}{\omega}}{\func{\distQ_{\lambda}}{\omega}} \right)^{\alpha} \expFunc{-\left(\alpha - 1 \right) \func{\varphi_{\max}}{\omega}}} \\
& \eqRes{a}{\le} \expectation{\omega \sim \distQ_{\lambda}}{\left(\lambda \frac{\func{\distP_{0}}{\omega}}{\func{\distQ_{\lambda}}{\omega}} \left(\frac{\func{\distP_{0}}{\omega}}{\func{\distQ_{0}}{\omega}}\right)^{\alpha - 1} + \left(1 - \lambda \right) \frac{\func{\distP_{1}}{\omega}}{\func{\distQ_{\lambda}}{\omega}} \left(\frac{\func{\distP_{1}}{\omega}}{\func{\distQ_{1}}{\omega}}\right)^{\alpha - 1} \right) \expFunc{-\left(\alpha - 1 \right) \func{\varphi_{\max}}{\omega}}} \\
& = \lambda \expectation{\omega \sim \distQ_{\lambda}}{\frac{\func{\distP_{0}}{\omega}}{\func{\distQ_{\lambda}}{\omega}} \left(\frac{\func{\distP_{0}}{\omega}}{\func{\distQ_{0}}{\omega}}\right)^{\alpha - 1} \expFunc{-\left(\alpha - 1 \right) \func{\varphi_{\max}}{\omega}}} \\
& ~~~~ + \left(1 - \lambda \right) \expectation{\omega \sim \distQ_{\lambda}}{\frac{\func{\distP_{1}}{\omega}}{\func{\distQ_{\lambda}}{\omega}} \left(\frac{\func{\distP_{1}}{\omega}}{\func{\distQ_{1}}{\omega}}\right)^{\alpha - 1} \expFunc{-\left(\alpha - 1 \right) \func{\varphi_{\max}}{\omega}}} \\
& \eqRes{b}{=} \lambda \expectation{\omega \sim \distP_{0}}{\expFunc{\left(\alpha - 1 \right) \left( \lnFunc{\frac{\func{\distP_{0}}{\omega}}{\func{\distQ_{0}}{\omega}}} - \func{\varphi_{\max}}{\omega} \right)}} \\
& ~~~~ + \left(1 - \lambda \right) \expectation{\omega \sim \distP_{1}}{\expFunc{\left(\alpha - 1 \right) \left( \lnFunc{\frac{\func{\distP_{1}}{\omega}}{\func{\distQ_{1}}{\omega}}} - \func{\varphi_{\max}}{\omega} \right)}} \\
& \eqRes{c}{\le} \lambda \expectation{\omega \sim \distP_{0}}{\expFunc{\left(\alpha - 1 \right) \left( \lnFunc{\frac{\func{\distP_{0}}{\omega}}{\func{\distQ_{0}}{\omega}}} - \func{\varphi_{0}}{\omega} \right)}} \\
& ~~~~ + \left(1 - \lambda \right) \expectation{\omega \sim \distP_{1}}{\expFunc{\left(\alpha - 1 \right) \left( \lnFunc{\frac{\func{\distP_{1}}{\omega}}{\func{\distQ_{1}}{\omega}}} - \func{\varphi_{1}}{\omega} \right)}} \\
& \eqRes{d}{\le} 1 \\
\end{align*}
where (a) results from the previous inequality, (b) is a renormalization step, (c) from the the definition of $\varphi_{\max}$, and (d) from the R\'{e}nyi dissimilarity assumption.

\textbf{Part 2}:
\begin{align*}
\underset{\omega \sim \distP}{\mathbb{E}} & \left[\expFunc{\left(\alpha - 1 \right) \left( \lnFunc{\frac{\distPfunc{\omega}}{\func{\distQ_{\lambda}}{\omega}}} - \func{\varphi_{\lambda}}{\omega} \right)} \right] \\
& = \expectation{\omega \sim \distP}{\expFunc{\left(\alpha - 1 \right) \left( \lnFunc{\frac{\distPfunc{\omega}}{\lambda \func{\distQ_{0}}{\omega} + \left(1 - \lambda \right) \func{\distQ_{1}}{\omega}}} - \func{\varphi_{\lambda}}{\omega} \right)}} \\
& \eqRes{a}{\le} \expectation{\omega \sim \distP}{\expFunc{\left(\alpha - 1 \right) \left( \lambda \lnFunc{\frac{\distPfunc{\omega}}{\func{\distQ_{0}}{\omega}}} + \left(1 - \lambda \right) \lnFunc{\frac{\distPfunc{\omega}}{\func{\distQ_{1}}{\omega}}} - \func{\varphi_{\lambda}}{\omega} \right)}} \\
& \eqRes{b}{=} \expectation{\omega \sim \distP}{\expFunc{\lambda \left(\alpha - 1 \right) \left(\lnFunc{\frac{\distPfunc{\omega}}{\func{\distQ_{0}}{\omega}}} - \func{\varphi_{0}}{\omega}\right) + \left(1 - \lambda \right) \left(\alpha - 1 \right) \left(\lnFunc{\frac{\distPfunc{\omega}}{\func{\distQ_{1}}{\omega}}} - \func{\varphi_{1}}{\omega} \right)}} \\
& \eqRes{c}{\le} \lambda \expectation{\omega \sim \distP}{\expFunc{\left(\alpha - 1 \right) \left(\lnFunc{\frac{\distPfunc{\omega}}{\func{\distQ_{0}}{\omega}}} - \func{\varphi_{0}}{\omega}\right)}} \\
& ~~~~ + \left(1 - \lambda \right) \expectation{\omega \sim \distP}{\expFunc{\left(\alpha - 1 \right) \left(\lnFunc{\frac{\distPfunc{\omega}}{\func{\distQ_{1}}{\omega}}} - \func{\varphi_{1}}{\omega} \right)}} \\
& \eqRes{d}{\le} 1
\end{align*}
where (a) results from Jensen's inequality for the convex function $\lnFunc{\frac{1}{x}}$, (b) from the definition of $\varphi_{\lambda}$, (c) from Jensen's inequality for the convex function $e^{x}$ over the $\lambda$ weighted combination, and (d) from the R\'{e}nyi dissimilarity assumption.
\end{proof}

\subsection{Stability measures} \label{sec:stabMeas}
The various divergence and dissimilarity measures considered in the previous sections were used to define several stability notions; the divergence (dissimilarity) between the distributions over responses induced by two differing input datasets serves as a measure of the stability of the mechanism producing those responses.

One major application of stability notions is in privacy-preserving mechanisms.
\begin{definition} [Differential privacy~\citep{DMNS06}] \label{def:DP}
Given $\epsilon, \delta \ge 0$, a mechanism $\mechanism$ will be called \emph{$\tuple{\epsilon}{\delta}$-differentially private} (or DP, for short) if for any two datasets $\sampleSet, \sampleSet' \in \domainOfSets$ that differ only in one element, and any query $\query \in \queryFamily$, the $\delta$-approximate max divergence between the two distributions defined over $\responseFamily$ by $\mechanismFunc{\sampleSet, \query}$ and $\mechanismFunc{\sampleSet', \query}$ is bounded by $\epsilon$.
\end{definition}

As in Definition \ref{def:acc}, $\delta$ can be viewed as a function of $\epsilon$. In this case, $\delta$ is essentially a tail bound on the distribution of the privacy loss random variable (Definition \ref{def:stabLoss}). An alternative stability notion can be based on a bound on the moments of the stability loss.

\begin{definition} [R\'{e}nyi differential privacy~\citep{Mironov17}] \label{def:RDP}
Given $\alpha, \phi \ge 0$, a mechanism $\mechanism$ will be called \emph{$\tuple{\alpha}{\phi}$-R\'{e}nyi differentially private} (or RDP, for short) if for any two datasets $\sampleSet, \sampleSet' \in \domainOfSets$ that differ only in one element, and any query $\query \in \queryFamily$, the $\alpha$-R\'{e}nyi divergence between the two distributions defined over $\responseFamily$ by $\mechanismFunc{\sampleSet, \query}$ and $\mechanismFunc{\sampleSet', \query}$ is bounded by $\phi$.
\end{definition}

The following two definitions bound the shape of the privacy loss curve.

\begin{definition} [Concentrated differential privacy~\citep{DR16}] \label{def:CDP} 
Given $\mu, \sigma \ge 0$, a mechanism $\mechanism$ will be called \emph{$\tuple{\mu}{\sigma}$-concentrated differentially private} (or CDP, for short) if for any two datasets $\sampleSet, \sampleSet' \in \domainOfSets$ that differ only in one element, and any query $\query \in \queryFamily$, the stability loss random variable (Definition \ref{def:stabLoss}) is a $\sigma^{2}$-sub-Gaussian random variable (Definition \ref{def:subGauss}) with expectation bounded by $\mu$. Formally, $\expectation{\responseRV \sim \mechanismFunc{\sampleSet, \query}}{\loss{\sampleSet}{\sampleSet'}{\responseRV}} \le \mu$ and for any $\alpha \ge 1$ we have 
\[
\expectation{\responseRV \sim \mechanismFunc{\sampleSet, \query}}{\expFunc{\left(\alpha - 1 \right) \left(\loss{\sampleSet}{\sampleSet'}{\responseRV} - \expectation{\responseRV' \sim \mechanismFunc{\sampleSet, \query}}{\loss{\sampleSet}{\sampleSet'}{\responseRV'}} \right)}} \le e^{\frac{\alpha^{2}}{2 \sigma^{2}}}.
\]
\end{definition}

\begin{definition} [Zero-concentrated differential privacy~\citep{BS16}] \label{def:zCDP}
Given $\phi \ge 0$, a mechanism $\mechanism$ will be called \emph{$\phi$-zero concentrated differentially private} (or zCDP, for short) if for any two datasets $\sampleSet, \sampleSet' \in \domainOfSets$ that differ only in one element, any query $\query \in \queryFamily$, and $\alpha \ge 1$, the $\alpha$-R\'{e}nyi divergence between the two distributions defined over $\responseFamily$ by $\mechanismFunc{\sampleSet, \query}$ and $\mechanismFunc{\sampleSet', \query}$ is bounded by $\alpha \phi$ in both directions.\footnote{The original definition includes an additional parameter that is equal to $0$ in the case of the Gaussian mechanism, so we omit it here for simplicity.}

This definition can be viewed as a special case of the CDP definition, with $\mu = \phi$ and $\sigma^{2} = 2 \phi$. The comparison is formalized in Lemma 24 in~\citep{BS16}.
\end{definition}

The above notions all hold over \emph{all} pairs of neighboring datasets, which can be achieved by scaling the randomness added by the mechanism to the worst-case sensitivity---that is, to the worst-case change in the output that can be induced by replacing one dataset by a neighboring one. As discussed in detail in Appendix \ref{apd:relNot}, to avoid the dependence on the range of the queries, our approach instead focuses on a version of the the ``local sensitivity,'' which, in turn, involves transitioning from divergences to dissimilarity notions.

\begin{definition} [Pairwise concentration; equivalent to Definition \ref{def:PC} but restated in terms of divergence and dissimilarity] Given a non-negative function $\PC: \domainOfSets \times \domainOfSets \times \queryFamily \rightarrow \pos$ which is symmetric in its first two arguments, a mechanism $\mechanism$ will be called \emph{$\PC$-Pairwise Concentrated} (or PC, for short), if for any $\sampleSet, \sampleSet' \in \domainOfSets$, query $\query \in \queryFamily$, and $\alpha \ge 0$, the $\alpha$-R\'{e}nyi divergence between the two distributions defined over $\responseFamily$ by $\mechanismFunc{\sampleSet, \query}$ and $\mechanismFunc{\sampleSet', \query}$ is bounded by $\alpha \PCqFunc{\sampleSet}{\sampleSet'}$ in both directions.

Given a non-negative function $\PC: \domainOfSets \times \domainOfSets \times \viewFamily \rightarrow \pos$ which is symmetric in its first two arguments and an analyst $\analyst$, a mechanism $\mechanism$ will be called $\PC$-Pairwise Concentrated with respect to $\analyst$, if for any $\sampleSet, \sampleSet' \in \domainOfSets$ and $\alpha \ge 0$, the $\alpha$-R\'{e}nyi dissimilarity between the two distributions defined over $\viewFamily$ by $\mechanismFunc{\sampleSet, \analyst}$ and $\mechanismFunc{\sampleSet', \analyst}$ is bounded by $\alpha \PCfuncL{\sampleSet}{\sampleSet'}{\cdot}$ in both directions.
\end{definition}

Analogously to how one can generalize the zCDP stability notion by replacing the divergence by a dissimilarity measure, one can also generalize other stability notions such as differential privacy.

\begin{definition} [Pairwise indistinguishability] Given $\delta \ge 0$ and a non-negative function $\PI: \domainOfSets \times \domainOfSets \times \queryFamily \rightarrow \pos$ which is symmetric in its first two arguments, a mechanism $\mechanism$ will be called \emph{$\PI$-Pairwise Indistinguishable} (or PI, for short), if for any $\sampleSet, \sampleSet' \in \domainOfSets$, and any query $\query \in \queryFamily$, the max divergence between the two distributions defined over $\responseFamily$ by $\mechanismFunc{\sampleSet, \query}$ and $\mechanismFunc{\sampleSet', \query}$ is bounded by $\PIqFunc{\sampleSet}{\sampleSet'}$ in both directions.

Given a non-negative function $\PI: \domainOfSets \times \domainOfSets \times \viewFamily \rightarrow \pos$ which is symmetric in its first two arguments and an analyst $\analyst$, a mechanism $\mechanism$ will be called $\PI$-Pairwise Indistinguishable with respect to $\analyst$, if for any $\sampleSet, \sampleSet' \in \domainOfSets$, the max dissimilarity between the two distributions defined over $\viewFamily$ by $\mechanismFunc{\sampleSet, \analyst}$ and $\mechanismFunc{\sampleSet', \analyst}$ is bounded by $\PIfuncL{\sampleSet}{\sampleSet'}{\cdot}$ in both directions.
\end{definition}

We note that Theorem \ref{thm:PCstab} can be stated in terms of pairwise indistinguishability as well.

\newpage
\section{Missing parts from Section \ref{sec:gen}} \label{apd:gen}

\subsection{Missing proofs}

\begin{lemma} \label{lem:probExp}
    Given a probability distribution $\distP$ over a product domain $\mathcal{X} \times \mathcal{Y}$ and a function $f : \mathcal{X} \times \mathcal{Y} \rightarrow \reals$, for any $\epsilon, \xi > 0$ we have,
    \[
        \prob{X \sim \distP_{\mathcal{X}}}{\abs{\expectation{Y \sim \distP_{\mathcal{Y} \vert \mathcal{X}}}{\func{f}{X, Y} \, \vert \, X}} > \epsilon + \xi} \le \frac{1}{\xi} \int_{\epsilon}^{\infty} \prob{\left(X, Y \right) \sim \distP}{\abs{\func{f}{X, Y}} > t} dt.
    \]
\end{lemma}

\begin{proof} 
    \begin{align*}
        \prob{X \sim \distP_{\mathcal{X}}}{\abs{\expectation{Y \sim \distP_{\mathcal{Y} \vert \mathcal{X}}}{\func{f}{X, Y} \, \vert \, X}} > \epsilon + \xi} & \eqRes{a}{\le} \prob{X \sim \distP_{\mathcal{X}}}{\left[\abs{\expectation{Y \sim \distP_{\mathcal{Y} \vert \mathcal{X}}}{\func{f}{X, Y} \, \vert \, X}} - \epsilon \right]^{+} > \xi}
        \\ & \eqRes{b}{\le} \frac{1}{\xi} \expectation{X \sim \distP_{\mathcal{X}}}{\left[\abs{\expectation{Y \sim \distP_{\mathcal{Y} \vert \mathcal{X}}}{\func{f}{X, Y} \, \vert \, X}} - \epsilon \right]^{+}}
        \\ & \eqRes{c}{\le} \frac{1}{\xi} \expectation{X \sim \distP_{\mathcal{X}}}{\left[\expectation{Y \sim \distP_{\mathcal{Y} \vert \mathcal{X}}}{\abs{\func{f}{X, Y}} \, \vert \, X} - \epsilon \right]^{+}}
        \\ & \eqRes{d}{\le} \frac{1}{\xi} \expectation{X \sim \distP_{\mathcal{X}}}{\expectation{Y \sim \distP_{\mathcal{Y} \vert \mathcal{X}}}{\left[\abs{\func{f}{X, Y}} - \epsilon \right]^{+} \, \vert \, X}}
        \\ & = \frac{1}{\xi} \expectation{\left(X, Y \right) \sim \distP}{\left[\abs{\func{f}{X, Y}} - \epsilon \right]^{+}}
        \\ & \eqRes{e}{=} \frac{1}{\xi} \int_{0}^{\infty} \prob{\left(X, Y \right) \sim \distP}{\left[\abs{\func{f}{X, Y}} - \epsilon \right]^{+} > t} dt
        \\ & = \frac{1}{\xi} \int_{\epsilon}^{\infty} \prob{\left(X, Y \right) \sim \distP}{\abs{\func{f}{X, Y}} > t} dt
    \end{align*}
    where (a) $\left[x \right]^{ + } \coloneqq \max \left\{x, 0 \right\}$ is the ReLU function, (b) results from Markov's inequality, (c) from the triangle inequality which implies $\abs{\expectation{}{X}} \le \expectation{}{\abs{X}}$ and $\left[\expectation{}{X} \right]^{+} \le \expectation{}{\left[X \right]^{+}}$, (d) from Jensen's inequality for the convex  ReLU function, and (e) from the fact that the expectation is taken over a non-negative random variable.
    \end{proof}

\pstLem{samAccImpPostAcc}

\begin{proof}
    For any $\view \in \viewFamily$, denote $\func{I}{\view} \coloneqq \underset{i \in \left[ \numOfIterations \right]}{\arg \max} \abs{\response_{i} - \func{\query_{i}}{\distPost}}$, and $\response_{\view} \coloneqq \response_{\func{I}{\view}}, \query_{\view} \coloneqq \query_{\func{I}{\view}}$. 
    When the view is a random variable $\viewRV$, the corresponding query $\queryRV_{\viewRV}$ and response $\responseRV_{\viewRV}$ are also random variables, and hence we denote them with capital letters.
    \begin{align*}
        \\ \prob{\viewRV \sim \dist}{\underset{i \in \rangeOfIterations}{\max} \abs{\responseRV_{i} - \func{\queryRV_{i}}{\distUpInd{}{\viewRV}}} > \epsilon} & = \prob{\viewRV \sim \dist}{\abs{\responseRV_{\viewRV} - \func{\queryRV_{\viewRV}}{\distUpInd{}{\viewRV}}} > \epsilon}
        \\ & = \prob{\viewRV \sim \dist}{\abs{\expectation{\sampleSetRV \sim \distFuncDep{\cdot}{\viewRV}}{\responseRV_{\viewRV} - \func{\queryRV_{\viewRV}}{\sampleSetRV}}} > \epsilon}
        \\ & \eqRes{a}{\le} \frac{1}{\xi} \int_{\epsilon - \xi}^{\infty} \prob{\sampleSetRV \sim \iiDist, \viewRV \sim \mechanismFunc{\sampleSetRV, \analyst}}{\abs{\responseRV_{\viewRV} - \func{\queryRV_{\viewRV}}{\sampleSetRV}} > u} du
        \\ & = \frac{1}{\xi} \int_{\epsilon - \xi}^{\infty} \prob{\sampleSetRV \sim \iiDist, \viewRV \sim \mechanismFunc{\sampleSetRV, \analyst}}{\underset{i \in \rangeOfIterations}{\max} \abs{\responseRV_{i} - \func{\queryRV_{i}}{\sampleSetRV}} > u} du
        \\ & \eqRes{b}{\le} \frac{1}{\xi} \int_{\epsilon - \xi}^{\infty} \func{\delta}{t} dt
        \end{align*}
    where (a) results from part one of Lemma \ref{lem:probExp} (where the term $\epsilon$ that appears in Lemma \ref{lem:probExp} is set to $\epsilon' = \epsilon - \xi$) and (b) from the definition of sample accuracy (Definition \ref{def:acc}).
\end{proof}

\pstLem{GaussAcc}

\begin{proof}
In the case of a Gaussian distribution, $\func{\delta}{\epsilon}$ is known as the Q-function and for any dataset $\sampleSet$ and query $\query$,
\[
\prob{\responseRV \sim \mechanismFunc{\sampleSet, \query}}{\abs{\responseRV - \queryFunc{\sampleSet}} > \epsilon} = \func{\text{erfc}}{\frac{\epsilon}{\sqrt{2} \eta}} \le \frac{2}{\sqrt{\pi}} e^{-\frac{\epsilon^{2}}{2 \eta^{2}}}.
\]

Combining this with a union bound completes the proof of sample accuracy.

Invoking Lemma \ref{lem:samAccImpPostAcc} with $\xi = \epsilon - \sqrt{\epsilon^{2} - 2 \eta^{2}}$ implies for all $\epsilon \ge \sqrt{2} \eta$ and any $i \in \rangeOfIterations$,

\begin{align*}
\prob{\sampleSetRV \sim \iiDist, \viewRV \sim \mechanismFunc{\sampleSetRV, \analyst}}{\func{\text{err}_{P}}{\sampleSetRV, \viewRV, i} > \epsilon} & \le \frac{1}{\xi} \int_{\epsilon - \xi}^{\infty} \func{\text{erfc}}{\frac{t}{\sqrt{2 \eta}}} dt \\
& \eqRes{u \coloneqq \frac{t}{\sqrt{2}\eta}}{=} \frac{\sqrt{2} \eta}{\xi} \int_{\frac{\epsilon - \xi}{\sqrt{2} \eta}}^{\infty} \func{\text{erfc}}{u} du \\
& \eqRes{a}{<} \frac{\sqrt{2} \eta \cdot e^{-\frac{\epsilon^{2} - 2 \eta^{2}}{2 \eta^{2}}}}{\sqrt{\pi} \epsilon \left( 1 - \sqrt{1 - \frac{2 \eta^{2}}{\epsilon^{2}}} \right)} \\
& \eqRes{b}{\le} \frac{\sqrt{2} e \epsilon}{\sqrt{\pi} \eta} e^{-\frac{\epsilon^{2}}{2 \eta^{2}}} \\
& \eqRes{c}{\le} \frac{2 e}{\sqrt{\pi}} e^{-\frac{\epsilon^{2}}{4 \eta^{2}}}
\\ & \le 4 e^{-\frac{\epsilon^{2}}{4 \eta^{2}}}
\end{align*}
where (a) results from the fact that $\int_{a}^{\infty} \func{\text{erfc}}{x} dx = \frac{e^{- a^{2}}}{\sqrt{\pi}} - a \cdot \func{\text{erfc}}{a} < \frac{e^{- a^{2}}}{\sqrt{\pi}}$ and the definition of $\xi$, (b) from the bound on $\epsilon$ and the inequality $\sqrt{1 + x} \le 1 + \frac{x}{2}$, and (c) from the inequality $x \cdot e^{-x^{2}} \le e^{-\frac{x^{2}}{2}}$.

Combining this with a union bound completes the proof of posterior accuracy.
\end{proof}

\pstThm{BayesStabImpdistPcc}

\begin{proof}
    Consider an analyst $\analyst$, which once completed an interaction of length $\numOfIterations$ resulting in a view $\view$, chooses $\query_{\numOfIterations + 1} \coloneqq \underset{i \in \rangeOfIterations}{\arg \max} \abs{\func{\query_{i}}{\distPost} - \func{\query_{i}}{\dist}}$. In this case,
    \begin{align*}
        \underset{\genfrac{}{}{0pt}{}{\sampleSetRV \sim \iiDist}{\viewRV \sim \mechanismFunc{\sampleSetRV, \analyst}}}{\text{Pr}} & \left[\underset{i \in \rangeOfIterations}{\max} \abs{\responseRV_{i} - \func{\queryRV_{i}}{\dist}} > \epsilon \right]
        \\ & \le \prob{\genfrac{}{}{0pt}{}{\sampleSetRV \sim \iiDist}{\viewRV \sim \mechanismFunc{\sampleSetRV, \analyst}}}{\underset{i \in \rangeOfIterations}{\max} \abs{\responseRV_{i} - \func{\queryRV_{i}}{\distUpInd{}{\viewRV}}} + \underset{i \in \rangeOfIterations}{\max} \abs{\func{\queryRV_{i}}{\distUpInd{}{\viewRV}} - \func{\queryRV_{i}}{\dist}} > \epsilon}
        \\ & \le \prob{\genfrac{}{}{0pt}{}{\sampleSetRV \sim \iiDist}{\viewRV \sim \mechanismFunc{\sampleSetRV, \analyst}}}{\underset{i \in \rangeOfIterations}{\max} \abs{\responseRV_{i} - \func{\queryRV_{i}}{\distUpInd{}{\viewRV}}} > \epsilon - \epsilon'} + \prob{\genfrac{}{}{0pt}{}{\sampleSetRV \sim \iiDist}{\viewRV \sim \mechanismFunc{\sampleSetRV, \analyst}, \queryRV \sim \func{\analyst}{\viewRV}}}{\abs{\func{\queryRV}{\distUpInd{}{\viewRV}} - \func{\queryRV}{\dist}} > \epsilon'}
        \\ & \eqRes{a}{\le} \frac{1}{\xi} \int_{\epsilon - \epsilon' - \xi}^{\infty} \func{\delta}{t} dt + \func{\delta}{\epsilon'},
    \end{align*}
    where (a) results from Lemma \ref{lem:samAccImpPostAcc} and the definition of Bayes stability (Definition \ref{def:bysStab}).
    \end{proof}

\subsection{Generalized results for arbitrary queries} \label{sec:arbQuer}
While the main results in this paper were stated for linear queries, some of the intermediate results extend to arbitrary queries. Denoting by $\bar{\Delta}_{\query} \coloneqq \underset{\sampleSet, \sampleSet' \in \elementsDomain}{\sup}\abs{\queryFunc{\sampleSet} - \queryFunc{\sampleSet}}$ and $\bar{\sigma}_{\query}^{2} \coloneqq \expectation{\elementRV \sim \distDomain}{\left(\queryFunc{\elementRV} - \queryFunc{\distInd{\domainOfSets}} \right)^{2}}$ and extending the posterior distribution notations to $\distUpInd{\domainOfSets}{\view} \coloneqq \func{\distUpInd{\domainOfSets | \viewFamily}{\analyst}}{\cdot \,|\, \view}$, Lemma \ref{lem:covStab} can be restated in terms of arbitrary queries.

\begin{lemma}[Covariance stability for arbitrary queries] 
    Given a view $\view \in \viewFamily$ and a query $\query$,
    \[
        \queryFunc{\dist_{\domainOfSets}^{\view}} - \queryFunc{\dist_{\domainOfSets}} = \cov{\sampleSetRV \sim \distDomainOfSets}{\queryFunc{\sampleSetRV}}{\BayesF{\sampleSetRV}{\view}}.
    \]
    
    Furthermore, given $\Delta, \sigma > 0$,
    \[
        \underset{q \in \queryFamily ~\text{s.t.}~ \bar{\Delta}_{\query} \le \Delta}{\sup} \abs{\queryFunc{\dist_{\domainOfSets}^{\view}} - \queryFunc{\dist_{\domainOfSets}}} = 2 \Delta \cdot \func{\mathbf{D}_{\text{TV}}}{\dist_{\domainOfSets}^{\view} \Vert \dist_{\domainOfSets}}
    \]
    and
    \[
        \underset{q \in \queryFamily ~\text{s.t.}~ \bar{\sigma}_{\query}^{2} \le \sigma^{2}}{\sup} \abs{\queryFunc{\dist_{\elementsDomain}^{\view}} - \queryFunc{\dist_{\elementsDomain}}} = \sigma \sqrt{\chiSqr{\dist_{\domainOfSets}^{\view}}{\dist_{\domainOfSets}}},
    \]
    where $\BayesF{\sampleSet}{\view} \coloneqq \frac{\distFuncDep{\sampleSet}{\view}}{\distFunc{\sampleSet}} = \frac{\distFuncDep{\view}{\sampleSet}}{\distFunc{\view}}$ 
    is the Bayes factor of $\sampleSet$ given $\view$ (and vice-versa).
\end{lemma}

Combining this lemma with the PC definition for arbitrary sample sets, we can state a simple version of Theorem \ref{thm:PCstab} for arbitrary queries, and achieve results similar to those guaranteed using typical stability ~\citep{BF16}. Like in the case of typical stability, this method does not achieve optimal accuracy rates for the iid case as discussed by \citet{KSS22}.

\newpage
\section{Missing parts from Section \ref{sec:PC}}
\label{apd:PC}
\subsection{PC mechanisms}

\pstLem{GaussPC}

\begin{proof}
From Lemma 16 in~\citep{BS16}, for any query $\query$, we have $\PCqFunc{\sampleSet}{\sampleSet'} = \frac{\left(\queryFunc{\sampleSet} - \queryFunc{\sampleSet'} \right)^{2}}{2 \eta^{2}}$.\footnote{The lemma is stated for the range $\alpha \in \left[1, \infty \right]$, but holds for the range $\left[0, 1 \right)$ as well, directly from the integral definition.} Combining this result with the composition theorem (Theorem~\ref{thm:PCcomp}) completes the proof.
\end{proof}

\subsection{Composition}

\pstThm{PCcomp}

\begin{proof}
Notice that
\[\loss{\sampleSet}{\sampleSet'}{\view_{\numOfIterations}} = \lnFunc{\frac{\distFuncDep{\view_{\numOfIterations}}{\sampleSet}}{\distFuncDep{\view_{\numOfIterations}}{\sampleSet'}}} = \sum_{i=1}^{\numOfIterations} \lnFunc{\frac{\distFuncDep{\response_{i}}{\sampleSet, \view_{i-1}}}{\distFuncDep{\response_{i}}{\sampleSet', \view_{i-1}}}} \eqRes{a}{=} \sum_{i=1}^{\numOfIterations} \lnFunc{\frac{\func{\dist^{\query_{i}}}{\response_{i} \,|\, \sampleSet}}{\func{\dist^{\query_{i}}}{\response_{i} \,|\, \sampleSet'}}} = \sum_{i=1}^{\numOfIterations} \loss{\sampleSet}{\sampleSet'}{\response_{i}}
\]

where (a) results from the fact that given $\sampleSet/\sampleSet'$ the additional information of $\view_{i-1}$ does not change the distribution over $\response_{i}$, since it can be encoded in the auxiliary parameter $\theta_{i}$. In case of a non-deterministic analyst, the sum includes an additional term of the form $\lnFunc{\frac{\distFuncDep{c}{\sampleSet}}{\distFuncDep{c}{\sampleSet'}}}$ for the coin-tossing step, as discussed in Section \ref{sec:formAdap}. But since $c$ is independent of the dataset, the ratio is equal to $1$ and this term has no effect.

Using this fact we can prove the theorem by induction over $\numOfIterations$. The base case $\numOfIterations = 0$ is trivial. For every $\numOfIterations > 0$,
\begin{align*}
\underset{\viewRV_{\numOfIterations} \sim \mechanismFunc{\sampleSet, \analyst}}{\mathbb{E}} & \left[e^{\left(\alpha - 1 \right) \left( \loss{\sampleSet}{\sampleSet'}{\viewRV_{\numOfIterations}} - \alpha \PCfuncL{\sampleSet}{\sampleSet'}{\viewRV_{\numOfIterations}} \right)} \right] \\
& = \expectation{\viewRV_{\numOfIterations - 1} \sim \mechanismFunc{\sampleSet, \analyst}}{ \expectation{\viewRV_{\numOfIterations} \sim \mechanismFunc{\sampleSet, \analyst}}{e^{\left(\alpha - 1 \right) \left( \loss{\sampleSet}{\sampleSet'}{\viewRV_{\numOfIterations}} - \alpha \PCfuncL{\sampleSet}{\sampleSet'}{\viewRV_{\numOfIterations}} \right)} \,|\, \viewRV_{\numOfIterations - 1}}} \\
& \eqRes{a}{=} \expectation{\viewRV_{\numOfIterations - 1} \sim \mechanismFunc{\sampleSet, \analyst}}{e^{\left(\alpha - 1 \right) \left( \loss{\sampleSet}{\sampleSet'}{\viewRV_{\numOfIterations - 1}} - \alpha \PCfuncL{\sampleSet}{\sampleSet'}{\viewRV_{\numOfIterations - 1}} \right)} \cdot \overset{\le 1}{\overbrace{\expectation{\responseRV_{\numOfIterations} \sim \mechanismFunc{\sampleSet, \queryRV_{\numOfIterations}}}{e^{\left(\alpha - 1 \right) \left( \loss{\sampleSet}{\sampleSet'}{\responseRV_{\numOfIterations}} - \alpha \PCfuncL{\sampleSet}{\sampleSet'}{\queryRV_{\numOfIterations}} \right)}}}}} \\
& \eqRes{b}{\le} \expectation{\viewRV_{\numOfIterations - 1} \sim \mechanismFunc{\sampleSet, \analyst}}{e^{\left(\alpha - 1 \right) \left( \loss{\sampleSet}{\sampleSet'}{\viewRV_{\numOfIterations - 1}} - \alpha \PCfuncL{\sampleSet}{\sampleSet'}{\viewRV_{\numOfIterations - 1}} \right)}} \\
& \eqRes{c}{\le} 1,
\end{align*}
where (a) results from the definition of $\PCfuncL{\cdot}{\cdot}{\bar{\query}}$, the previous equation and the fact that both $\loss{\sampleSet}{\sampleSet'}{\viewRV_{\numOfIterations - 1}}$ and $\PCfuncL{\sampleSet}{\sampleSet'}{\viewRV_{\numOfIterations - 1}}$ are independent of $\responseRV_{\numOfIterations}$, (b) follows from the PC definition, and (c) follows from the induction assumption.
\end{proof}

In the next section we prove Lemma \ref{lem:PCelem}, as well as a bound on the R\'{e}nyi divergence between the prior and posterior distributions over elements, which will later be used in the proof of Theorem \ref{thm:PCstab}.

\subsection{Expected stability bound}
\pstLem{PCelem}

\begin{proof}
    The notation used in this proof is defined in Appendix \ref{apd:divAndDis}. 
    
    Notice that since the dataset was sampled iid, we have $\distFuncDep{\view}{\element} = \expectation{\sampleSetRV \sim \dist^{\left(\sampleSize - 1 \right)}}{\distFuncDep{\view}{\tuple{\sampleSetRV}{\element}}}$, so the first part is a direct result of Part 1 of Theorem \ref{thm:dynRenDisConv}, where $\distP_{\lambda} \coloneqq \distFuncDep{\cdot}{\element}$, $\distQ_{\lambda} \coloneqq \distFuncDep{\cdot}{\elementY}$, and the convexity is taken over the convex combination of the sample sets of size $\sampleSize - 1$.

    For the second part, notice that $\distFunc{\view} = \expectation{\elementRV \sim \dist}{\distFuncDep{\view}{\elementRV}}$, so
    \begin{align*}
        \underset{\viewRV \sim \distFuncDep{\cdot}{\element}}{\mathbb{E}} & \left[\expFunc{\left(\alpha - 1 \right) \left(\lossS{\element}{\viewRV} - \alpha \PCfuncS{\element}{\viewRV} \right)} \right] \\
        & \eqRes{a}{\le} \expectation{\viewRV \sim \distFuncDep{\cdot}{\element}}{\expFunc{\left(\alpha - 1 \right) \left(\lossS{\element}{\viewRV} - \alpha \expectation{\elementRVY \sim \dist}{\PCvFunc{\element}{\elementRVY}} \right)}} \\ & \eqRes{b}{\le} 1,
    \end{align*}
    where (a) results from Jensen's inequality for the convex function $e^{x}$ and (b) from combining the first part with Part 2 of Theorem \ref{thm:dynRenDisConv}, where $\distP \coloneqq \distFuncDep{\cdot}{\element}$, $\distQ_{\lambda} \coloneqq \dist$, and the convexity is taken over the convex combination of the elements.
\end{proof}

Next, we bound the R\'{e}nyi divergence between the prior and posterior distributions for PC mechanisms.

\begin{lemma} \label{lem:PCexp}
Given an element $\element \in \elementsDomain$, a similarity function over views $\PC$ and an analyst $A$, if a mechanism $\mechanism$ that is $\PC$-PC with respect to $\analyst$ receives an iid sample from $\domainOfSets$, then for any $\alpha \ge 1$ we have
\[
\func{\mathbf{D}_{\alpha}}{\func{\distInd{\viewFamily}}{\cdot | \element} \Vert \distInd{\viewFamily}} \le \frac{1}{2 \left(\alpha - 1 \right)} \lnFunc{\expectation{\viewRV \sim \dist}{\expFunc{2 \alpha \left(2 \alpha - 1 \right) \PCfuncS{\element}{\viewRV}}}},
\]
where $\PCfuncS{\element}{\view}$ is as was defined in Lemma \ref{lem:PCelem}.
\end{lemma}

\begin{proof}
\begin{align*}
\exp & \left(\left(\alpha - 1 \right) \func{\mathbf{D}_{\alpha}}{\func{\distInd{\viewFamily}}{\cdot | \elementRV} \Vert \distInd{\viewFamily}} \right) \\
& = \expectation{\viewRV \sim \distFuncDep{\cdot}{\element}}{\left(\frac{\distFuncDep{\viewRV}{\element}}{\distFunc{\viewRV}} \right)^{\alpha - 1}} \\
& = \expectation{\viewRV \sim \dist}{\left(\frac{\distFuncDep{\viewRV}{\element}}{\distFunc{\viewRV}} \right)^{\alpha}} \\
& = \expectation{\viewRV \sim \dist}{\expFunc{\alpha\left(\lossS{\element}{\viewRV} - \left(2 \alpha - 1 \right) \PCfuncS{\element}{\viewRV} \right)} \cdot \expFunc{\alpha \left(2 \alpha - 1 \right) \PCfuncS{\element}{\viewRV}}} \\
& \eqRes{a}{\le} \sqrt{\expectation{\viewRV \sim \dist}{\expFunc{2 \alpha \left(\lossS{\element}{\viewRV} - \left(2 \alpha - 1 \right) \PCfuncS{\element}{\viewRV} \right)}} \cdot \expectation{\viewRV \sim \dist}{\expFunc{2 \alpha \left(2 \alpha - 1 \right) \PCfuncS{\element}{\viewRV}}}} \\
& = \sqrt{\expectation{\viewRV \sim \dist}{\left(\frac{\distFuncDep{\viewRV}{\element}}{\distFunc{\viewRV}} \right)^{2 \alpha} \expFunc{- 2 \alpha \left(2 \alpha - 1 \right) \PCfuncS{\element}{\viewRV}}} \cdot \expectation{\viewRV \sim \dist}{\expFunc{2 \alpha \left(2 \alpha - 1 \right) \PCfuncS{\element}{\viewRV}}}} \\
& = \sqrt{\expectation{\viewRV \sim \distFuncDep{\cdot}{\element}}{\left(\frac{\distFuncDep{\viewRV}{\element}}{\distFunc{\viewRV}} \right)^{2 \alpha - 1} \expFunc{- 2 \alpha \left(2 \alpha - 1 \right) \PCfuncS{\element}{\viewRV}}} \cdot \expectation{\viewRV \sim \dist}{\expFunc{2 \alpha \left(2 \alpha - 1 \right) \PCfuncS{\element}{\viewRV}}}} \\
& = \sqrt{\expectation{\viewRV \sim \distFuncDep{\cdot}{\element}}{\expFunc{\left(2 \alpha - 1 \right) \left(\lossS{\element}{\viewRV} - 2 \alpha \PCfuncS{\element}{\viewRV} \right)}} \cdot \expectation{\viewRV \sim \dist}{\expFunc{2 \alpha \left(2 \alpha - 1 \right) \PCfuncS{\element}{\viewRV}}}} \\
& \eqRes{b}{\le} \sqrt{\expectation{\viewRV \sim \dist}{\expFunc{2 \alpha \left(2 \alpha - 1 \right) \PCfuncS{\element}{\viewRV}}}} \\
\end{align*}
where (a) results from the Cauchy-Schwarz inequality and (b) from the second part of Lemma \ref{lem:PCelem}.

Taking the log of both sides and dividing by $\alpha - 1$ completes the proof.
\end{proof}

In the next section we prove a high probability bound on the stability loss in terms of the similarity function, as a second step toward proving Theorem \ref{thm:PCstab}.

\subsection{High probability stability bound}
\begin{lemma} \label{lem:PChighProb}
Given a similarity function over views $\PC$ and an analyst $A$, if a mechanism $\mechanism$ that is $\PC$-PC with respect to $\analyst$ receives an iid sample from $\domainOfSets$, then
\[
\prob{\genfrac{}{}{0pt}{}{\elementRV \sim \dist, \sampleSetRV \sim \iiDist}{\viewRV \sim \mechanismFunc{\sampleSetRV, \analyst}}}{\abs{\lossS{\elementRV}{\viewRV}} > \PIfuncS{\elementRV}{\viewRV}} \le \frac{2^{\frac{1}{4}} \left(e^{\phi} + 1 \right) \delta^{2}}{e^{\phi} - 1} \left(\expectation{\genfrac{}{}{0pt}{}{\elementRV \sim \dist}{\viewRV \sim \dist}}{\expFunc{12 \PCfuncS{\elementRV}{\viewRV}}} \right)^{\frac{1}{8}}
\]
for any $\delta > 0$, $\phi \ge \expectation{\genfrac{}{}{0pt}{}{\elementRV \sim \dist, \sampleSetRV \sim \iiDist}{\viewRV \sim \mechanismFunc{\sampleSetRV, \analyst}}}{\PCfuncS{\elementRV}{\viewRV}}$, where $\PIfuncS{\element}{\view} \coloneqq \PCfuncS{\element}{\view} + \phi + 2 \sqrt{2 \lnFunc{\frac{1}{\delta}} \left(\PCfuncS{\element}{\view} + \phi \right)}$ and $\PCfuncS{\element}{\view}$ is as was defined in Lemma \ref{lem:PCelem}.
\end{lemma}

Before we prove this lemma, we present several supporting claims. Throughout, we will fix a similarity function over views $\PC$, an analyst $A$, and a mechanism $\mechanism$ which is $\PC$-PC with respect to $\analyst$ receiving an iid sample from $\domainOfSets$. The notation used here is all defined in Appendix \ref{apd:divAndDis}.

\begin{claim} \label{clm:PChighProb_1}
Given $\phi, \delta > 0$ and two sample sets $\sampleSet, \sampleSet' \in \domainOfSets$, we have $\distFuncDep{\cdot}{\sampleSet} \DPMD{\sampleSet, \sampleSet'}{\sampleSet, \sampleSet'} \distFuncDep{\cdot}{\sampleSet'}$, where $\func{\varepsilon_{\sampleSet, \sampleSet'}}{\view} \coloneqq \PCvFunc{\sampleSet}{\sampleSet'} + 2 \sqrt{2 \lnFunc{\frac{1}{\delta}} \left(\PCvFunc{\sampleSet}{\sampleSet'} + \phi \right)}$ and $\delta_{\sampleSet, \sampleSet'} \coloneqq \delta \sqrt{\expectation{\viewRV \sim \distFuncDep{\cdot}{\sampleSet}}{\sqrt{\frac{\PCfuncL{\sampleSet}{\sampleSet'}{\viewRV}}{\phi} + 1}}}$.
\end{claim}

\begin{proof}
Denote $\Omega \coloneqq \viewFamily$, $\distP \coloneqq \distFuncDep{\cdot}{\sampleSet}$, and $\distQ \coloneqq \distFuncDep{\cdot}{\sampleSet'}$. From the PC definition, for any $\alpha \ge 0$ the $\alpha$-R\'{e}nyi dissimilarity between $\distP$ and $\distQ$ is bounded by $\alpha \cdot \PCfuncL{\sampleSet}{\sampleSet'}{\cdot}$.

Invoking Theorem \ref{thm:dynRenImpDynMax} we get $\prob{\viewRV \sim \distP}{\abs{\lnFunc{\frac{\distPfunc{\viewRV}}{\distQfunc{\viewRV}}}} > \func{\tilde{\varepsilon}_{\sampleSet, \sampleSet'}}{\viewRV}} \le \delta_{\sampleSet, \sampleSet'}$, where 
\[
\func{\tilde{\varepsilon}_{\sampleSet, \sampleSet'}}{\view} \coloneqq \PCvFunc{\sampleSet}{\sampleSet'} + 2 \sqrt{2 \lnFunc{\frac{1}{\delta}} \left(\PCvFunc{\sampleSet}{\sampleSet'} + \phi \right)}.
\]

Combining this with Part 1 of Theorem \ref{thm:maxDisImp} we get that $\distP \DPMD{\sampleSet, \sampleSet'}{\sampleSet, \sampleSet'} \distQ$.
\end{proof}

\begin{claim} \label{clm:PChighProb_2}
Given $\phi, \delta > 0$ and two elements $\element, \elementY \in \elementsDomain$, we have $\distFuncDep{\cdot}{\element} \DPMD{\element, \elementY}{\element, \elementY} \distFuncDep{\cdot}{\elementY}$ and $\distFuncDep{\cdot}{\element} \DNMD{\element, \elementY}{\element, \elementY} \distFuncDep{\cdot}{\elementY}$, where $\func{\varepsilon_{\element, \elementY}}{\view} \coloneqq \func{\underset{\sampleSet \in \elementsDomain^{\sampleSize-1}}{\sup}}{\func{\varepsilon_{\tuple{\sampleSet}{\element}, \tuple{\sampleSet}{\elementY}}}{\view}}$, $\delta_{\element, \elementY} \coloneqq \expectation{\sampleSetRV \sim \dist^{\left(\sampleSize - 1 \right)}}{\delta_{\tuple{\sampleSetRV}{\element}, \tuple{\sampleSetRV}{\elementY}}}$, and $\varepsilon_{\sampleSet, \sampleSet'}$, $\delta_{\sampleSet, \sampleSet'}$ are defined as in Claim \ref{clm:PChighProb_1}.
\end{claim}

\begin{proof}
Notice that from the iid assumption, $\distFuncDep{\view}{\element} = \expectation{\sampleSetRV \sim \dist^{\left(\sampleSize - 1 \right)}}{\distFuncDep{\view}{\tuple{\sampleSetRV}{\element}}}$, so combining Claim \ref{clm:PChighProb_1} and Part 1 of Theorem \ref{thm:dynMaxDisConv} implies $\distFuncDep{\cdot}{\element} \DPMD{\element, \elementY}{\element, \elementY} \distFuncDep{\cdot}{\elementY}$. Combining it with Part 2 of Theorem \ref{thm:maxDisImp} implies $\distFuncDep{\cdot}{\element} \DNMD{\element, \elementY}{\element, \elementY} \distFuncDep{\cdot}{\elementY}$.
\end{proof}

\begin{claim} \label{clm:PChighProb_3}
Given $\phi, \delta > 0$ and an element $\element \in \elementsDomain$, we have $\distInd{\viewFamily} \DPMD{\element}{\element, *} \func{\distInd{\viewFamily \vert \elementsDomain}}{\cdot \vert \element}$ and $\func{\distInd{\viewFamily \vert \elementsDomain}}{\cdot \vert \element} \DNMD{\element}{*, \element} \distInd{\viewFamily}$, where $\func{\varepsilon_{\element}}{\view} \coloneqq \lnFunc{\expectation{\elementRVY \sim \dist}{e^{\PIvFunc{\element}{\elementRVY}}}}$, $\delta_{\element, *} \coloneqq \expectation{\elementRVY \sim \dist}{\delta_{\element, \elementRVY}}$, $\delta_{*, \element} \coloneqq \expectation{\elementRVY \sim \dist}{\delta_{\elementRVY, \element}}$, and $\varepsilon_{\element, \elementY}$, $\delta_{\element, \elementY}$ are defined as in Claim \ref{clm:PChighProb_2}.
\end{claim}

\begin{proof}
Notice that $\distFunc{\view} = \expectation{\elementRVY \sim \dist}{\distFuncDep{\view}{\elementRVY}}$, so combining the first part of Claim \ref{clm:PChighProb_2} and Part 2 of Theorem \ref{thm:dynMaxDisConv} implies $\distFunc{\cdot} \DPMD{\element}{\element, *} \distFuncDep{\cdot}{\element}$.

Similarly, combining the second part of Claim \ref{clm:PChighProb_2} and Part 3 of Theorem \ref{thm:dynMaxDisConv} implies $\distFuncDep{\cdot}{\element} \DNMD{\element}{*, \element} \distFunc{\cdot}$. Notice that the symmetry of $\PC$ implies $\PI$ is symmetric as well, so the same $\PI$ function bounds both terms.
\end{proof}

\begin{claim} \label{clm:PChighProb_4}
Given $\phi, \epsilon, \delta > 0$ and an element $\element \in \elementsDomain$, we have $\prob{\viewRV \sim \dist}{\lnFunc{\frac{\distFunc{\viewRV}}{\distFuncDep{\viewRV}{\element}}} > \func{\PI_{\element}}{\viewRV} + \epsilon} < \frac{e^{\epsilon} \delta_{\element, *}}{e^{\epsilon} - 1}$ and $\prob{\viewRV \sim \dist}{\lnFunc{\frac{\distFuncDep{\viewRV}{\element}}{\distFunc{\viewRV}}} > \func{\PI_{\element}}{\omega} + \epsilon} < \frac{\delta_{*, \element}}{e^{\epsilon} - 1}$, where $\PI_{\element}$, $\delta_{\element, *}$, and $\delta_{*, \element}$ are defined as in Claim \ref{clm:PChighProb_3}.
\end{claim}

\begin{proof}
The first part results from combining the first part of Claim \ref{clm:PChighProb_3} and Part 3 of Theorem \ref{thm:maxDisImp}. The second part results from combining the second part of Claim \ref{clm:PChighProb_3} and Part 4 of Theorem \ref{thm:maxDisImp}.
\end{proof}

\begin{proof} [Proof of Lemma \ref{lem:PChighProb}]
Notice that for any $\phi, \tilde{\delta} > 0$ we have
\begin{align*}
\func{\varepsilon_{\element, \elementY}}{\view} & = \func{\underset{\sampleSet \in \elementsDomain^{\sampleSize-1}}{\sup}}{\func{\varepsilon_{\tuple{\sampleSet}{\element}, \tuple{\sampleSet}{\elementY}}}{\view}} \\
& = \func{\underset{\sampleSet \in \elementsDomain^{\sampleSize-1}}{\sup}}{\PCvFunc{\tuple{\sampleSet}{\element}}{\tuple{\sampleSet}{\elementY}} + 2 \sqrt{2 \lnFunc{\frac{1}{\tilde{\delta}}} \left(\PCvFunc{\tuple{\sampleSet}{\element}}{\tuple{\sampleSet}{\elementY}} + \phi \right)}} \\
& \le \func{\underset{\sampleSet \in \elementsDomain^{\sampleSize-1}}{\sup}}{\PCvFunc{\tuple{\sampleSet}{\element}}{\tuple{\sampleSet}{\elementY}}} + 2 \sqrt{2 \lnFunc{\frac{1}{\tilde{\delta}}} \left(\func{\underset{\sampleSet \in \elementsDomain^{\sampleSize-1}}{\sup}}{\PCvFunc{\tuple{\sampleSet}{\element}}{\tuple{\sampleSet}{\elementY}}} + \phi \right)} \\
& = \PCvFunc{\element}{\elementY} + 2 \sqrt{2 \lnFunc{\frac{1}{\tilde{\delta}}} \left(\PCvFunc{\element}{\elementY} + \phi \right)} \\
 \end{align*}
which implies
\begin{align*}
\func{\varepsilon_{\element}}{\view} & = \expectation{\elementRVY \sim \dist}{\func{\varepsilon_{\element, \elementRVY}}{\view}} \\
& \eqRes{a}{\le} \expectation{\elementRVY \sim \dist}{\PCvFunc{\element}{\elementRVY} + 2 \sqrt{2 \lnFunc{\frac{1}{\tilde{\tilde{\delta}}}} \left(\PCvFunc{\element}{\elementRVY} + \phi \right)}} \\
& \eqRes{b}{\le} \expectation{\elementRVY \sim \dist}{\PCvFunc{\element}{\elementRVY}} + 2 \sqrt{2 \lnFunc{\frac{1}{\delta}} \left(\expectation{\elementRVY \sim \dist}{\PCvFunc{\element}{\elementRVY}} + \phi \right)} \\
& = \PCfuncS{\element}{\view} + 2 \sqrt{2 \lnFunc{\frac{1}{\tilde{\delta}}} \left(\PCfuncS{\element}{\view} + \phi \right)},
\end{align*}
where (a) results from the previous inequality and (b) from Jensen's inequality for the concave function $\sqrt{x}$.

Setting $\epsilon \coloneqq \phi$ and $\tilde{\delta} \coloneqq \delta^{2}$ we get
\begin{align*}
\underset{\genfrac{}{}{0pt}{}{\elementRV \sim \dist, \sampleSetRV \sim \iiDist}{\viewRV \sim \mechanismFunc{\sampleSetRV, \analyst}}}{\text{Pr}} & \left[\abs{\lossS{\elementRV}{\viewRV}} > \PIfuncS{\elementRV}{\viewRV} \right] \\
& = \expectation{\elementRV \sim \dist}{\prob{\genfrac{}{}{0pt}{}{\sampleSetRV \sim \iiDist}{\viewRV \sim \mechanismFunc{\sampleSetRV, \analyst}}}{\abs{\lnFunc{\frac{\distFuncDep{\viewRV}{\elementRV}}{\distFunc{\viewRV}}}} > \PCfuncS{\elementRV}{\viewRV} + \phi + 2 \sqrt{2 \lnFunc{\frac{1}{\tilde{\delta}}} \left(\PCfuncS{\elementRV}{\viewRV} + \phi \right)}}} \\
& \eqRes{a}{\le} \expectation{\elementRV \sim \dist}{\prob{\genfrac{}{}{0pt}{}{\sampleSetRV \sim \iiDist}{\viewRV \sim \mechanismFunc{\sampleSetRV, \analyst}}}{\abs{\lnFunc{\frac{\distFuncDep{\viewRV}{\elementRV}}{\distFunc{\viewRV}}}} > \func{\varepsilon_{\elementRV}}{\viewRV} + \epsilon}} \\
& \eqRes{b}{\le} \frac{e^{\epsilon}}{e^{\epsilon} - 1} \expectation{\elementRV \sim \dist}{\delta_{*, \elementRV}} + \frac{1}{e^{\epsilon} - 1} \expectation{\elementRV \sim \dist}{\delta_{\elementRV, *}} \\
& \eqRes{c}{=} \frac{e^{\epsilon} + 1}{e^{\epsilon} - 1} \expectation{\genfrac{}{}{0pt}{}{\elementRV \sim \dist, \elementRVY \sim \dist}{\sampleSetRV \sim \dist^{\left(\sampleSize - 1 \right)}}}{\tilde{\delta} \sqrt{\expectation{\viewRV \sim \mechanismFunc{\tuple{\sampleSetRV}{\elementRV}, \analyst}}{\sqrt{\frac{\PCfuncL{\tuple{\sampleSetRV}{\elementRV}}{\tuple{\sampleSetRV}{\elementRVY}}{\viewRV}}{\phi} + 1}}}} \\
& \eqRes{d}{\le} \frac{\left(e^{\phi} + 1 \right) \delta^{2}}{e^{\phi} - 1} \sqrt{\expectation{\genfrac{}{}{0pt}{}{\elementRV \sim \dist, \elementRVY \sim \dist, \sampleSetRV \sim \dist^{\left(\sampleSize - 1 \right)}}{\viewRV \sim \mechanismFunc{\tuple{\sampleSetRV}{\elementRV}, \analyst}}}{\sqrt{\frac{\PCfuncL{\tuple{\sampleSetRV}{\elementRV}}{\tuple{\sampleSetRV}{\elementRVY}}{\viewRV}}{\phi} + 1}}}, \\
\end{align*}
where (a) results from the previous bound, (b) from the two parts of Claim \ref{clm:PChighProb_4}, (c) from the definitions of $\delta_{\element, *}$ and $\delta_{*, \element}$, and (d) from Jensen's inequality for the concave function $\sqrt{x}$.

Analyzing the expectation term we get
\begin{align*}
\underset{\genfrac{}{}{0pt}{}{\elementRV \sim \dist, \elementRVY \sim \dist, \sampleSetRV \sim \dist^{\left(\sampleSize - 1 \right)}}{\viewRV \sim \mechanismFunc{\tuple{\sampleSetRV}{\elementRV}, \analyst}}}{\mathbb{E}} & \left[\sqrt{\frac{\PCfuncL{\tuple{\sampleSetRV}{\elementRV}}{\tuple{\sampleSetRV}{\elementRVY}}{\viewRV}}{\phi} + 1} \right] \\
& \eqRes{a}{\le} \expectation{\genfrac{}{}{0pt}{}{\elementRV \sim \dist, \elementRVY \sim \dist}{\viewRV \sim \distFuncDep{\cdot}{\elementRV}}}{\sqrt{\frac{\PCfuncL{\elementRV}{\elementRVY}{\viewRV}}{\phi} + 1}} \\
& = \expectation{\genfrac{}{}{0pt}{}{\elementRV \sim \dist, \elementRVY \sim \dist}{\viewRV \sim \dist}}{\frac{\distFuncDep{\viewRV}{\elementRV}}{\distFunc{\viewRV}} \sqrt{\frac{\PCfuncL{\elementRV}{\elementRVY}{\viewRV}}{\phi} + 1}} \\
& \eqRes{b}{\le} \sqrt{\expectation{\genfrac{}{}{0pt}{}{\elementRV \sim \dist}{\viewRV \sim \dist}}{\left(\frac{\distFuncDep{\viewRV}{\elementRV}}{\distFunc{\viewRV}} \right)^{2}} \cdot \expectation{\genfrac{}{}{0pt}{}{\elementRV \sim \dist, \elementRVY \sim \dist}{\viewRV \sim \dist}}{\frac{\PCfuncL{\elementRV}{\elementRVY}{\viewRV}}{\phi} + 1}} \\
& \eqRes{c}{\le} \sqrt{2 \expectation{\genfrac{}{}{0pt}{}{\elementRV \sim \dist}{\viewRV \sim \distFuncDep{\cdot}{\elementRV}}}{\frac{\distFuncDep{\elementRV}{\viewRV}}{\distFunc{\elementRV}}}} \\
& = \sqrt{2 \expectation{\elementRV \sim \dist}{\expFunc{\func{\mathbf{D}_{2}}{\func{\distInd{\viewFamily}}{\cdot | \elementRV} \Vert \distInd{\viewFamily}}}}} \\
& \eqRes{d}{\le} \sqrt{2} \left(\expectation{\genfrac{}{}{0pt}{}{\elementRV \sim \dist}{\viewRV \sim \dist}}{\expFunc{12 \PCfuncS{\elementRV}{\viewRV}}} \right)^{\frac{1}{4}},\\
\end{align*}
where (a) results from the definition of $\PCvFunc{\element}{\elementY}$, (b) from the Cauchy-Schwarz inequality, (c) from the bound on $\phi$, and (d) from Lemma \ref{lem:PCexp}.

Plugging this term back in completes the proof.
\end{proof}

We can now prove the main stability theorem for PC mechanisms.
\subsection{Bayes stability of PC mechanisms}
\begin{theorem} [Exact version of Theorem \ref{thm:PCstab}] \label{thm:PCstabExt}
    Given a similarity function over views $\PC$ and an analyst $A$, if a mechanism $\mechanism$ that is $\PC$-PC with respect to $\analyst$ receives an iid sample from $\domainOfSets$, then
    \begin{align*}
        \underset{\genfrac{}{}{0pt}{}{\sampleSetRV \sim \iiDist}{\viewRV \sim \mechanismFunc{\sampleSetRV, \analyst}, \queryRV \sim \func{\analyst}{\viewRV}}}{{\text{Pr}}} & \left[\abs{\func{\queryRV}{\dist_{\elementsDomain}^{\view}} - \func{\queryRV}{\dist_{\elementsDomain}}} > \sigma_{\queryRV} \cdot \sqrt{\epsilon^{2} + \xi} \right] \\
        & \le \prob{\genfrac{}{}{0pt}{}{\sampleSetRV \sim \iiDist}{\viewRV \sim \mechanismFunc{\sampleSetRV, \analyst}, \queryRV \sim \func{\analyst}{\viewRV}}}{\expectation{\elementRV \sim \dist}{e^{27  \lnFunc{\frac{1}{\delta}} \left(\PCfuncS{\elementRV}{\view} + \phi \right)}} > 1 + \frac{\epsilon^{2}}{3}} \\
        & + \frac{\delta}{\xi} \left(4 \func{\psi}{12} \right)^{\frac{1}{16}} \sqrt{\left(\sqrt{\func{\psi}{56}} + 1 \right) \cdot \frac{e^{\phi} + 1}{e^{\phi} - 1}} \\
    \end{align*}
    for any $\epsilon, \xi, \delta > 0$, $\phi \ge \expectation{\genfrac{}{}{0pt}{}{\elementRV \sim \dist, \sampleSetRV \sim \iiDist}{\viewRV \sim \mechanismFunc{\sampleSetRV, \analyst}}}{\PCfuncS{\elementRV}{\viewRV}}$, where $\PCfuncS{\element}{\view}$ is as was defined in Lemma \ref{lem:PCelem} and $\func{\psi}{\alpha} \coloneqq \expectation{\genfrac{}{}{0pt}{}{\elementRV \sim \dist, \sampleSetRV \sim \iiDist}{\viewRV \sim \mechanismFunc{\sampleSetRV, \analyst}}}{e^{\alpha \PCfuncS{\elementRV}{\viewRV}}}$.
\end{theorem}

We first prove several supporting claims.

\begin{claim} \label{clm:PCstabSupp_1}
Given a function $f : \elementsDomain \times \viewFamily \rightarrow \reals$, an analyst $A$, and a mechanism $\mechanism$, then for any $\xi > 0$ we have
\begin{align*}
\underset{\genfrac{}{}{0pt}{}{\sampleSetRV \sim \iiDist}{\viewRV \sim \mechanismFunc{\sampleSetRV, \analyst}}}{{\text{Pr}}} & \left[\chiSqr{\dist_{\elementsDomain}^{\view}}{\dist_{\elementsDomain}} > \expectation{\elementRV \sim \dist}{ \left(e^{\func{f}{\elementRV, \viewRV}} - 1 \right)^{2}} + \xi \right] \\
& \le \frac{1}{\xi} \sqrt{\left(\expectation{\genfrac{}{}{0pt}{}{\elementRV \sim \dist, \sampleSetRV \sim \iiDist}{\viewRV \sim \mechanismFunc{\sampleSetRV, \analyst}}}{e^{4 \lossS{\elementRV}{\viewRV}}} + 1 \right) \cdot \prob{\genfrac{}{}{0pt}{}{\elementRV \sim \dist, \sampleSetRV \sim \iiDist}{\viewRV \sim \mechanismFunc{\sampleSetRV, \analyst}}}{\abs{\lossS{\elementRV}{\viewRV}} > \func{f}{\elementRV, \viewRV}}}.
\end{align*}
\end{claim}

\begin{proof}
Denoting $\func{B}{f} \coloneqq \left\{\tuple{\element}{\view} \in \elementsDomain \times \viewFamily \,|\, \abs{\lossS{\element}{\view}} > \PIfuncS{\element}{\view} \right\}$ and $\func{\bar{B}}{f} \coloneqq \left(\elementsDomain \times \viewFamily \right) \backslash \func{B}{f}$ we get

\begin{align*}
\chiSqr{\dist_{\elementsDomain}^{\view}}{\dist_{\elementsDomain}} & - \expectation{\elementRV \sim \dist}{ \left(e^{\func{f}{\elementRV, \viewRV}} - 1 \right)^{2}} \\
& = \expectation{\elementRV \sim \dist}{\left(e^{\lossS{\elementRV}{\viewRV}} - 1 \right)^{2}} - \expectation{\elementRV \sim \dist}{\left(e^{\func{f}{\elementRV, \viewRV}} - 1 \right)^{2}} \\
& = \expectation{\elementRV \sim \dist}{\left(\left(e^{\lossS{\elementRV}{\viewRV}} - 1 \right)^{2} - \left(e^{\func{f}{\elementRV, \viewRV}} - 1 \right)^{2} \right) \cdot \indicator{\func{\bar{B}}{f}}{\tuple{\elementRV}{\viewRV}}} \\
& ~~~~ + \expectation{\elementRV \sim \dist}{\left(\left(e^{\lossS{\elementRV}{\viewRV}} - 1 \right)^{2} - \left(e^{\func{f}{\elementRV, \viewRV}} - 1 \right)^{2} \right) \cdot \indicator{\func{B}{f}}{\tuple{\elementRV}{\viewRV}}} \\
& \eqRes{a}{\le} \expectation{\elementRV \sim \dist}{\left(\left(e^{\abs{\lossS{\elementRV}{\viewRV}}} - 1 \right)^{2} - \left(e^{\func{f}{\elementRV, \viewRV}} - 1 \right)^{2} \right) \cdot \indicator{\func{\bar{B}}{f}}{\tuple{\elementRV}{\viewRV}}} \\
& ~~~~ + \expectation{\elementRV \sim \dist}{\left(\left(e^{\lossS{\elementRV}{\viewRV}} - 1 \right)^{2} - \left(e^{\func{f}{\elementRV, \viewRV}} - 1 \right)^{2} \right) \cdot \indicator{\func{B}{f}}{\tuple{\elementRV}{\viewRV}}} \\
& \eqRes{b}{\le} \expectation{\elementRV \sim \dist}{\left(e^{\lossS{\elementRV}{\viewRV}} - 1 \right)^{2} \cdot \indicator{\func{B}{f}}{\tuple{\elementRV}{\viewRV}}} \\
\end{align*}
where (a) results from the fact that $\left(e^{x} - 1 \right)^{2} \le \left(e^{\abs{x}} - 1 \right)^{2}$ and (b) from removing negative terms.

Using this bound we get
\begin{align*}
\underset{\genfrac{}{}{0pt}{}{\sampleSetRV \sim \iiDist}{\viewRV \sim \mechanismFunc{\sampleSetRV, \analyst}}}{{\text{Pr}}} & \left[\chiSqr{\dist_{\elementsDomain}^{\view}}{\dist_{\elementsDomain}} > \expectation{\elementRV \sim \dist}{ \left(e^{\func{f}{\elementRV, \viewRV}} - 1 \right)^{2}} + \xi \right] \\
& \eqRes{a}{\le} \prob{\genfrac{}{}{0pt}{}{\sampleSetRV \sim \iiDist}{\viewRV \sim \mechanismFunc{\sampleSetRV, \analyst}}}{\expectation{\elementRV \sim \dist}{\left(e^{\lossS{\elementRV}{\viewRV}} - 1 \right)^{2} \cdot \indicator{\func{B}{f}}{\tuple{\elementRV}{\viewRV}}} > \xi} \\
& \eqRes{b}{\le} \frac{1}{\xi} \expectation{\genfrac{}{}{0pt}{}{\elementRV \sim \dist, \sampleSetRV \sim \iiDist}{\viewRV \sim \mechanismFunc{\sampleSetRV, \analyst}}}{\left(e^{\lossS{\elementRV}{\viewRV}} - 1 \right)^{2} \cdot \indicator{\func{B}{f}}{\tuple{\elementRV}{\viewRV}}} \\
& \eqRes{c}{\le} \frac{1}{\xi} \sqrt{\expectation{\genfrac{}{}{0pt}{}{\elementRV \sim \dist, \sampleSetRV \sim \iiDist}{\viewRV \sim \mechanismFunc{\sampleSetRV, \analyst}}}{\left(e^{\lossS{\elementRV}{\viewRV}} - 1 \right)^{4}} \cdot \expectation{\genfrac{}{}{0pt}{}{\elementRV \sim \dist, \sampleSetRV \sim \iiDist}{\viewRV \sim \mechanismFunc{\sampleSetRV, \analyst}}}{\func{\mathbbm{1}_{\func{B}{f}}^{2}}{\tuple{\elementRV}{\viewRV}}}} \\
& \eqRes{d}{\le} \frac{1}{\xi} \sqrt{\left(\expectation{\genfrac{}{}{0pt}{}{\elementRV \sim \dist, \sampleSetRV \sim \iiDist}{\viewRV \sim \mechanismFunc{\sampleSetRV, \analyst}}}{e^{4 \lossS{\elementRV}{\viewRV}}} + 1 \right) \cdot \prob{\genfrac{}{}{0pt}{}{\elementRV \sim \dist, \sampleSetRV \sim \iiDist}{\viewRV \sim \mechanismFunc{\sampleSetRV, \analyst}}}{\tuple{\elementRV}{\viewRV} \in \func{B}{f}}} \\
\end{align*}
where (a) results from the previous bound, (b) from Markov's inequality, (c) from the Cauchy-Schwarz inequality, and (d) from the inequality $\left(x - 1 \right)^{4} \le x^{4} + 1$ for any $x \ge 0$.
\end{proof}

\begin{claim} \label{clm:PCstabSupp_2}
Given $\phi > 0   $, $\delta \in \left(0 , \frac{1}{e} \right)$, and a view $\view \in \viewFamily$, we have
\[
\expectation{\elementRV \sim \dist}{ \left(e^{\PIfuncS{\elementRV}{\view}} - 1 \right)^{2}} \le 3 \left(\expectation{\elementRV \sim \dist}{e^{27  \lnFunc{\frac{1}{\delta}} \left(\PCfuncS{\elementRV}{\view} + \phi \right)}} - 1 \right)
\]
where $\PIfuncS{\element}{\view}$ was defined in Lemma \ref{lem:PChighProb}.
\end{claim}

\begin{proof}
    We first notice that for any $\alpha \ge 1$,
    \begin{equation}
        \expectation{\elementRV \sim \dist}{e^{\alpha \PCfuncS{\elementRV}{\view
        }}} \eqRes{a}{=} \expectation{\elementRV \sim \dist}{\left(\expectation{\elementRVY \sim \dist}{e^{\PCvFunc{\elementRV}{\elementRVY}}} \right)^{\alpha}} \eqRes{b}{\le} \expectation{\elementRV, \elementRVY \sim \dist}{e^{\alpha \PCvFunc{\elementRV}{\elementRVY}}}
    \end{equation}
    where (a) follows from the definition of $\PCfuncS{\element}{\view}$ and (b) from Jensen's inequality for the convex function $x^{\alpha}$.
    
    Next we prove a useful identity. For any $a, b \ge 0$,
    \[
        e^{a+b} - 1 = e^{a} \left(e^{b} - 1 \right) + \left(e^{a} - 1 \right) = e^{a} \sqrt{\left(e^{b} - 1 \right)^{2}} + \sqrt{\left(e^{a} - 1 \right)^{2}} \le	e^{a} \sqrt{\left(e^{\frac{3}{2} b^{2}} - 1 \right)} + \sqrt{\left(e^{2a} - 1 \right)}
    \]
    where the inequality results from the inequalities $\left(e^{x} - 1 \right)^{2} \le e^{\frac{3}{2} x^{2}} - 1$ for any $x \in \reals$ and $\left(e^{x} - 1 \right)^{2} \le e^{2 x} - 1$ for any $x \ge 0$.
    
    Using this bound we get
    \begin{align*}
        \left(e^{a + b} - 1 \right)^{2} & \eqRes{a}{\le} \left(e^{a} \sqrt{\left(e^{\frac{3}{2}b^{2}} - 1 \right)} + \sqrt{\left(e^{2a} - 1 \right)} \right)^{2}
        \\ & =	e^{2a} \left(e^{\frac{3}{2}b^{2}} - 1 \right) + 2e^{a} \sqrt{\left(e^{\frac{3}{2}b^{2}} - 1 \right) \left(e^{2a} - 1 \right)} + \left(e^{2a} - 1 \right)
        \\ & =	\left(e^{2a + \frac{3}{2}b^{2}} - 1 \right) + 2e^{a} \sqrt{\left(e^{\frac{3}{2}b^{2}} - 1 \right) \left(e^{2a} - 1 \right)}
        \\ & \eqRes{b}{\le} \left(e^{2 a + \frac{3}{2} b^{2}} - 1 \right) + 2 \left(e^{3a + \frac{3}{2}b^{2}} - 1 \right)
        \\ & \le 3\left(e^{3 a + \frac{3}{2} b^{2}} - 1 \right)
    \end{align*}
    where (a) results from the previous bound and (b) from the inequalities $\left(e^{x} - 1 \right) \left(e^{y} - 1 \right) \le \left(e^{x + y} - 1 \right)^{2}$ and $e^{x} \left(e^{y} - 1 \right) \le \left(e^{x + y} - 1 \right)$ for any $x, y \ge 0$.
    
    Using this inequality we get
    \begin{align*}
        \underset{\elementRV \sim \dist}{\mathbb{E}} & \left[\left(e^{\PIfuncS{\elementRV}{\view}} - 1 \right)^{2} \right]
        \\ & = \expectation{\elementRV \sim \dist}{ \left(\expFunc{\PCfuncS{\elementRV}{\view} + \phi + 4 \sqrt{\lnFunc{\frac{1}{\delta}} \left(\PCfuncS{\elementRV}{\view} + \phi \right)}} - 1 \right)^{2}}
        \\ & \eqRes{a}{\le} \expectation{\elementRV \sim \dist}{3 \left(\expFunc{3 \left(1 + 8 \lnFunc{\frac{1}{\delta}} \right) \left(\PCfuncS{\elementRV}{\view} + \phi \right)} - 1 \right)}
        \\ & \eqRes{b}{\le} 3 \left(\expectation{\elementRV \sim \dist}{e^{27  \lnFunc{\frac{1}{\delta}} \left(\PCfuncS{\elementRV}{\view} + \phi \right)}} - 1 \right)
    \end{align*}
    where (a) results from the previous inequality and (b) from the bound on $\delta$.
\end{proof}

\begin{proof} [Proof of Theorem \ref{thm:PCstabExt}]
First notice
\begin{align*}
\expectation{\genfrac{}{}{0pt}{}{\elementRV \sim \dist, \sampleSetRV \sim \iiDist}{\viewRV \sim \mechanismFunc{\sampleSetRV, \analyst}}}{e^{4 \lossS{\elementRV}{\viewRV}}} & = \expectation{\elementRV \sim \dist}{\expectation{\viewRV \sim \dist}{\left(\frac{\distFuncDep{\viewRV}{\elementRV}}{\distFunc{\viewRV}} \right)^{4}}} \\
& = \expectation{\elementRV \sim \dist}{\expectation{\viewRV \sim \distFuncDep{\cdot}{\elementRV}}{\left(\frac{\distFuncDep{\viewRV}{\elementRV}}{\distFunc{\viewRV}} \right)^{3}}} \\
& = \expectation{\elementRV \sim \dist}{\expFunc{\func{\mathbf{D}_{4}}{\func{\distInd{\viewFamily}}{\cdot | \elementRV} \Vert \distInd{\viewFamily}}}} \\
& \eqRes{a}{\le} \expectation{\elementRV \sim \dist}{\sqrt{\expectation{\viewRV \sim \dist}{e^{56 \PCfuncS{\elementRV}{\viewRV}}}}} \\
& \eqRes{b}{\le} \sqrt{\expectation{\elementRV \sim \dist, \viewRV \sim \dist}{e^{56 \PCfuncS{\elementRV}{\viewRV}}}} \\
& \eqRes{c}{=} \sqrt{\func{\psi}{56}} \\
\end{align*}
where (a) results from Lemma \ref{lem:PCexp}, (b) from Jensen's inequality for the concave function $\sqrt{x}$, and (c) from the definition of $\psi$.

Using this bound we get
\begin{align*}
\underset{\genfrac{}{}{0pt}{}{\sampleSetRV \sim \iiDist}{\viewRV \sim \mechanismFunc{\sampleSetRV, \analyst}, \queryRV \sim \func{\analyst}{\viewRV}}}{{\text{Pr}}} & \left[\abs{\func{\queryRV}{\dist_{\elementsDomain}^{\view}} - \func{\queryRV}{\dist_{\elementsDomain}}} > \sigma_{\queryRV} \cdot \sqrt{\epsilon^{2} + \xi} \right] \\
& \eqRes{a}{\le} \prob{\genfrac{}{}{0pt}{}{\sampleSetRV \sim \iiDist}{\viewRV \sim \mechanismFunc{\sampleSetRV, \analyst}, \queryRV \sim \func{\analyst}{\viewRV}}}{\chiSqr{\dist_{\elementsDomain}^{\viewRV}}{\dist_{\elementsDomain}} > \epsilon^{2} + \xi} \\
& \eqRes{b}{\le} \prob{\genfrac{}{}{0pt}{}{\sampleSetRV \sim \iiDist}{\viewRV \sim \mechanismFunc{\sampleSetRV, \analyst}, \queryRV \sim \func{\analyst}{\viewRV}}}{\chiSqr{\dist_{\elementsDomain}^{\view}}{\dist_{\elementsDomain}} > \expectation{\elementRV \sim \dist}{ \left(e^{\PIfuncS{\elementRV}{\viewRV}} - 1 \right)^{2}} + \xi} \\
& ~~~~ + \prob{\genfrac{}{}{0pt}{}{\sampleSetRV \sim \iiDist}{\viewRV \sim \mechanismFunc{\sampleSetRV, \analyst}, \queryRV \sim \func{\analyst}{\viewRV}}}{\expectation{\elementRV \sim \dist}{ \left(e^{\PIfuncS{\elementRV}{\viewRV}} - 1 \right)^{2}} > \epsilon^{2}} \\
& \eqRes{c}{\le} \frac{1}{\xi} \sqrt{\left(\expectation{\genfrac{}{}{0pt}{}{\elementRV \sim \dist, \sampleSetRV \sim \iiDist}{\viewRV \sim \mechanismFunc{\sampleSetRV, \analyst}}}{e^{4 \lossS{\elementRV}{\viewRV}}} + 1 \right) \cdot \prob{\genfrac{}{}{0pt}{}{\elementRV \sim \dist, \sampleSetRV \sim \iiDist}{\viewRV \sim \mechanismFunc{\sampleSetRV, \analyst}}}{\abs{\lossS{\elementRV}{\viewRV}} > \PIfuncS{\elementRV}{\viewRV}}} \\
& ~~~~ + \prob{\genfrac{}{}{0pt}{}{\sampleSetRV \sim \iiDist}{\viewRV \sim \mechanismFunc{\sampleSetRV, \analyst}, \queryRV \sim \func{\analyst}{\viewRV}}}{3 \left(\expectation{\elementRV \sim \dist}{e^{27  \lnFunc{\frac{1}{\delta}} \left(\PCfuncS{\elementRV}{\view} + \phi \right)}} - 1 \right) > \epsilon^{2}} \\
& \eqRes{d}{\le} \frac{1}{\xi} \sqrt{\left(\sqrt{\func{\psi}{56}} + 1 \right) \cdot \frac{2^{\frac{1}{4}} \left(e^{\phi} + 1 \right) \delta^{2}}{e^{\phi} - 1} \left(\func{\psi}{12}\right)^{\frac{1}{8}}} \\
& ~~~~ + \prob{\genfrac{}{}{0pt}{}{\sampleSetRV \sim \iiDist}{\viewRV \sim \mechanismFunc{\sampleSetRV, \analyst}, \queryRV \sim \func{\analyst}{\viewRV}}}{\expectation{\elementRV \sim \dist}{e^{27  \lnFunc{\frac{1}{\delta}} \left(\PCfuncS{\elementRV}{\view} + \phi \right)}} > 1 + \frac{\epsilon^{2}}{3}} \\
\end{align*}
where (a) results from Corollary \ref{cor:BayesStabBnd}, (b) from the inequality $\prob{}{a > b} \le \prob{}{a > c \vee c > b} \le \prob{}{a > c} + \prob{}{c > b}$, (c) from Claims \ref{clm:PCstabSupp_1} and \ref{clm:PCstabSupp_1}, and (d) from the previous bound and Lemma \ref{lem:PChighProb}.
\end{proof}

\newpage

\section{Missing parts from Section \ref{sec:varAcc}}
\label{apd:varAcc}
Using Theorem \ref{thm:PCstabExt}, one only has to bound $\expectation{\elementRV \sim \dist}{e^{27  \lnFunc{\frac{1}{\delta}} \left(\PCfuncS{\elementRV}{\view} + \phi \right)}}$ and $\func{\psi}{\alpha}$ to prove Bayes stability. To do so, we first prove a supporting claim and then provide separate bounds for the bounded case (Section \ref{sec:varAccBndQ}) and sub-Gaussian case (Section \ref{sec:varAccSubG}).

\begin{claim} \label{clm:varAccSup}
Given $\numOfIterations \in \naturals$; $\eta > 0$ and a view $\view \in \viewFamily$ consisting of responses to $\numOfIterations$ linear queries, produced by a Gaussian mechanism $\mechanism$ with noise parameter $\eta$ which receives an iid sampled dataset of size $\sampleSize$, then for any $\alpha \ge 1$ we have
\[
\expectation{\elementRV \sim \dist}{e^{\alpha \PCfuncS{\elementRV}{\view}}} \le \expectation{\elementRV, \elementRVY \sim \dist}{e^{\alpha \PCvFunc{\elementRV}{\elementRVY}}} = \expectation{\elementRV, \elementRVY \sim \dist}{\expFunc{\frac{\alpha}{2 \sampleSize^{2} \eta^{2}} \sum_{i = 1}^{\numOfIterations} \left(\func{\query_{i}}{\elementRV} - \func{\query_{i}}{\elementRVY} \right)^{2}}},
\]
where $\PCfuncS{\element}{\view}$ and $\phi$ are defined as in Theorem \ref{thm:PCstab}.
\end{claim}

\begin{proof} Notice that
\begin{align*}
\expectation{\elementRV \sim \dist}{e^{\alpha \PCfuncS{\elementRV}{\view}}} & \eqRes{a}{=} \expectation{\elementRV \sim \dist}{\left(\expectation{\elementRVY \sim \dist}{e^{\PCvFunc{\elementRV}{\elementRVY}}} \right)^{\alpha}} \\
& \eqRes{b}{\le} \expectation{\elementRV, \elementRVY \sim \dist}{e^{\alpha \PCvFunc{\elementRV}{\elementRVY}}} \\
& \eqRes{c}{=} \expectation{\elementRV, \elementRVY \sim \dist}{\expFunc{\alpha \supFunc{\sampleSet \in \elementsDomain^{\sampleSize - 1}}{\frac{\norm{\func{\viewQuery}{\tuple{\sampleSet}{\elementRV}} - \func{\viewQuery}{\tuple{\sampleSet}{\elementRVY}}}^{2}}{2 \eta^{2}}}}} \\
& = \expectation{\elementRV, \elementRVY \sim \dist}{\expFunc{\frac{\alpha}{2 \eta^{2}} \supFunc{\sampleSet \in \elementsDomain^{\sampleSize - 1}}{\sum_{i = 1}^{\numOfIterations} \left(\func{\query_{i}}{\tuple{\sampleSet}{\elementRV}} - \func{\query_{i}}{\tuple{\sampleSet}{\elementRVY}} \right)^{2}}}} \\
& = \expectation{\elementRV, \elementRVY \sim \dist}{\expFunc{\frac{\alpha}{2 \sampleSize^{2} \eta^{2}} \sum_{i = 1}^{\numOfIterations} \left(\func{\query_{i}}{\elementRV} - \func{\query_{i}}{\elementRVY} \right)^{2}}}
\end{align*}
where (a) results from the definition of $\PCfuncS{\element}{\view}$, (b) from Jensen's inequality for the convex function $x^{a}$ for any $a \ge 1$, and (c) from Theorem \ref{lem:GaussPC}.
\end{proof}

\subsection{Generalization guarantees, bounded case} \label{sec:varAccBndQ}
\begin{claim} \label{clm:varAccBndSup}
Given $\numOfIterations \in \naturals$; $\eta, \Delta, \sigma > 0$; $\phi \ge 0$, and a view $\view \in \viewFamily$ consisting of responses to $\numOfIterations$ $\Delta$-bounded linear queries with variance bounded by $\sigma^{2}$, produced by a Gaussian mechanism $\mechanism$ with noise parameter $\eta$ which receives an iid sampled dataset of size $\sampleSize$, then for any $1 \le \alpha \le \frac{1}{\frac{\numOfIterations \Delta^{2}}{2 \sampleSize^{2} \eta^{2}} + \phi}$ we have
\[
\expectation{\elementRV \sim \dist}{e^{\alpha \left(\PCfuncS{\elementRV}{\view} + \phi \right)}} \le \expectation{\elementRV, \elementRVY \sim \dist}{e^{\alpha \left(\PCvFunc{\elementRV}{\elementRVY} + \phi \right)}} \le 1 + 2 \alpha \left(\phi + \frac{\numOfIterations \sigma^{2}}{\sampleSize^{2} \eta^{2}} \right),
\]
where $\PCfuncS{\element}{\view}$ is as was defined in Theorem \ref{thm:PCstab}.
\end{claim}

\begin{proof}
First notice that for any linear query $\query$ with variance $\sigma^{2}$,
\begin{align} \label{eq:varAccSup_2_1}
\expectation{\elementRV, \elementRVY \sim \dist}{\left(\queryFunc{\elementRV} - \queryFunc{\elementRVY} \right)^{2}} & = \expectation{\elementRV, \elementRVY \sim \dist}{\left(\left(\queryFunc{\elementRV} - \queryFunc{\dist} \right) + \left(\queryFunc{\dist} - \queryFunc{\elementRVY} \right) \right)^{2}} \nonumber \\
& \eqRes{a}{=} \expectation{\elementRV \sim \dist}{\left(\queryFunc{\elementRV} - \queryFunc{\dist} \right)^{2}} + \expectation{\elementRVY \sim \dist}{\left(\queryFunc{\dist} - \queryFunc{\elementRVY} \right)^{2}} \\
& ~~~~ + 2 \expectation{\elementRV \sim \dist}{\queryFunc{\elementRV} - \queryFunc{\dist}} \cdot \expectation{\elementRVY \sim \dist}{\queryFunc{\dist} - \queryFunc{\elementRVY}} \nonumber  \\
& \eqRes{b}{=} 2 \sigma^{2} \nonumber 
\end{align}
where (a) results from the fact $\elementRV, \elementRVY$ are independent and (b) from the fact that $\queryFunc{\elementRV}$ is a random variable with expectation $\queryFunc{\dist}$ and variance $\sigma^{2}$.

For any two elements $\element, \elementY \in \elementsDomain$ we have
\begin{equation} \label{eq:varAccSup_2_2}
\PCvFunc{\element}{\elementY} = \frac{1}{2 \sampleSize^{2} \eta^{2}} \sum_{i = 1}^{\numOfIterations} \left(\func{\query_{i}}{\elementRV} - \func{\query_{i}}{\elementRVY} \right)^{2} \le \frac{\numOfIterations \Delta^{2}}{2 \sampleSize^{2} \eta^{2}} .
\end{equation}

Using this bound we get
\begin{align*}
\expectation{\elementRV \sim \dist}{e^{\alpha \left(\PCfuncS{\elementRV}{\view} + \phi \right)}} & \eqRes{a}{\le} \expectation{\elementRV, \elementRVY \sim \dist}{e^{\alpha \left(\PCvFunc{\elementRV}{\elementRVY} + \phi \right)}} \nonumber \\
& = \expectation{\elementRV, \elementRVY \sim \dist}{\expFunc{\alpha \left(\phi + \frac{1}{2 \sampleSize^{2} \eta^{2}} \sum_{i = 1}^{\numOfIterations} \left(\func{\query_{i}}{\elementRV} - \func{\query_{i}}{\elementRVY} \right)^{2} \right)}} \\
& \eqRes{b}{\le} \expectation{\elementRV, \elementRVY \sim \dist}{1 + 2 \alpha \left(\phi + \frac{1}{2 \sampleSize^{2} \eta^{2}} \sum_{i = 1}^{\numOfIterations} \left(\func{\query_{i}}{\elementRV} - \func{\query_{i}}{\elementRVY} \right)^{2} \right)} \\
& = 1 + \alpha \left(2 \phi + \frac{1}{\sampleSize^{2} \eta^{2}} \sum_{i = 1}^{\numOfIterations} \expectation{\elementRV, \elementRVY \sim \dist}{\left(\func{\query_{i}}{\elementRV} - \func{\query_{i}}{\elementRVY} \right)^{2}} \right) \\
& \eqRes{c}{\le} 1 + \alpha \left(2 \phi + \frac{1}{\sampleSize^{2} \eta^{2}} \sum_{i = 1}^{\numOfIterations} 2 \sigma^{2} \right) \\
& = 1 + 2 \alpha \left(\phi + \frac{\numOfIterations \sigma^{2}}{\sampleSize^{2} \eta^{2}} \right)
\end{align*}
where (a) results from combining the assumption $\alpha \ge 1$ with Claim \ref{clm:varAccSup}, (b) from combining Equation \ref{eq:varAccSup_2_2}, the assumption $\alpha \le \frac{1}{\frac{\numOfIterations \Delta^{2}}{2 \sampleSize^{2} \eta^{2}} + \phi}$, and the inequality $e^{x} \le 1 + 2 x$ for any $0 \le x \le 1$, and (c) from Equation \ref{eq:varAccSup_2_1}.
\end{proof}

\begin{lemma}  \label{lem:GaussMechBayesStabBnd}
Given $\numOfIterations \in \naturals$; $\Delta, \sigma, \eta \ge 0$; $0 < \delta \le \frac{1}{e}$; and an analyst $\analyst$ issuing $\numOfIterations$ $\Delta$-bounded linear queries with variance bounded by $\sigma^{2}$, if $\mechanism$ is a Gaussian mechanism with noise parameter $\eta$ that receives an iid dataset of size $\sampleSize \ge \frac{9 \sqrt{\numOfIterations} \Delta}{\eta} \sqrt{\lnFunc{\frac{1}{\delta}}}$, 
then $\mechanism$ is $\tuple{\frac{\sigma^{2}}{\sampleSize \eta} \sqrt{488 \numOfIterations \lnFunc{\frac{1}{\delta}}}}{\frac{2 \sampleSize^{3} \eta^{3} \delta}{\numOfIterations^{\frac{3}{2}} \sigma^{3}}}$-Bayes stable.
\end{lemma}

\begin{proof}
Setting $\phi \coloneqq \frac{2 \numOfIterations \sigma^{2}}{\sampleSize^{2} \eta^{2}}$ we get 
\[
27 \lnFunc{\frac{1}{\delta}} \eqRes{a}{<} \frac{2 \sampleSize^{2} \eta^{2}}{5 \numOfIterations \Delta^{2}} = \frac{1}{\frac{5 \numOfIterations \Delta^{2}}{2 \sampleSize^{2} \eta^{2}}} \eqRes{b}{<} \frac{1}{\frac{\numOfIterations \Delta^{2}}{2 \sampleSize^{2} \eta^{2}} + \phi}
\]
where (a) results from the assumption $\sampleSize \ge \frac{9 \sqrt{\numOfIterations} \Delta}{\eta} \sqrt{\lnFunc{\frac{1}{\delta}}}$ and (b) from the definition of $\phi$.

Using this bound we can invoke Claim \ref{clm:varAccBndSup}, and get that for any view $\view \in \viewFamily$,
\begin{equation} \label{eq:GaussMechBayesStabBnd_1}
\expectation{\elementRV \sim \dist}{e^{27 \lnFunc{\frac{1}{\delta}} \left(\PCfuncS{\elementRV}{\view} + \phi \right)}} \le 1 + 54 \left(\frac{\numOfIterations \sigma^{2}}{\sampleSize^{2} \eta^{2}} + \phi \right) \lnFunc{\frac{1}{\delta}}.
\end{equation}

Next, we notice that
\begin{align} \label{eq:GaussMechBayesStabBnd_2}
\expectation{\elementRV \sim \dist, \viewRV \sim \dist}{\PCfuncS{\elementRV}{\viewRV}} & = \expectation{\elementRV \sim \dist, \viewRV \sim \dist}{\lnFunc{\expectation{\elementRVY \sim \dist}{e^{\PCfuncL{\elementRV}{\elementRVY}{\viewRV}}}}} \nonumber \\
& \eqRes{a}{\le} \lnFunc{\expectation{\elementRV, \elementRVY \sim \dist, \viewRV \sim \dist}{e^{\PCfuncL{\elementRV}{\elementRVY}{\viewRV}}}} \nonumber \\
& \eqRes{b}{\le} \lnFunc{1 + \frac{2\numOfIterations \sigma^{2}}{\sampleSize^{2} \eta^{2}}} \\
& \eqRes{c}{\le} \frac{2\numOfIterations \sigma^{2}}{\sampleSize^{2} \eta^{2}} \nonumber \\
& \eqRes{d}{=} \phi \nonumber
\end{align}
where (a) results from Jensen's inequality for the concave function $\lnFunc{x}$, (b) from Claim \ref{clm:varAccBndSup} where the condition on $\alpha$ is fulfilled by the assumption that $\sampleSize \ge \frac{9 \sqrt{\numOfIterations} \Delta}{\eta} \sqrt{\lnFunc{\frac{1}{\delta}}}$, (c) from the inequality $\lnFunc{1 + x} \le x$ for any $x \ge 0$, and (d) from the definition of $\phi$.

Similarly, for any $\alpha \le \frac{2 \sampleSize^{2} \eta^{2}}{\numOfIterations \Delta^{2}}$ we have
\begin{equation} \label{eq:GaussMechBayesStabBnd_3}
\func{\psi}{\alpha} \eqRes{a}{\le} 1 + \frac{2 \alpha \numOfIterations \sigma^{2}}{\sampleSize^{2} \eta^{2}} \eqRes{b}{\le} 1 + \frac{\alpha \numOfIterations \Delta^{2}}{2 \sampleSize^{2} \eta^{2}} \eqRes{c}{\le} 1 + \frac{\alpha \numOfIterations \Delta^{2}}{2 \sampleSize^{2} \eta^{2}} \eqRes{d}{\le} 1 + \frac{\alpha}{162}
\end{equation}
where (a) results from the assumption on $\alpha$ and Claim \ref{clm:varAccBndSup}, (b) from the fact that for any distribution, $\sigma^{2} \le \frac{\Delta^{2}}{4}$, (c) from the assumption that $\delta \le \frac{1}{e}$, and (d) from the assumption that $\sampleSize \ge \frac{9 \sqrt{\numOfIterations} \Delta}{\eta} \sqrt{\lnFunc{\frac{1}{\delta}}}$.

Setting $\xi \coloneqq \frac{2 \numOfIterations \sigma^{2}}{\sampleSize^{2} \eta^{2}}$ we get
\begin{align*}
\underset{\genfrac{}{}{0pt}{}{\sampleSetRV \sim \iiDist}{\viewRV \sim \mechanismFunc{\sampleSetRV, \analyst}, \queryRV \sim \func{\analyst}{\viewRV}}}{\text{Pr}} & \left[\abs{\func{\queryRV}{\dist_{\elementsDomain}^{\view}} - \func{\queryRV}{\dist_{\elementsDomain}}} > \frac{\sigma^{2}}{\sampleSize \eta} \sqrt{488 \numOfIterations \lnFunc{\frac{1}{\delta}}} \right]
\\ & \eqRes{a}{\le} \prob{\genfrac{}{}{0pt}{}{\sampleSetRV \sim \iiDist}{\viewRV \sim \mechanismFunc{\sampleSetRV, \analyst}, \queryRV \sim \func{\analyst}{\viewRV}}}{\abs{\func{\queryRV}{\dist_{\elementsDomain}^{\view}} - \func{\queryRV}{\dist_{\elementsDomain}}} > \sigma_{\queryRV} \sqrt{\frac{488 \numOfIterations \sigma^{2}}{\sampleSize^{2} \eta^{2}} \lnFunc{\frac{1}{\delta}}}}
\\ & \eqRes{b}{\le} \prob{\genfrac{}{}{0pt}{}{\sampleSetRV \sim \iiDist}{\viewRV \sim \mechanismFunc{\sampleSetRV, \analyst}, \queryRV \sim \func{\analyst}{\viewRV}}}{\abs{\func{\queryRV}{\dist_{\elementsDomain}^{\view}} - \func{\queryRV}{\dist_{\elementsDomain}}} > \sigma_{\queryRV} \sqrt{162 \left(\frac{\numOfIterations \sigma^{2}}{\sampleSize^{2} \eta^{2}} + \frac{2 \numOfIterations \sigma^{2}}{\sampleSize^{2} \eta^{2}} \right) \lnFunc{\frac{1}{\delta}} + \frac{2 \numOfIterations \sigma^{2}}{\sampleSize^{2} \eta^{2}}}}
\\ & \eqRes{c}{=} \prob{\genfrac{}{}{0pt}{}{\sampleSetRV \sim \iiDist}{\viewRV \sim \mechanismFunc{\sampleSetRV, \analyst}, \queryRV \sim \func{\analyst}{\viewRV}}}{\abs{\func{\queryRV}{\dist_{\elementsDomain}^{\view}} - \func{\queryRV}{\dist_{\elementsDomain}}} > \sigma_{\queryRV} \sqrt{162 \left(\frac{\numOfIterations \sigma^{2}}{\sampleSize^{2} \eta^{2}} + \phi \right) \lnFunc{\frac{1}{\delta}} + \xi}}
\\ & \eqRes{d}{\le} \prob{\genfrac{}{}{0pt}{}{\sampleSetRV \sim \iiDist}{\viewRV \sim \mechanismFunc{\sampleSetRV, \analyst}, \queryRV \sim \func{\analyst}{\viewRV}}}{\expectation{\elementRV \sim \dist}{e^{27  \lnFunc{\frac{1}{\delta}} \left(\PCfuncS{\elementRV}{\view} + \phi \right)}} > 1 + 54 \left(\frac{\numOfIterations \sigma^{2}}{\sampleSize^{2} \eta^{2}} + \phi \right) \lnFunc{\frac{1}{\delta}}}
\\ & ~~~~ + \frac{\delta}{\xi} \left(4 \func{\psi}{12} \right)^{\frac{1}{16}} \sqrt{\left(\sqrt{\func{\psi}{56}} + 1 \right) \frac{e^{\phi} + 1}{e^{\phi} - 1}}
\\ & \eqRes{e}{\le} \frac{\delta}{\xi} \left(4 \func{\psi}{12} \right)^{\frac{1}{16}} \sqrt{\left(\sqrt{\func{\psi}{56}} + 1 \right) \frac{e^{\phi} + 1}{e^{\phi} - 1}}
\\ & \eqRes{f}{\le} \frac{\delta}{\xi} \left(\frac{116}{27} \right)^{\frac{1}{16}} \sqrt{\left(\sqrt{\frac{109}{81}} + 1 \right) \frac{3}{\phi}}
\\ & \le \frac{3 \delta}{\xi \sqrt{\phi}} 
\\ & \eqRes{c}{=} \frac{2 \sampleSize^{3} \eta^{3} \delta}{\numOfIterations^{\frac{3}{2}} \sigma^{3}}
\end{align*}
where (a) results from fact that the variance of all queries issued by $\analyst$ is bounded by $\sigma^{2}$, (b) from the assumption that $\delta \le \frac{1}{e}$, (c) from the definition of $\phi$ and $\xi$, (d) from Theorem \ref{thm:PCstabExt} where the condition was proven in Equation \ref{eq:GaussMechBayesStabBnd_2}, (e) from Equation \ref{eq:GaussMechBayesStabBnd_1}, and (f) from Equation \ref{eq:GaussMechBayesStabBnd_3}, and the fact that $\frac{e^{x} + 1}{e^{x} - 1} \le \frac{3}{x}$ for any $0 < x < 2$.
\end{proof}

\begin{theorem} [Exact version of Theorem \ref{thm:GaussMechAccBnd}] \label{thm:GaussMechAccBndExt}
Given $\numOfIterations \in \naturals$; $\Delta, \sigma, \epsilon \ge 0$; $0 < \delta \le \frac{1}{e}$; and an analyst $\analyst$ issuing $\numOfIterations$ $\Delta$-bounded linear queries with variance bounded by $\sigma^{2}$, if $\mechanism$ is a Gaussian mechanism with noise parameter $\eta = \left(122 \right)^{\frac{1}{4}} \sigma \sqrt{\frac{\sqrt{\numOfIterations}}{\sampleSize}}$ that receives an iid dataset of size $\sampleSize \ge \max \left\{\frac{9 \Delta}{\epsilon}, \frac{89 \sigma^{2}}{\epsilon^{2}} \right\} \sqrt{\numOfIterations} \lnFunc{\frac{8 \numOfIterations}{\delta}}$, and $\numOfIterations \ge 2 \sampleSize$, then $\mechanism$ is $\tuple{\epsilon}{\delta}$-distribution accurate. 
\end{theorem}

\begin{proof}
From Lemma \ref{lem:GaussMechBayesStabBnd} setting $\delta' \coloneqq \frac{\delta}{8 \numOfIterations}$ (where $\delta'$ denotes the parameter $\delta$ used in the Lemma) we get that $\mechanism$ is $\tuple{\frac{\sigma^{2}}{\sampleSize \eta} \sqrt{488 \numOfIterations \lnFunc{\frac{4 \numOfIterations}{\delta}}}}{\frac{\sampleSize^{3} \eta^{3} \delta}{\numOfIterations^{\frac{5}{2}} \sigma^{3}}}$-Bayes stable.
From Lemma \ref{lem:GaussAcc}, $\mechanism$ is $\tuple{2 \eta \sqrt{\lnFunc{\frac{4 \numOfIterations}{\delta}}}}{\frac{\delta}{2}}$-posterior accurate.

Optimizing over $\eta$ we get that $\eta = \left(122 \right)^{\frac{1}{4}} \sigma \sqrt{\frac{\sqrt{\numOfIterations}}{\sampleSize}}$.
Combining this with the bound on $\numOfIterations$ implies $\frac{\sampleSize^{3} \eta^{3}}{\numOfIterations^{\frac{3}{2}} \sigma^{3}} = \frac{3 \left(122 \right)^{\frac{1}{4}} \sampleSize^{\frac{3}{2}}}{\sqrt{2} \numOfIterations^{\frac{3}{4}}} \le 2 \numOfIterations^{\frac{3}{4}} \le 2 \numOfIterations$. 
Plugging the definition of $\eta$ back in we get that $\mechanism$ is $\tuple{\sigma \sqrt{\frac{\sqrt{488 \numOfIterations}}{\sampleSize} \lnFunc{\frac{4 \numOfIterations}{\delta}}}}{\frac{\delta}{2}}$-Bayes stable and posterior accurate. 

Using the bound on $\sampleSize$ we get that $\mechanism$ is $\tuple{\frac{\epsilon}{2}}{\frac{\delta}{2}}$-Bayes stable and posterior accurate. 
Combining this with Theorem \ref{thm:BayesStabImpdistPcc} completes the proof.
\end{proof}

\subsection{Generalization guarantees, sub-Gaussian case} \label{sec:varAccSubG}
To prove the generalization guarantees in the sub-Gaussian case, we must first introduce the notion of sub-exponential random variables, and recall several folklore facts about sub-exponential random variables.

\begin{definition}
    Given $\sigma, \alpha \ge 0$, a random variable $X$ is called $\tuple{\alpha}{\sigma^{2}}$-sub-exponential if for all $\lambda \in \left[-\frac{1}{\alpha}, \frac{1}{\alpha}\right]$ we have $\expectation{}{e^{\lambda X}} \le e^{\frac{\lambda^{2} \sigma^{2}}{2}}$.
\end{definition}

\begin{fact}\label{fct:subGaussToSubExp}
    Given $\sigma \ge 0$ and a $\sigma^{2}$-sub-Gaussian random variable $X$, the random variable $Y \coloneqq X^{2} - \expectation{}{X'^{2}}$ is $\tuple{4 \sigma^{2}}{64 \sigma^{4}}$-sub-exponential.
\end{fact}

\begin{fact} \label{fct:sumSubGauss}
    Given $\sigma_{i},  \ge 0$ and independent $\sigma_{i}^{2}$-sub-Gaussian random variables $X_{i}$ for $i \in \left\{0, 1 \right\}$, the random variable $Y \coloneqq X_{0} + X_{1}$ is $\tuple{0}{\sigma_{0}^{2} + \sigma_{1}^{2}}$-sub-Gaussian.
\end{fact}

\begin{claim} \label{clm:varAccSubGaussSup}
    Given $\numOfIterations \in \naturals$; $\eta, \sigma > 0$ and a view $\view \in \viewFamily$ consisting of responses to $\numOfIterations$ $\sigma^{2}$-sub-Gaussian linear queries, produced by a Gaussian mechanism $\mechanism$ with noise parameter $\eta$ which receives an iid sampled dataset of size $\sampleSize$, then for any $1 \le \alpha \le \frac{\sampleSize^{2} \eta^{2}}{4 \numOfIterations \sigma^{2}}$ we have
    \[
        \expectation{\elementRV \sim \dist}{e^{\alpha \PCfuncS{\elementRV}{\view}}} \le \expectation{\elementRV, \elementRVY \sim \dist}{e^{\alpha \PCvFunc{\elementRV}{\elementRVY}}} \le \expFunc{\frac{9 \alpha \numOfIterations \sigma^{2}}{\sampleSize^{2} \eta^{2}}},
    \]
    where $\PCfuncS{\element}{\view}$ is as was defined in Theorem \ref{thm:PCstab}.
\end{claim}

\begin{proof}
    From Fact \ref{fct:sumSubGauss} and the assumption, the random variable $\queryFunc{\elementRV} - \queryFunc{\elementRVY}$ is a $\tuple{0}{2 \sigma^{2}}$-sub-Gaussian random variable. Combining this with Fact \ref{fct:subGaussToSubExp} we get that the random variable $\left(\queryFunc{\elementRV} - \queryFunc{\elementRVY} \right)^{2} - 2 \sigma^{2}$ is a $\tuple{8 \sigma^{2}}{256 \sigma^{4}}$-sub-exponential random variable, where we rely on the fact that
    \begin{align*}
        \expectation{\elementRV, \elementRVY \sim \dist}{\left(\queryFunc{\elementRV} - \queryFunc{\elementRVY} \right)^{2}} & = \expectation{\elementRV, \elementRVY \sim \dist}{\left(\left(\queryFunc{\elementRV} - \queryFunc{\dist} \right) + \left(\queryFunc{\dist} - \queryFunc{\elementRVY} \right) \right)^{2}}
        \\ & \eqRes{a}{=} \expectation{\elementRV \sim \dist}{\left(\queryFunc{\elementRV} - \queryFunc{\dist} \right)^{2}} + \expectation{\elementRVY \sim \dist}{\left(\queryFunc{\dist} - \queryFunc{\elementRVY} \right)^{2}}
        \\ & ~~~~ + 2 \expectation{\elementRV \sim \dist}{\queryFunc{\elementRV} - \queryFunc{\dist}} \cdot \expectation{\elementRVY \sim \dist}{\queryFunc{\dist} - \queryFunc{\elementRVY}}
        \\ & \eqRes{b}{\le} 2 \sigma^{2}
    \end{align*}
    where (a) results from the fact $\elementRV, \elementRVY$ are independent and (b) from the fact that $\queryFunc{\elementRV}$ is a random variable with expectation $\queryFunc{\dist}$ and variance proxy $\sigma^{2}$ which upper bounds its variance.
    
    Using this we get
    \begin{align*}
        \expectation{\elementRV \sim \dist}{e^{\alpha \PCfuncS{\elementRV}{\view}}} & \eqRes{a}{\le} \expectation{\elementRV, \elementRVY \sim \dist}{e^{\alpha \PCvFunc{\elementRV}{\elementRVY}}} \nonumber
        \\ & = \expectation{\elementRV, \elementRVY \sim \dist}{\expFunc{\frac{\alpha}{2 \sampleSize^{2} \eta^{2}} \sum_{i = 1}^{\numOfIterations} \left(\func{\query_{i}}{\elementRV} - \func{\query_{i}}{\elementRVY} \right)^{2}}}
        \\ & = e^{\frac{\alpha \numOfIterations \sigma^{2}}{\sampleSize^{2} \eta^{2}}} \expectation{\elementRV, \elementRVY \sim \dist}{\expFunc{\frac{\alpha}{2 \sampleSize^{2} \eta^{2}} \sum_{i = 1}^{\numOfIterations} \left(\left(\func{\query_{i}}{\elementRV} - \func{\query_{i}}{\elementRVY} \right)^{2} - 2 \sigma^{2} \right)}}
        \\ & = e^{\frac{\alpha \numOfIterations \sigma^{2}}{\sampleSize^{2} \eta^{2}}} \expectation{\elementRV, \elementRVY \sim \dist}{\prod_{i = 1}^{\numOfIterations} \expFunc{\frac{\alpha}{2 \sampleSize^{2} \eta^{2}} \left(\left(\func{\query_{i}}{\elementRV} - \func{\query_{i}}{\elementRVY} \right)^{2} - 2 \sigma^{2} \right)}}
        \\ & \eqRes{b}{\le} e^{\frac{\alpha \numOfIterations \sigma^{2}}{\sampleSize^{2} \eta^{2}}} \prod_{i = 1}^{\numOfIterations} \left(\expectation{\elementRV, \elementRVY \sim \dist}{\expFunc{\frac{\alpha \numOfIterations}{2 \sampleSize^{2} \eta^{2}} \left(\left(\func{\query_{i}}{\elementRV} - \func{\query_{i}}{\elementRVY} \right)^{2} - 2 \sigma^{2} \right)}} \right)^{\frac{1}{\numOfIterations}}
        \\ & \eqRes{c}{\le} e^{\frac{\alpha \numOfIterations \sigma^{2}}{\sampleSize^{2} \eta^{2}}} \prod_{i = 1}^{\numOfIterations} \left(\expFunc{\frac{32 \alpha^{2} \numOfIterations^{2} \sigma^{4}}{\sampleSize^{4} \eta^{4}}} \right)^{\frac{1}{\numOfIterations}}
        \\ & \eqRes{d}{\le} e^{\frac{\alpha \numOfIterations \sigma^{2}}{\sampleSize^{2} \eta^{2}}} \prod_{i = 1}^{\numOfIterations} \left(\expFunc{\frac{8 \alpha \numOfIterations \sigma^{2}}{\sampleSize^{2} \eta^{2}}} \right)^{\frac{1}{\numOfIterations}}
        \\ & = \expFunc{\frac{9 \alpha \numOfIterations \sigma^{2}}{\sampleSize^{2} \eta^{2}}}
    \end{align*}
    where (a) results from combining the assumption $\alpha \ge 1$ with the Claim \ref{clm:varAccSup}, (b) from the generalized version of H\"{o}lder's inequality, (c) from the fact $\left(\queryFunc{\elementRV} - \queryFunc{\elementRVY} \right)^{2} - 2 \sigma^{2}$ is a $\tuple{8 \sigma^{2}}{256 \sigma^{4}}$-sub-exponential random variable and the assumption $\alpha \le \frac{\sampleSize^{2} \eta^{2}}{4 \numOfIterations \sigma^{2}}$, and (d) from the the assumption $\alpha \le \frac{\sampleSize^{2} \eta^{2}}{4 \numOfIterations \sigma^{2}}$.
\end{proof}

\begin{lemma} \label{lem:GaussMechBayesStabSubGauss}
Given $\numOfIterations \in \naturals$; $\sigma, \eta > 0$; $0 < \delta \le \frac{1}{e}$; and an analyst $\analyst$ issuing $\numOfIterations$ $\sigma^{2}$-sub-Gaussian linear queries, if $\mechanism$ is a Gaussian mechanism with noise parameter $\eta$ that receives an iid dataset of size $\sampleSize \ge \frac{23 \sqrt{\numOfIterations} \sigma}{\eta} \sqrt{\lnFunc{\frac{1}{\delta}}}$, 
then $\mechanism$ is $\tuple{\frac{\sigma^{2}}{\sampleSize \eta} \sqrt{976 \numOfIterations \lnFunc{\frac{1}{\delta}}}}{\frac{\sampleSize^{3} \eta^{3} \delta}{\numOfIterations^{\frac{3}{2}} \sigma^{3}}}$-Bayes stable.
\end{lemma}

\begin{proof}
Setting $\phi \coloneqq \frac{9 \numOfIterations \sigma^{2}}{\sampleSize^{2} \eta^{2}}$ we get
\begin{align} \label{eq:GaussMechBayesStabSubGauss_1}
\expectation{\elementRV \sim \dist}{e^{27 \lnFunc{\frac{1}{\delta}} \left(\PCfuncS{\elementRV}{\view} + \phi \right)}} & \eqRes{a}{=} \expFunc{\frac{243 \numOfIterations \sigma^{2}}{\sampleSize^{2} \eta^{2}} \lnFunc{\frac{1}{\delta}}} \expectation{\elementRV \sim \dist}{e^{27 \lnFunc{\frac{1}{\delta}} \PCfuncS{\elementRV}{\view}}} \nonumber
\\ & \eqRes{b}{\le} \expFunc{\frac{243 \numOfIterations \sigma^{2}}{\sampleSize^{2} \eta^{2}} \lnFunc{\frac{1}{\delta}} + \frac{243 \numOfIterations \sigma^{2}}{\sampleSize^{2} \eta^{2}} \lnFunc{\frac{1}{\delta}}}
\\ & = \expFunc{\frac{486 \numOfIterations \sigma^{2}}{\sampleSize^{2} \eta^{2}} \lnFunc{\frac{1}{\delta}}} \nonumber
\\ & \eqRes{c}{\le} 1 + \frac{972 \numOfIterations \sigma^{2}}{\sampleSize^{2} \eta^{2}} \lnFunc{\frac{1}{\delta}} \nonumber
\end{align}
where (a) results from the definition of $\phi$, (b) from Claim \ref{clm:varAccSubGaussSup} where the condition on $\alpha$ is fulfilled by the assumption that $\sampleSize \ge \frac{23 \sqrt{\numOfIterations} \sigma}{\eta} \sqrt{\lnFunc{\frac{1}{\delta}}}$, and (c)  from the inequality $e^{x} \le 1 + 2x$ for any $x \le 1$ where the condition on $x$ is fulfilled by the assumption that $\sampleSize \ge \frac{23 \sqrt{\numOfIterations} \sigma}{\eta} \sqrt{\lnFunc{\frac{1}{\delta}}}$.

Next, we notice that
\begin{align} \label{eq:GaussMechBayesStabSubGauss_2}
\expectation{\elementRV \sim \dist, \viewRV \sim \dist}{\PCfuncS{\elementRV}{\viewRV}} & = \expectation{\elementRV \sim \dist, \viewRV \sim \dist}{\lnFunc{\expectation{\elementRVY \sim \dist}{e^{\PCfuncL{\elementRV}{\elementRVY}{\viewRV}}}}} \nonumber \\
& \eqRes{a}{\le} \lnFunc{\expectation{\elementRV, \elementRVY \sim \dist, \viewRV \sim \dist}{e^{\PCfuncL{\elementRV}{\elementRVY}{\viewRV}}}} \nonumber \\
& \eqRes{b}{\le} \lnFunc{\expFunc{\frac{9 \numOfIterations \sigma^{2}}{\sampleSize^{2} \eta^{2}}}} \\
& = \frac{9 \numOfIterations \sigma^{2}}{\sampleSize^{2} \eta^{2}} \nonumber \\
& \eqRes{d}{=} \phi \nonumber,
\end{align}
where (a) results from Jensen's inequality for the concave function $\lnFunc{x}$, (b) from the assumption $\sampleSize \ge \frac{23 \sqrt{\numOfIterations} \sigma}{\eta} \sqrt{\lnFunc{\frac{1}{\delta}}}$ and Claim \ref{clm:varAccSubGaussSup}, and (c) from the definition of $\phi$.

Similarly, for any $\alpha \le \frac{\sampleSize^{2} \eta^{2}}{4 \numOfIterations \sigma^{2}}$ we have
\begin{equation} \label{eq:GaussMechBayesStabSubGauss_3}
\func{\psi}{\alpha} \eqRes{a}{\le} \expFunc{\frac{9 \alpha \numOfIterations \sigma^{2}}{\sampleSize^{2} \eta^{2}}} \eqRes{b}{\le}\expFunc{\frac{9 \alpha}{529}},
\end{equation}
where (a) results from the assumption on $\alpha$ and Claim \ref{clm:varAccSubGaussSup} and (b) from the fact that $\sampleSize \ge \frac{23 \sqrt{\numOfIterations} \sigma}{\eta} \sqrt{\lnFunc{\frac{1}{\delta}}}$.

Setting $\xi \coloneqq \frac{6 \numOfIterations \sigma^{2}}{\sampleSize^{2} \eta^{2}}$ we get
\begin{align*}
\underset{\genfrac{}{}{0pt}{}{\sampleSetRV \sim \iiDist}{\viewRV \sim \mechanismFunc{\sampleSetRV, \analyst}, \queryRV \sim \func{\analyst}{\viewRV}}}{\text{Pr}} & \left[\abs{\func{\queryRV}{\dist_{\elementsDomain}^{\view}} - \func{\queryRV}{\dist_{\elementsDomain}}} > \frac{\sigma^{2}}{\sampleSize \eta} \sqrt{976 \numOfIterations \lnFunc{\frac{1}{\delta}}} \right]
\\ & \eqRes{a}{\le} \prob{\genfrac{}{}{0pt}{}{\sampleSetRV \sim \iiDist}{\viewRV \sim \mechanismFunc{\sampleSetRV, \analyst}, \queryRV \sim \func{\analyst}{\viewRV}}}{\abs{\func{\queryRV}{\dist_{\elementsDomain}^{\view}} - \func{\queryRV}{\dist_{\elementsDomain}}} > \sigma_{\queryRV} \sqrt{\frac{976 \numOfIterations \sigma^{2}}{\sampleSize^{2} \eta^{2}} \lnFunc{\frac{1}{\delta}}}}
\\ & \eqRes{b}{\le} \prob{\genfrac{}{}{0pt}{}{\sampleSetRV \sim \iiDist}{\viewRV \sim \mechanismFunc{\sampleSetRV, \analyst}, \queryRV \sim \func{\analyst}{\viewRV}}}{\abs{\func{\queryRV}{\dist_{\elementsDomain}^{\view}} - \func{\queryRV}{\dist_{\elementsDomain}}} > \sigma_{\queryRV} \sqrt{162 \left(\frac{9 \numOfIterations \sigma^{2}}{\sampleSize^{2} \eta^{2}} + \frac{9 \numOfIterations \sigma^{2}}{\sampleSize^{2} \eta^{2}} \right) \lnFunc{\frac{1}{\delta}} + \frac{6 \numOfIterations \sigma^{2}}{\sampleSize^{2} \eta^{2}}}}
\\ & \eqRes{c}{=} \prob{\genfrac{}{}{0pt}{}{\sampleSetRV \sim \iiDist}{\viewRV \sim \mechanismFunc{\sampleSetRV, \analyst}, \queryRV \sim \func{\analyst}{\viewRV}}}{\abs{\func{\queryRV}{\dist_{\elementsDomain}^{\view}} - \func{\queryRV}{\dist_{\elementsDomain}}} > \sigma_{\queryRV} \sqrt{162 \left(\frac{9 \numOfIterations \sigma^{2}}{\sampleSize^{2} \eta^{2}} + \phi \right) \lnFunc{\frac{1}{\delta}} + \xi}}
\\ & \eqRes{d}{\le} \prob{\genfrac{}{}{0pt}{}{\sampleSetRV \sim \iiDist}{\viewRV \sim \mechanismFunc{\sampleSetRV, \analyst}, \queryRV \sim \func{\analyst}{\viewRV}}}{\expectation{\elementRV \sim \dist}{e^{27  \lnFunc{\frac{1}{\delta}} \left(\PCfuncS{\elementRV}{\view} + \phi \right)}} > 1 + 54 \left(\frac{9 \numOfIterations \sigma^{2}}{\sampleSize^{2} \eta^{2}} + \phi \right) \lnFunc{\frac{1}{\delta}}}
\\ & ~~~~ + \frac{\delta}{\xi} \left(4 \func{\psi}{12} \right)^{\frac{1}{16}} \sqrt{\left(\sqrt{\func{\psi}{56}} + 1 \right) \frac{e^{\phi} + 1}{e^{\phi} - 1}}
\\ & \eqRes{e}{\le} \frac{\delta}{\xi} \left(4 \func{\psi}{12} \right)^{\frac{1}{16}} \sqrt{\left(\sqrt{\func{\psi}{56}} + 1 \right) \frac{e^{\phi} + 1}{e^{\phi} - 1}}
\\ & \eqRes{f}{\le} \frac{\delta}{\xi} e^{\frac{27}{2116}} \sqrt{\left(e^{\frac{252}{529}} + 1 \right) \frac{3}{\phi}}
\\ & \le \frac{\delta}{\xi \sqrt{\phi}} 
\\ & \eqRes{c}{=} \frac{\sampleSize^{3} \eta^{3} \delta}{\numOfIterations^{\frac{3}{2}} \sigma^{3}}
\end{align*}
where (a) results from fact that the variance of all queries issued by $\analyst$ is bounded by $\sigma^{2}$, (b) from the assumption that $\delta \le \frac{1}{e}$, (c) from the definition of $\phi$ and $\xi$, (d) from Theorem \ref{thm:PCstabExt} where the condition was proven in Equation \ref{eq:GaussMechBayesStabSubGauss_2}, (e) from Equation \ref{eq:GaussMechBayesStabSubGauss_1}, and (f) from Equation \ref{eq:GaussMechBayesStabSubGauss_3}, and the fact that $\frac{e^{x} + 1}{e^{x} - 1} \le \frac{3}{x}$ for any $0 < x < 2$.
\end{proof}

\begin{theorem} [Exact version of Theorem \ref{thm:GaussMechAccSubGauss}] \label{thm:GaussMechAccSubGaussExt}
Given $\numOfIterations \in \naturals$; $\sigma, \epsilon \ge 0$; $0 < \delta \le \frac{1}{e}$; and an analyst $\analyst$ issuing $\numOfIterations$ $\sigma^{2}$-sub-Gaussian linear queries, if $\mechanism$ is a Gaussian mechanism with noise parameter $\eta = \left(244 \right)^{\frac{1}{4}} \sigma \sqrt{\frac{\sqrt{\numOfIterations}}{\sampleSize}}$ that receives an iid dataset of size $\sampleSize \ge \frac{32 \sqrt{\numOfIterations} \sigma^{2}}{\epsilon^{2}} \lnFunc{\frac{2 \numOfIterations}{\delta}}$, and $\numOfIterations \ge 2 \sampleSize$, then $\mechanism$ is $\tuple{\epsilon}{\delta}$-distribution accurate. 
\end{theorem}

\begin{proof}
From Lemma \ref{lem:GaussMechBayesStabSubGauss} setting $\delta' \coloneqq \frac{\delta}{4 \numOfIterations}$ (where $\delta'$ denotes the parameter $\delta$ used in the Lemma) we get that $\mechanism$ is $\tuple{\frac{\sigma^{2}}{\sampleSize \eta} \sqrt{976 \numOfIterations \lnFunc{\frac{2 \numOfIterations}{\delta}}}}{\frac{\sampleSize^{3} \eta^{3} \delta}{2 \numOfIterations^{\frac{5}{2}} \sigma^{3}}}$-Bayes stable.
From Lemma \ref{lem:GaussAcc}, $\mechanism$ is $\tuple{2 \eta \sqrt{\lnFunc{\frac{4 \numOfIterations}{\delta}}}}{\frac{\delta}{2}}$ Posterior accurate.

Optimizing over $\eta$ we get that $\eta = \left(244 \right)^{\frac{1}{4}} \sigma \sqrt{\frac{\sqrt{\numOfIterations}}{\sampleSize}}$.
Combining this with the bound on $\numOfIterations$ implies $\frac{\sampleSize^{3} \eta^{3}}{\numOfIterations^{\frac{3}{2}} \sigma^{3}} = \frac{\left(224 \right)^{\frac{1}{4}} \sampleSize^{\frac{3}{2}}}{\numOfIterations^{\frac{3}{4}}} \le 2 \numOfIterations^{\frac{3}{4}} \le 2 \numOfIterations$. 
Plugging the definition of $\eta$ back in we get that $\mechanism$ is $\tuple{\sigma \sqrt{\frac{\sqrt{976 \numOfIterations}}{\sampleSize} \lnFunc{\frac{4 \numOfIterations}{\delta}}}}{\frac{\delta}{2}}$-Bayes stable and Posterior accurate. 

Using the bound on $\sampleSize$ we get that $\mechanism$ is $\tuple{\frac{\epsilon}{2}}{\frac{\delta}{2}}$-Bayes stable and Posterior accurate. 
Combining this with Theorem \ref{thm:BayesStabImpdistPcc} completes the proof.
\end{proof}

\subsection{Different or incorrect variances} \label{sec:difIncoVar}
In this section we discuss in detail the case of a mechanism receiving queries with different variances, and then extend the discussion to the effects of analyst relying upon an incorrect bound on the variance of one of the queries.

In the case of a Gaussian mechanism with queries $\query_{i}$, each with variance bounded by $\sigma_{i}^{2}$, the results we achieved translate to a guarantee that with probability $\ge 1 - \delta$, for any $i \in \numOfIterations$, we have
\[
    \abs{\func{\query_{i}}{\dist} - \response_{i}} \le \func{O}{\eta_{i}\sqrt{\lnFunc{\frac{1}{\delta}}} + \sigma_{i} \sqrt{\lnFunc{\frac{1}{\delta}} \sum_{j=1}^{\numOfIterations} \frac{\sigma_{j}^{2}}{\sampleSize^{2} \eta_{j}^{2}}}},
\]
where $\eta_{i}$ is the noise parameter chosen at each iteration.

The $\eta_{i}\sqrt{\lnFunc{\frac{1}{\delta}}}$ term  represents the sample / posterior accuracy bound, and the $\sigma_{i} \sqrt{\lnFunc{\frac{1}{\delta}} \sum_{j=1}^{\numOfIterations} \frac{\sigma_{j}^{2}}{\sampleSize^{2} \eta_{j}^{2}}}$ represents the Bayes stability. Optimizing over $\eta_{i}$, we get that the this error is minimized by setting $\eta_{i} = \func{O}{\sigma_{i} \sqrt{\frac{\sqrt{\numOfIterations}}{\sampleSize}}}$, which implies 
\[
    \abs{\func{\query_{i}}{\dist} - \response_{i}} \le \func{O}{\sigma_{i}\sqrt{\frac{\sqrt{\numOfIterations}}{\sampleSize} \lnFunc{\frac{1}{\delta}}}}.
\]

Now consider a situation where the correct variances were each bounded by $\sigma_{i}^{2}$, but the analyst mistakenly assumed a different bound $\tau_{i}^{2}$ which might be higher or lower than $\sigma_{i}^{2}$. In this case $\eta_{i}$ would have been set to $\func{O}{\tau_{i} \sqrt{\frac{\sqrt{\numOfIterations}}{\sampleSize}}}$, so 
\begin{align*}
    \abs{\func{\query_{i}}{\dist} - \response_{i}} & \le \func{O}{\eta_{i}\sqrt{\lnFunc{\frac{1}{\delta}}} + \sigma_{i} \sqrt{\lnFunc{\frac{1}{\delta}} \sum_{j=1}^{\numOfIterations} \frac{\sigma_{j}^{2}}{\sampleSize^{2} \eta_{j}^{2}}}}
    \\ & = \func{O}{\left(\frac{\tau_{i}}{\sigma_{i}} + \sqrt{\frac{1}{\numOfIterations} \sum_{j=1}^{\numOfIterations} \frac{\sigma_{j}^{2}}{\tau_{j}^{2}}} \right) \sigma_{i} \sqrt{\frac{\sqrt{\numOfIterations}}{\sampleSize} \lnFunc{\frac{1}{\delta}}}},
\end{align*}
which is the same term as in the case of correctly estimated variances, multiplied by the term $\frac{\tau_{i}}{\sigma_{i}} + \sqrt{\frac{1}{\numOfIterations} \sum_{j=1}^{\numOfIterations} \frac{\sigma_{j}^{2}}{\tau_{j}^{2}}}$.

Analyzing this term allows us to precisely understand the implications of an analyst making a mistake in the assumed variance bound.

If $\tau_{i} \ge \sigma_{i}$ for a single query $\query_{i}$ (which means we added too much noise to the response to that query), the first term increased only for that query, and the second term decreased for all queries. On the other hand, if $\tau_{i} < \sigma_{i}$ (which means we did not add enough noise to the response to that query), the first term decreased for that query, but the second term increased for all queries.

The effect on the first term is proportional to the square root of the ratio of the wrong variance to the correct variance of that query, while the effect on the second term is proportional to the square root of average over all $k$ queries of the ratio of the correct variance to the wrong variances (notice the switch between numerator and denominator).

\newpage
\section{Relationships to other stability notions} \label{apd:relNot}
Intuitively speaking, ensuring $\tuple{\epsilon}{\delta}$-differential privacy means that for any query $\query$ and any two neighboring datasets $\sampleSet, \sampleSet'$, with probability $> 1 - \delta$ over the choice of the response $\response$, the log conditional probability ratio $\lnFunc{\frac{\distFuncDep{\response}{\sampleSet}}{\distFuncDep{\response}{\sampleSet'}}}$ is bounded by $\epsilon$ (this quantity is sometimes referred to as the \emph{privacy loss}).\footnote{This intuition is formalized with small degradation in the $\epsilon$ and $\delta$ terms in Lemma 3.3.2 of~\citet{KS14}.}
The most basic method for ensuring $\tuple{\epsilon}{\delta}$-DP in the context of numerical queries is by randomizing the response through noise addition, calibrating the added noise to the \emph{global sensitivity}, that is, the worst-case privacy loss over all queries and all pairs of neighboring datasets.
A series of works initiated by~\citet{DFHPR15} leverages bounds on the privacy loss to provide generalization guarantees for mean estimation under adaptive data analysis. \citet{BNSSSU21} asymptotically improved this guarantee and extended it to low-sensitivity queries and optimization queries. 

Differential privacy's dependence on the global sensitivity naturally yields generalization guarantees that scale with the range of the query's values. However, a dependence on the range of the queries is not required to protect against overfitting in the non-adaptive setting, where the generalization guarantees achievable via Bernstein's inequality scale with the queries' standard deviation. The worst-case nature of DP is not a bug but a feature, since privacy must be ensured for \emph{all} participants under \emph{any} circumstances against \emph{every} potential attack. On the contrary, the task of statistical estimation focuses only on the \emph{typical} case, even in the non-adaptive setting. To overcome the dependence on the queries' range, we seek to decouple DP's stability-preserving powers from its dependence on global sensitivity. This can be done by, in a sense, considering the distribution over the \emph{local sensitivity} induced by the \emph{underlying distribution} over the data, as we discuss in the next sections.

\subsection{Scaling back the noise} \label{sec:locSen}

One approach one might consider to attempt to circumvent DP's worst-case nature in the context of adaptive data analysis is by adding noise that scales like the \emph{local sensitivity} of each query, that is, the privacy loss given the specific query, the specific dataset under study, and the worst-case neighbor of that specific dataset. A variant of this notion was first introduced by~\citet{NRS07}, who illustrated that if the magnitude of noise addition depends na\"{i}vely on the local sensitivity, then the chosen magnitude ``leaks information'' about the dataset, resulting in a mechanism that is not differentially private. To address this, they introduce the \emph{smoothed sensitivity} notion, which smooths out the addition of locally-tailored noise, achieving improved DP guarantees for the task of median estimation (or any other quantile estimation), but not for mean estimation.

Several works have leveraged this result to improve private mean estimation as well, using more involved mechanisms. For example,~\citet{FS17} propose an algorithm that splits a dataset into $m$ subsets, estimates the empirical mean over each one in a non-private manner, and then privately reports the median of the means. They prove that this mechanism is still DP, and provide accuracy guarantees that scale with the queries' standard deviation. Another line of work uses truncation-based specialized mechanisms to provide differential privacy guarantees for Gaussian and sub-Gaussian queries, even in the case of multivariate distribution unknown covariance~\citep{KV17, AL21, DHK23}.~\citet{FS18} use a more natural mechanism for the purpose of generalization under adaptivity, which adds noise that scales like the empirical standard deviation of the query. The generalization guarantees they prove scale like the true standard deviation, but apply only to the expected error.

An alternative method for avoiding the privacy loss that results from scaling the added noise addition to parameters of the actual dataset is keeping the noise level constant, and instead using local sensitivity only when analyzing the mechanism's accuracy and privacy/stability guarantees. This line of work was initiated by~\citet{GR11} and later extended by~\citet{ESS15}, who use similar notions to provide privacy accounting for each individual in the dataset, but do not improve generalization (or privacy) guarantees in the general case. For a survey of this type of analysis, see~\citet{Wang19}. One notable exception is a work by~\citet{FZ20}, who provide improved guarantees under sparsity assumptions on the data---that is, assuming every sample element is relevant only to a small portion of the issued queries.

\subsection{Accounting for the underlying distribution}

The degradation in the guaranteed generalization capabilities caused by scaling the added noise to the global sensitivity of a query might be also mitigated by considering the ``natural noise'' induced by sampling the dataset from some underlying distribution. A variety of relaxations of the DP definition in this vein were introduced by~\citet{GHLP12} and~\citet{BGKS13}.~\citet{BBGLT11}  even consider an alternative privacy notion that relies solely on noise from sampling without any noise added by the mechanism, but all of these approaches suffer from a similar problem: while the privacy guarantee of a single query might improve due to noise introduced by sampling, these definitions have little to no adaptive composition guarantees. A key step in DP's composition proof is the identity $\lnFunc{\frac{\distFuncDep{\response_{\numOfIterations}}{\sampleSet, \view_{\numOfIterations-1}}}{\distFuncDep{\response_{\numOfIterations}}{\sampleSet', \view_{\numOfIterations-1}}}} = \lnFunc{\frac{\distFuncDep{\response_{\numOfIterations}}{\sampleSet}}{\distFuncDep{\response_{\numOfIterations}}{\sampleSet'}}}$, which relies on the guarantee that the mechanism's response is stable, given any auxiliary information (as discussed in the proof of Theorem \ref{thm:PCcomp}). This guarantee can hold only if we condition on the full dataset, which one cannot do when using a definition that leverages the sampling of the data.

Failure to compose---or to compose well---is a common issue with alternative notions of stability that attempt to weaken DP. \citet{LS19} propose a stability notion that is both necessary and sufficient for adaptive generalization (under certain assumptions), but it suffers from limited composition guarantees. \citet{TF20} propose the notion of \emph{Bayesian differential privacy}, which does not enjoy advanced composition guarantees. Instead they prove that the generalization guarantees implied by this notion do compose, but their generalization guarantees still scale with the queries' range in the general case.
A notable exception to these composition limitations is the notion of \emph{Typical stability} introduced by~\citet{BF16}; their approach only requires that the stability notion hold with respect to a subset the of possible datasets, so long as the probability of drawing a dataset outside the set is negligible. Using this notion, they provide generalization guarantees that extend beyond datasets where each element is drawn independently and bounded linear queries, and they obtain guarantees that scale with the variance proxy in case of sub-Gaussian distributions. Unfortunately, this notion only provides accuracy guarantees for a number of queries that scales linearly with the sample size, rather than the quadratic guarantee achieved using DP in the iid case.

Considering the underlying data distribution also paves the way for a Bayesian interpretation of stability. This perspective actually dates back to the work that introduced differential privacy. \citet{DMNS06} interpret the bound on the privacy loss as a bound on the change in the a posteriori probability belief of an analyst, over the sample set held by the mechanism. This interpretation was later formalized by \citet{KS14}, who prove that DP is essentially equivalent to a bound on the Total Variation distance (Definition \ref{def:TVDist}) between the two possible posteriors. Later, \citet{JLNRSS20} use the posterior distribution induced by the transcript of the interaction between the mechanism and the analyst as a key component in their analysis of DP's generalization guarantees under adaptivity, which forms the basis for the ideas presented in this paper.

Some works consider a true Bayesian setting, where the analyst and/or the mechanism hold a prior over possible data distributions.~\citet{MGG09} and~\citet{LQSWY13} use such an assumption to provide stronger privacy guarantees under the assumption that the analyst's knowledge about the distribution is limited.~\citet{YSN15} use this setting to account for non iid data distributions, where knowing of one element in the dataset changes the probability that another element is included in it, thus degrading the privacy of other elements. In the context of adaptive data analysis,~\citet{Elder16} considered a more realistic setting, which removes the gap between the knowledge of the analyst and the mechanism regarding the underlying data distribution. Unfortunately, any assumption about the analyst's knowledge is hard to justify in many realistic scenarios, where implicit priors might be embedded in her algorithm. The guarantees we provide hold even in the case of a fixed distribution which is known to the analyst.

\subsection{Our approach}

In this work we combine these two approaches, leveraging the potential improved stability resulting from low local sensitivity and the properties of the underlying distribution, to avoid the aforementioned potential pitfalls. We do so by introducing a localized stability notion that depends on the specific dataset(s) and on the queries issued during the interaction. Using this notion we can leverage the distribution over the local sensitivity induced by the underlying distribution over the domain. By considering the local sensitivity only in the analysis stage, we avoid the induced stability loss that might be caused, had we scaled the noise with the local sensitivity. Similarly, by choosing a stability notion that depends on the full dataset(s), saving the underlying distribution for the analysis stage, we guarantee that this stability notion composes well.

By transitioning from a global stability notion to a local one, we provide high-probability generalization guarantees that scale with the queries' standard deviation. Furthermore, by removing the dependency on the range, our guarantees extend to unbounded queries as well, so long as they are well-concentrated. Unlike previous works, both of these advantages are achieved even for simple noise-addition mechanisms.

Moving our focus from an a priori stability guarantee to an a posteriori guarantee provided after the fact is similar to the notion of \emph{ex-post privacy} presented by~\citet{LNRWW17}. This approach can also be viewed as a variant of the \emph{metric differential privacy} (mDP) notion~\citep{CABP13, IKWAT22}, for the purpose of adaptive data analysis. Unlike mDP, our notion does not rely on some external fixed metric over elements domain, but rather uses a separate metric at each iteration which is induced by the corresponding query.

The dependence of the stability function on the issued queries also introduces a new challenge. Since these queries are adaptively chosen, the sensitivity is not predefined even by fixing the two datasets, and, in fact, is a random variable. This challenge has been considered with different motivation under the name \emph{full adaptivity}, first by~\citet{RRUV16} and later by~\citet{WRRW23}. In their framework, the sensitivity is fixed, but the privacy parameters are adaptively chosen. The dependence of our PC definition on the actual issued queries during the interaction is analogous to their \emph{privacy odometer}.

Although differential privacy is defined with respect to given $\epsilon$ and $\delta$ parameters, most mechanisms in fact establish a relationship between the two, where each $\epsilon$ has a corresponding $\delta$, and vice-versa (see~\citet{ZDW22} for a survey of possible representations of this relationship). \moshe{Be aware of the order things are presented} ~\citet{Mironov17} provided a simple representation of this relation, by introducing R\'{e}nyi DP (Definition \ref{def:RDP}), which replaces max divergence (Definition \ref{def:maxDiv}) by R\'{e}nyi divergence (Definition \ref{def:RenDiv}) as the stability measure. Building on this notion,~\citet{BS16} introduced a new privacy definition which they call \emph{zero-concentration differential privacy (zCDP)} (Definition \ref{def:zCDP}), and prove that it implies a relationship between the $\epsilon$ and $\delta$ parameters for DP. This definition can also be viewed as a sub-Gaussian bound on the distribution of the privacy loss (Definition \ref{def:stabLoss}), in the style of \emph{Concentrated differential privacy} introduced by~\citet{DR16} (Definition \ref{def:CDP}). Our PC stability notion (Definition \ref{def:PC}) can be viewed as a ``localized'' version of zCDP, replacing the divergence by a dissimilarity measure as discussed in Appendix \ref{apd:divAndDis}, but the same technique could be used to creating a ``localized'' version of DP as well (Definition \ref{def:dynMaxDis}). \citet{TF20} also propose a notion similar to PC, but they do not leverage it to achieve variance-dependent accuracy guarantees.

Although we have managed to transition from worst-case to typical-case guarantees by considering the issued queries and chosen sample set, the guarantees we provide still protect against a worst-case analyst. Using the Cauchy-Schwarz inequality in the proof of the Covariance Lemma (Lemma \ref{lem:covStab}), we bound the stability loss even for queries that are adversarially correlated with the previous view to increase the chance of overfitting. This might be viewed as overkill for a real-world setting that is typically non-adversarial. One possible extension for future work could be to  consider a version of ``natural analysts'', a notion introduced by~\citet{FZ20}, who improve generalization guarantees for such analysts. Alternatively, \citet{DD22} consider a setting where, in each iteration, the issued query is related to a different covariant of the data elements, which are weakly correlated. Although we do not consider this type of relaxation in the present work, the tools we develop here might be used to give improved guarantees under similar assumptions.

\end{document}